\documentclass{article}

\usepackage[numbers,sort&compress]{natbib}

\usepackage[main,final]{neurips_2026}

\usepackage[utf8]{inputenc}
\usepackage[T1]{fontenc}
\usepackage{hyperref}
\usepackage{url}
\usepackage{booktabs}
\usepackage{amsfonts}
\usepackage{nicefrac}
\usepackage{microtype}
\usepackage[table]{xcolor}
\usepackage{amsmath,amssymb}
\usepackage{chngcntr}
\usepackage{multirow}
\usepackage{enumitem}
\usepackage{subcaption}
\usepackage{amsthm}
\usepackage{tabularx}
\usepackage{cleveref}
\usepackage{multicol}
\usepackage{wrapfig}
\theoremstyle{definition}
\usepackage{adjustbox}
\newtheorem{remark}{Remark}[section]
\theoremstyle{plain}
\newtheorem{theorem}{Theorem}[section]

\usepackage{amsmath,amsfonts,bm}

\hypersetup{pdftitle={Solver-as-Teacher: Solver-Guided On-Policy Post-Training Framework for PDE Foundation Models},pdfauthor={Hao Wei, Shangzhe Li, Nils Thuerey}}
\title{Solver-as-Teacher: Solver-Guided On-Policy Post-Training Framework for PDE Foundation Models}

\author{Hao Wei\\
  Technical University of Munich \\
  \texttt{hao.wei@tum.de}
  \And
  Shangzhe Li\\
  UNC Chapel Hill\\
  \texttt{shangzhe@unc.edu}
  \And
  Nils Thuerey\\
  Technical University of Munich \& \\
  Munich Center for Machine Learning \\ \texttt{nils.thuerey@tum.de}
}
\begin{document}

\maketitle

\begin{abstract}
PDE foundation models (PDE-FMs) promise general-purpose surrogate simulation, yet deploying a pretrained model to a new physical regime inevitably introduces distribution mismatch, demanding task-specific post-training. The dominant approach, supervised fine-tuning (SFT), is fundamentally off-policy: the model trains on pre-generated states but is evaluated autoregressively on its own. In chaotic regimes, this mismatch is catastrophic as per-step errors compound exponentially along the Lyapunov spectrum, and exact pointwise trajectory matching beyond the predictability horizon becomes an ill-posed training signal.
To resolve this, we introduce \textbf{Solver-as-Teacher (SaT)}, an on-policy post-training framework for PDE-FMs. During training, the student rolls out autoregressively; a numerical solver branches multi-step correction targets from each student-visited state; and the student is updated against these solver corrections, learning to correct its own rollout dynamics. We instantiate SaT in a unified post-training suite with SFT, physics-informed temporal alignment, and learned-teacher on-policy distillation. Across seven 2D/3D PDE benchmarks and six pretrained PDE-FMs, SaT variants achieve the lowest 50-step rollout RMSE in 23 of 24 in-distribution model--task pairs. The best SaT variant reduces RMSE over SFT up to 90.0\%, and SaT gives the lowest RMSE on all seven OOD parameter-shift benchmarks. Code is available at \url{https://github.com/tum-pbs/Solver-as-Teacher}.
\end{abstract}
\section{Introduction}
\label{sec:introduction}

The solution of partial differential equations (PDEs) is a cornerstone of modern scientific computing.
A rapidly growing body of work trains \emph{PDE foundation models} (PDE-FMs): large neural surrogates pretrained on broad simulation data that learn a time-stepping operator $\mathcal{G}_\theta\colon u(t)\mapsto u(t{+}\Delta t)$ and can be composed autoregressively at inference~\cite{holzschuh2025pdetransformer,herde2024poseidon,mccabe2024mpp,hao2024dpot,mccabe2025walrus,zhang2025omnifluids,rautela2025morph,holzschuh2025p3d,lai2025panda}. Once pretrained, PDE-FMs promise to generalize across diverse physical systems, positioning them as a general engine for scientific computing.

This promise, however, rests on the assumption that pretraining covers the deployment distribution. Systems governed by PDEs differ across physical regimes: boundary conditions, initial conditions, or physical parameters such as the Reynolds number. But even slight mismatches push the model into state-space regions it has rarely seen during pretraining~\cite{koupai2024geps,gupta2025subgridfm}, leading to distribution mismatch. Moreover, there is a fundamental challenge for chaotic regimes: small per-step errors are amplified exponentially along the Lyapunov spectrum, so trajectories diverge from the physical solution manifold, degrade into spectral blur, corrupt attractor statistics, or blow up entirely, even for models with excellent short-horizon metrics~\cite{lippe2023pderefiner,li2022dissipative,he2025chaosattention,nayak2025tideeponet,shikhman2026fnofailure,zou2025uqstability}. Therefore, deploying a pretrained PDE-FM to a new scenario or simply extending the rollout horizon demands task-specific adaptation with proper post-training; without it, long-horizon fidelity is not merely degraded but infeasible.

Yet while pretraining for PDE-FMs has matured rapidly, their post-training methods have not. The most common approach remains supervised fine-tuning (SFT), where the model is trained to minimize pointwise discrepancies between its predictions and reference trajectories. This objective is natural for short-horizon accuracy but fundamentally \textit{off-policy}: the model is optimized on states sampled from a fixed, pre-generated dataset, whereas at inference, it is evaluated autoregressively on states generated by its own previous predictions. This creates a distribution mismatch~\cite{ross2010efficient} between the trajectories seen during training and those generated by the PDE-FMs during inference. The issue is more severe for chaotic and turbulent systems, where small local errors are exponentially amplified by positive Lyapunov exponents, leading to long-horizon distribution drift~\cite{mikhaeil2022difficulty,jiang2023invariant,weiinc}. In such regimes, exact pointwise trajectory matching beyond the predictability horizon becomes an ill-posed training signal: two physically valid trajectories may rapidly decorrelate pointwise while still sharing the same governing dynamics~\cite{li2022dissipative,jiang2023invariant}. Thus, post-training PDE-FMs should not only fit reference trajectories, but learn to recover \textbf{physically meaningful dynamics} from the states produced by their own rollouts.

Although post-training algorithms beyond SFT for LLMs have recently flourished~\cite{tie2025survey}, most do not directly apply to PDE-FMs. Current PDE-FMs are predominantly deterministic neural operators: given a continuous field state, they output a single next-state prediction rather than a probability distribution over tokens, making standard generative alignment techniques inapplicable. Nor does on-policy distillation~\cite{agarwal2024onpolicy} with a larger FM as teacher solve the problem. It closes the distribution gap but inherits the teacher's own approximation errors: in chaotic regimes where per-step errors grow exponentially, the teacher FM's hallucinations transfer directly to the student~\cite{gudibande2023false,li2026rethinking}.

To address this gap, we introduce \textbf{Solver-as-Teacher (SaT)}, a solver-guided on-policy post-training framework designed explicitly for PDE-FMs. We keep the core idea of on-policy distillation: train the model on the continuous states it actually visits during autoregressive rollout, and provide dense expert supervision at those states. SaT uses a numerical solver $\mathcal{S}$ as the teacher. Unlike a learned teacher, the solver provides high-fidelity physical correction targets governed by the underlying PDE, up to discretization and time-stepping error. This enables dense supervision not only on the student's own state distribution, but also ensures the supervision is aligned with the correct physical dynamics. As shown in~\cref{fig:pipeline}, SaT lets the student drive training: (1) at every iteration, the student rolls out autoregressively $\hat{u}_1, \ldots, \hat{u}_K$ from initial state $u_0$; (2) the numerical solver branches forward from each student-visited state $\hat{u}_k$ to produce multi-step correction targets $\mathcal{S}^{j}(\hat{u}_k)$; and (3) the student parameters are updated to minimize the discrepancy between its predictions and the solver-induced dynamics at the states the student actually visits. In this way, SaT converts post-training from static trajectory fitting into solver-guided correction of the model's own rollout dynamics.

Building on this core mechanism, we further cast SaT as one element of a \emph{unified post-training suite} that instantiates complementary training objectives within the same infrastructure: \textbf{On-Policy Distillation (OPD)} reuses the on-policy loop but replaces the solver with a frozen, larger foundation model as teacher~\cite{agarwal2024onpolicy}, isolating the contribution of the solver-teacher signal; and \textbf{PITA}~\cite{zhu2025physics} applies physics-informed PDE residuals as a self-supervised signal for trajectory alignment. Together with standard SFT as the off-policy baseline, SaT, OPD, and PITA form a coherent spectrum of post-training strategies, enabling the first systematic comparison for PDE foundation models. Our main contributions are summarized as follows:
\begin{itemize}[leftmargin=*,topsep=2pt,itemsep=1pt]
    \item \textbf{Solver-as-Teacher on-policy distillation.} We propose SaT, the first on-policy distillation framework for PDE foundation models that uses a numerical solver as the teacher, training the student on its own autoregressive trajectory with dense, per-step solver-derived corrections.
    \item \textbf{Unified post-training suite for PDE-FMs.} We integrate SFT (standard baseline), OPD (larger-FM on-policy distillation), SaT, and PITA (physics-informed trajectory alignment) into a single framework, enabling the first systematic comparison of post-training strategies for PDE-FMs.
    \item \textbf{Extensive evaluation on diverse PDEs and backbones.} We evaluate the suite on seven 2D/3D PDE benchmarks using six pretrained PDE-FM backbones, demonstrating that SaT variants achieve the best 50-step rollout RMSE in 23 of 24 in-distribution model--task pairs and reduce aggregate RMSE over SFT by 29.1\%--90.0\%. Additional experiments extend SaT to $96^3$ resolution and to a second solver family, a finite-volume solver for wall-bounded turbulent channel flows.
\end{itemize}

\section{Related Work}
\label{sec:related}

\textbf{PDE Foundation Models.}
The success of the first PDE-FM~\cite{subramanian2023towards} has spurred a wave of foundation models.
MPP~\cite{mccabe2024mpp} established the multi-physics pretraining paradigm, and its successor Walrus~\cite{mccabe2025walrus} scales to 1.3B parameters.
Poseidon~\cite{herde2024poseidon} introduces a multiscale operator transformer (scOT) with time-conditioned LayerNorm, enabling a single model to handle variable time horizons.
DPOT~\cite{hao2024dpot} adopts a Fourier-attention transformer pre-trained with noise-perturbed inputs on a large collection of PDE simulations.
PDE-Transformer~\cite{holzschuh2025pdetransformer} proposes a Swin-style architecture pretrained on 16 PDE types, while its 3D extension P3D~\cite{holzschuh2025p3d} handles up to 512\textsuperscript{3} resolution via windowed attention with global context fusion.
OmniFluids~\cite{zhang2025omnifluids} combines physics-only pretraining, operator distillation, and few-shot fine-tuning through a mixture-of-operators routing mechanism.
MORPH~\cite{rautela2025morph} supports arbitrary data modalities from 1D to 3D via axial attention with native LoRA support.
Recent studies focus on generalization and robustness during inference. Some incorporate PDE residuals into pretraining~\cite{totounferoush2025paving}, while others leverage In-Context Learning for PDE-FMs~\cite{serrano2024zebra, cao2024vicon, yang2023context}. Others build physical symmetries, such as rotational equivariance, directly into transformer backbones for neural PDE solvers~\cite{weirevit}.
Despite this architectural diversity, most PDE-FMs remain vulnerable to distribution drift between training and inference, especially in chaotic regimes where per-step errors compound exponentially, causing trajectories to diverge from the physical attractor~\cite{lippe2023pderefiner,li2022dissipative,zou2025uqstability,shikhman2026fnofailure}.

\textbf{Remedies for Distribution Drift.}
Remedies for distribution drift have been proposed prior to the current wave of PDE-FMs. The pushforward trick~\cite{brandstetter2022message} trains on multi-step unrolled predictions to reduce exposure bias;
Solver-in-the-Loop~\cite{um2020solverinloop} leverages a differentiable solver to advance the solution in time;
INC~\cite{weiinc} improves stability of hybrid methods by coupling corrections into the governing equations, and PhysicsCorrect~\cite{huang2025physicscorrect} applies solver-residual corrections at inference without updating weights.
For statistical fidelity in chaotic regimes, DySLIM~\cite{schiff2024dyslim} and Jiang et al.~\cite{jiang2023invariant} propose invariant measure matching losses, while Li et al.~\cite{li2022dissipative} demonstrate that Sobolev loss with dissipativity regularization enables stable long-horizon emulation. Other remedies, such as multi-model training with multi-timescales~\cite{bi2023accurate}, noise injection~\cite{stachenfeld2021learned}, and multiresolution targets~\cite{sun2023neural}, reduce errors with increased complexity and, more importantly, are difficult to transfer to PDE-FM post-training.

\textbf{Post-training methods.}
Post-training has become a central tool for improving foundation models, including supervised fine-tuning, RL-based alignment, and knowledge distillation~\cite{tie2025survey,wei2021finetuned,schulman2017proximal,shao2024deepseekmath}. However, these methods are not directly transferable to PDE-FMs. Most PDE-FMs are deterministic, mapping a continuous field to a single next-state prediction rather than producing token-level probability distributions. Thus, LLM-style objectives based on log probabilities, policy ratios, or stochastic sampling require substantial reformulation. Moreover, as long-horizon PDE prediction is sensitive to error accumulation, distillation from a larger teacher may propagate the teacher's own approximation errors~\cite{agarwal2024onpolicy,gudibande2023false,li2026rethinking}.
This gap has motivated post-training methods tailored to PDE-FMs. LoRA-based parameter-efficient adaptation~\cite{hu2022lora} has been explored~\cite{holzschuh2025pdetransformer,rautela2025morph}, but such methods are typically offline fine-tuning procedures. F-Adapter allocates adapter capacity according to spectral complexity~\cite{zhang2025fadapter}, and Spectral-Refiner fine-tunes FNO spectral layers using a Sobolev norm~\cite{cao2025spectralrefiner}; however, these designs are closely tied to spectral architectures and are less directly applicable to transformer-based PDE-FMs. Mansingh et al.~\cite{mansingh2026ttc} introduce inference-time scaling, but rely on converting deterministic PDE-FMs into stochastic samplers via dropout at inference. PITA incorporates physics-informed constraints into self-supervised trajectory alignment~\cite{zhu2025physics}, yet physics residuals serve as soft constraints and, by themselves, do not guarantee long-horizon rollout performance.

\section{Method: Solver-as-Teacher}
\label{sec:method}

\begin{figure}[t]
    \centering
    \includegraphics[width=\linewidth]{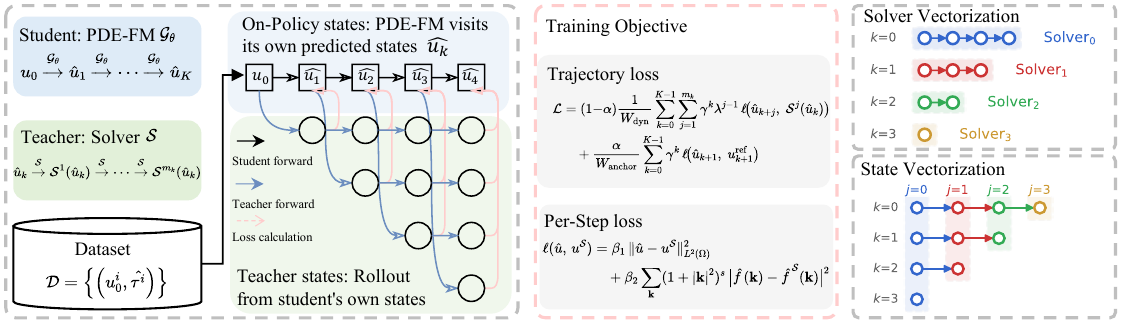}
    \caption{\textbf{SaT overview.} \emph{Left:} the student FM $\mathcal{G}_\theta$ autoregressively rolls out on-policy states $\hat{u}_1,\ldots,\hat{u}_K$ from an initial condition $u_0$;
    the teacher $\mathcal{S}$ branches multi-step correction targets from each student-visited state, as \(\mathcal{S}^j(\hat{u}_k)\) for \(j=1,\dots,m_k\).
    \emph{Center:} the combined training objective from \cref{eq:unified-loss} consists of (i) a dynamics, on-policy multi-step loss that compares the FM's autoregressive predictions against solver-generated chains, and (ii) an anchor loss that directly penalizes deviation from a pre-computed reference trajectory \(u_{k+1}^{\text{ref}}\). The per-step loss in \cref{eq:composite-step} is a weighted combination of an \(L^2\) pointwise error and a Sobolev term.
    \emph{Right:} two vectorization strategies for the multi-step solver branching. \textbf{Solver vectorization}: each branch $k$ is assigned to an independent solver instance, computing its full chain in parallel. \textbf{State vectorization}: all active branches are concatenated into a mega-batch at each depth level $j$ and processed in a single vectorized solver call.}
\vspace{-5pt}
\label{fig:pipeline}
\end{figure}

We now formalize the SaT framework. We first establish notation, then describe the multi-step solver branching mechanism, efficient dispatch strategies, and the training objective.

\paragraph{Preliminaries}
Consider a time-dependent PDE on a periodic spatial domain $\Omega$:
\begin{equation}
    \frac{\partial u}{\partial t} = \mathcal{F}(u, \nabla u, \nabla^2 u, \ldots;\, \mu),
    \qquad u(x,0) = u_0(x),
    \label{eq:pde}
\end{equation}
where $\mathcal{F}$ is the spatial operator and $\mu$ denotes PDE parameters. A \emph{student} FM $\mathcal{G}_\theta \colon u(t) \mapsto u(t{+}\Delta t)$ is applied autoregressively to produce a $K$-step rollout $\hat{\tau} = (\hat{u}_0, \hat{u}_1, \ldots, \hat{u}_K)$ with $\hat{u}_{i+1} = \mathcal{G}_\theta(\hat{u}_i)$. A \emph{teacher} numerical solver $\mathcal{S}$ maps any state $\hat{u}_i$ to the next physical state $\mathcal{S}(\hat{u}_i)$, accurate up to time-stepping and spatial discretization error. Given this setup, the central question is how the teacher solver provides correction targets to the student during on-policy rollout. We write $\mathcal{S}^j(\hat{u}_k)$ for $j$ sequential solver applications starting from $\hat{u}_k$. Throughout, we use $K$ for the rollout horizon and $m_{\max}$ for the lookahead depth of the teacher.

\paragraph{Multi-Step Solver Branching}

From each student-visited state $\hat{u}_k$, the teacher branches forward $m_k$ steps, producing a tree of correction targets (\cref{fig:pipeline}, left):
\begin{equation}
    m_k = \min(m_{\max},\; K - k),
    \qquad
    \mathcal{S}^{j}(\hat{u}_k) \;\;\text{for}\;\; j = 1, \ldots, m_k.
    \label{eq:branching}
\end{equation}
The natural truncation gives longer lookahead early in the trajectory, where the student is closer to the physical manifold and solver targets are most reliable, and shorter lookahead toward the end, where the student has drifted further. Setting $m_{\max} = 1$ recovers one-step on-policy distillation; with $m_{\max} > 1$, the student receives supervision not just on its immediate output but on how its predictions compound over multiple steps. Crucially, the teacher is treated as a black-box oracle: no gradient flows through $\mathcal{S}$. This permits the use of any solver without differentiability requirements, unlike hybrid pipelines~\cite{um2020solverinloop} that backpropagate through the full computation graph of the solver. The black-box property also enables flexible dispatch strategies for the solver evaluations.

\paragraph{Efficient Solver Dispatch}
The multi-step branching structure requires $\sum_{k=0}^{K-1} m_k$ solver evaluations per training step, each applied to a mini-batch of $B$ samples. We implement two complementary vectorization strategies:
\textbf{(1)~Solver vectorization} assigns each branch $k$ to an independent solver instance that computes its full chain $[\mathcal{S}^1(\hat{u}_k), \ldots, \mathcal{S}^{m_k}(\hat{u}_k)]$ in a single fused dispatch. Because each solver instance processes one branch independently, this strategy does not require the solver to support batched evaluation---any black-box time-stepper suffices.
\textbf{(2)~State vectorization} iterates over depth level $j$; at each level, all branches that have not yet terminated are concatenated into a single mega-batch and processed via one batched solver call. This strategy exploits solvers that natively support vectorized batch evaluation (e.g.\ via \texttt{jax.vmap}), amortizing dispatch overhead across all active branches at each depth level and maximizing GPU occupancy.
The two variants are shown in \cref{fig:pipeline}, right.
State vectorization consistently achieves the highest throughput in our benchmarks (${\sim}13\times$ over sequential execution; see~\cref{sec:solver_backends}), while the PyTorch--JAX bridge cost is in \cref{sec:bridge}. We use state vectorization as the default backend throughout all experiments.

\paragraph{Training Objective}
With the solver targets efficiently computed, we now define how the student is trained against them.
Analogous to OPD in language models~\cite{agarwal2024onpolicy}, we formulate a training objective that combines on-policy solver correction with an optional off-policy reference anchor. At each step $k$ of the student's rollout, two supervision signals are available: (1)~the \emph{solver correction} $\mathcal{S}(\hat{u}_k)$, computed on-policy from the student's own state, providing the physically correct next state from wherever the student currently is; and (2)~the \emph{reference trajectory} $u^{\mathrm{ref}}_{k+1}$, the solver-derived reference state from a pre-computed trajectory, providing an off-policy anchor toward the reference solution path.

The full training objective combines a dynamics loss $\mathcal{L}_{\mathrm{dyn}}$ from multi-step solver corrections with an anchor loss $\mathcal{L}_{\mathrm{anchor}}$ from reference trajectories:
\begin{equation}
    \mathcal{L} = (1{-}\alpha)\frac{1}{W_{\mathrm{dyn}}}\sum_{k=0}^{K-1} \sum_{j=1}^{m_k} \gamma^{k} \lambda^{j-1}\;\ell\!\left(\hat{u}_{k+j},\; \mathcal{S}^{j}(\hat{u}_k)\right) +\;
    \frac{\alpha}{W_{\mathrm{anchor}}}\sum_{k=0}^{K-1} \gamma^{k} \ell \left(\hat{u}_{k+1},\; u^{\mathrm{ref}}_{k+1}\right),
    \label{eq:unified-loss}
\end{equation}
where $\gamma,\lambda\in (0,1]$ are decay weights for trajectory position and solver branch depth, $\alpha\in[0,1]$ interpolates between pure on-policy correction and reference-trajectory supervision, and $W_{\mathrm{dyn}},W_{\mathrm{anchor}}$ are normalization constants that ensure a stable learning rate across rollout configurations.
Different values of $\alpha$ yield the key training methods as special cases:
\begin{itemize}[leftmargin=*,topsep=2pt,itemsep=1pt]
    \item $\alpha = 0$: \textbf{SaT-0}, pure on-policy solver correction. The student learns exclusively from the teacher's evaluation of its own states.
    \item $\alpha = 0.5$: \textbf{SaT-.5}, interpolating between on-policy solver correction and off-policy reference supervision. The dynamics loss teaches the model to recover from its own errors; the anchor loss guides it back toward the reference trajectory.
    \item $\alpha = 1$: \textbf{SaT-1}, which reduces to uniformly weighted unrolled supervised fine-tuning~\cite{um2020solverinloop}, with no on-policy correction signal.
\end{itemize}
We recommend $\alpha=0$ (SaT-0) as the default, especially for chaotic systems: it gives the lowest cross-model RMSE on six of seven tasks, is within $1.2\%$ of SaT-.5 on the seventh (KS3D), and is the best method in 10 of the 14 model--task pairs on the chaotic KS and KF tasks (\cref{tab:cross_model_diagnostics_id,tab:rmse_comparison}).

\paragraph{Per-step composite loss.}
The per-step loss $\ell(\cdot,\cdot)$ combines pointwise and spectral supervision:
\begin{equation}
    \ell(\hat{u},\, u^{\mathcal{S}}) = \beta_1 \cdot \|\hat{u} - u^{\mathcal{S}}\|_{L^2(\Omega)}^2 + \beta_2 \cdot \sum_{\mathbf{k}} (1 + |\mathbf{k}|^2)^s \,\bigl|\hat{f}(\mathbf{k}) - \hat{f}^{\mathcal{S}}(\mathbf{k})\bigr|^2,
    \label{eq:composite-step}
\end{equation}
where $\beta_1, \beta_2$ weight the pointwise and spectral terms, and $\hat{f}, \hat{f}^{\mathcal{S}}$ denote the Fourier coefficients of the student and solver outputs, respectively. The Sobolev term penalizes high-frequency discrepancies via the $H^s$ norm~\cite{cao2025spectralrefiner}, computed exactly in Fourier space using Parseval's identity. Together, the unified objective and composite per-step loss define the complete SaT training procedure.

\section{Experimental Configurations}
\label{sec:experiments}

\paragraph{Baselines}

We integrate SaT with different values of $\alpha$ as described in~\cref{sec:method} into a unified post-training suite alongside three complementary methods, enabling the first systematic comparison of post-training strategies for PDE-FMs. \textbf{SFT} (supervised fine-tuning) serves as the baseline; it is widely applied in existing fine-tuning of PDE-FMs~\cite{mccabe2024mpp,hao2024dpot,herde2024poseidon}. \textbf{On-Policy Distillation (OPD)}~\cite{agarwal2024onpolicy} reuses the same on-policy loop as SaT-0, but replaces the numerical solver with a fine-tuned, larger foundation model as teacher, isolating the contribution of solver corrections versus approximate FM corrections.
\textbf{PITA}~\cite{zhu2025physics} (Physics-Informed Temporal Alignment) applies physics-informed constraints as a self-supervised signal for trajectory alignment: the student's autoregressive rollout is penalized by PDE residuals computed from its own predictions, encouraging temporal physical consistency without requiring solver labels at training time.
All methods share the same trainer, configuration system, and evaluation engine; they differ only in the teacher source and loss formulation.

\paragraph{PDE Benchmarks}
We evaluate on seven PDE tasks spanning both 2D and 3D, including chaotic and turbulent regimes. All 2D cases are sourced from APEBench~\cite{koehler2024apebench}, and the 3D cases are generated at higher resolution ($64^3$) by directly instantiating the same Exponax solver used by APEBench. The three 2D tasks are discretized at $160^2$ resolution on $\Omega=[0,2\pi]^2$: \textbf{Kuramoto--Sivashinsky (KS2D)}, a scalar fourth-order PDE exhibiting spatiotemporal chaos through the competition of anti-diffusion and hyper-diffusion; \textbf{Burgers (Burgers2D)}, a scalar conservation law with viscous diffusion that develops sharp fronts and shock structures; and \textbf{Kolmogorov Flow (KF2D)}, forced incompressible Navier--Stokes in vorticity formulation, driven by a sinusoidal body force at mode~4. The four 3D tasks are discretized at $64^3$ resolution on $\Omega=[0,2\pi]^3$: \textbf{KS3D} and \textbf{Burgers3D} extend their 2D counterparts to three spatial dimensions; \textbf{Kolmogorov Flow (KF3D)}, forced 3D incompressible Navier--Stokes in velocity formulation (3~channels, 200-step warm-up to reach statistical stationarity); and \textbf{Decaying Turbulence (DT3D)}, unforced 3D incompressible Navier--Stokes in velocity formulation (3~channels), where energy cascades freely from large to small scales without external forcing. We additionally evaluate on out-of-distribution (OOD) zero-shot test sets to probe generalization; dataset specifications are provided in \cref{sec:appendix:zeroshot}. The governing equations, complete solver parameter tables, and data splits are provided in \cref{sec:appendix:experiment}. Results at $96^3$ resolution and on wall-bounded turbulent channel flow (TCF3D) with a second solver are reported in \cref{sec:appendix:highres,sec:appendix:tcf}.

\paragraph{Models and Teacher}
We evaluate SaT across six pretrained PDE foundation models that span diverse architecture families and cover both 2D and 3D spatial domains. \textbf{PDE-S}~\cite{holzschuh2025pdetransformer} (33.2M params) is a Swin-like U-Net Transformer with hierarchical shifted-window attention. \textbf{MPP-B}~\cite{mccabe2024mpp} (115.6M params) employs an Axial Vision Transformer (AViT) with interleaved spatial and temporal axial attention, pretrained across heterogeneous physical systems. \textbf{DPOT-Ti}~\cite{hao2024dpot} uses Adaptive Fourier Neural Operator (AFNO) spectral mixing blocks that apply learnable complex-valued MLPs to retained Fourier modes; it is the only model applied to both 2D (5.2M params) and 3D (12.7M params), where the 3D instance keeps the same DPOT-Ti block width/depth but replaces the 2D patch embedding, positional embedding, and output reconstruction with 3D counterparts. \textbf{Poseidon-T}~\cite{herde2024poseidon} (20.7M params) builds on ScOT, a SwinV2-style scalable operator transformer with built-in spectral resolution adaptation. For 3D, \textbf{P3D-S}~\cite{holzschuh2025p3d} (11.2M params) combines a Conv3D U-Net encoder/decoder with windowed 3D Transformer blocks, and \textbf{MORPH-Ti}~\cite{rautela2025morph} (9.9M params) uses 4D axial attention decomposed along the temporal, depth, height, and width axes with inter-field cross-attention. All student models are fine-tuned with full weights in our experiments. Detailed model configurations are provided in~\cref{sec:appendix:model_details}. The teacher solver is Exponax~\cite{koehler2024apebench}, a JAX-based ETDRK pseudo-spectral solver. We deliberately choose a JAX solver within a PyTorch training pipeline to demonstrate that SaT is agnostic to the solver backend. To test a second solver family, \cref{sec:appendix:tcf} replaces Exponax with PICT~\cite{franz2025pict}, a PyTorch finite-volume PISO solver, on wall-bounded turbulent channel flow. For OPD, the teacher is PDE-L (518M params, 2D) or P3D-L (181M params, 3D), fine-tuned before being used for training the students. Experimental details and training hyperparameters are provided in~\cref{sec:appendix:experiment};  end-to-end wall-clock benchmarks comparing SaT, SFT, and OPD are in~\cref{sec:method_efficiency}; the JAX$\leftrightarrow$PyTorch bridge overhead is characterized in~\cref{sec:bridge}; and the efficiency between state vectorization and solver vectorization is detailed in~\cref{sec:solver_backends}.

\section{Results}
\label{sec:results}

\begin{table}[htbp]
\vspace{-5pt}
\centering
\caption{Cross-model aggregate diagnostics on in-distribution test sets for 50 steps, across 7 PDE benchmarks. Each entry is the mean $\pm$ std over evaluated architectures (4 for 2D, 3 for 3D); lower is better. Best per row \textbf{bolded}. RMSE, Sobolev $H^{1}$, and normalized energy-spectrum errors are defined in \cref{sec:appendix:physical_diagnostics}; per-model RMSE values are reported in \cref{tab:rmse_comparison}. PITA does not support 3D tasks (--).}
\label{tab:cross_model_diagnostics_id}
\vspace{2pt}
\scriptsize
\setlength{\tabcolsep}{2.4pt}
\begin{adjustbox}{max width=\textwidth}
\begin{tabular}{@{}ll c c c c c c c@{}}
\toprule
Task & Metric & SFT & PITA & OPD & SaT-1 & SaT-.5 & SaT-0 & N models \\
\midrule
\multirow{3}{*}{Burgers2D} & RMSE     & $0.0378 \pm 0.0298$ & $0.0232 \pm 0.0172$ & $0.0193 \pm 0.0096$ & $0.0138 \pm 0.0038$ & $0.0141 \pm 0.0097$ & $\mathbf{0.0118 \pm 0.0040}$ & \multirow{3}{*}{4} \\
 & $H^1$    & $0.0380 \pm 0.0300$ & $0.0232 \pm 0.0172$ & $0.0193 \pm 0.0096$ & $0.0139 \pm 0.0038$ & $0.0142 \pm 0.0097$ & $\mathbf{0.0118 \pm 0.0041}$ & \\
 & Spectrum & $0.0502 \pm 0.0576$ & $0.0234 \pm 0.0263$ & $0.0223 \pm 0.0163$ & $0.0090 \pm 0.0025$ & $0.0150 \pm 0.0194$ & $\mathbf{0.0077 \pm 0.0059}$ & \\
\midrule
\multirow{3}{*}{KF2D} & RMSE     & $0.9023 \pm 0.4394$ & $0.8017 \pm 0.2627$ & $0.7845 \pm 0.2827$ & $0.5884 \pm 0.2246$ & $0.7554 \pm 0.3107$ & $\mathbf{0.5409 \pm 0.2089}$ & \multirow{3}{*}{4} \\
 & $H^1$    & $0.9032 \pm 0.4399$ & $0.8024 \pm 0.2629$ & $0.7852 \pm 0.2829$ & $0.5890 \pm 0.2248$ & $0.7561 \pm 0.3110$ & $\mathbf{0.5414 \pm 0.2091}$ & \\
 & Spectrum & $0.0250 \pm 0.0135$ & $0.0169 \pm 0.0090$ & $0.0179 \pm 0.0009$ & $0.0166 \pm 0.0060$ & $0.0210 \pm 0.0098$ & $\mathbf{0.0126 \pm 0.0054}$ & \\
\midrule
\multirow{3}{*}{KS2D} & RMSE     & $0.1187 \pm 0.0784$ & $0.1084 \pm 0.0545$ & $0.1016 \pm 0.0422$ & $0.0913 \pm 0.0460$ & $0.0844 \pm 0.0489$ & $\mathbf{0.0841 \pm 0.0419}$ & \multirow{3}{*}{4} \\
 & $H^1$    & $0.1191 \pm 0.0787$ & $0.1088 \pm 0.0547$ & $0.1019 \pm 0.0423$ & $0.0916 \pm 0.0462$ & $0.0847 \pm 0.0491$ & $\mathbf{0.0843 \pm 0.0421}$ & \\
 & Spectrum & $0.0778 \pm 0.0733$ & $0.0438 \pm 0.0203$ & $0.0440 \pm 0.0171$ & $0.0338 \pm 0.0121$ & $0.0365 \pm 0.0050$ & $\mathbf{0.0307 \pm 0.0131}$ & \\
\midrule
\multirow{3}{*}{Burgers3D} & RMSE     & $0.0154 \pm 0.0042$ & -- & $0.0109 \pm 0.0012$ & $0.0082 \pm 0.0015$ & $0.0091 \pm 0.0016$ & $\mathbf{0.0080 \pm 0.0014}$ & \multirow{3}{*}{3} \\
 & $H^1$    & $0.0157 \pm 0.0043$ & -- & $0.0110 \pm 0.0012$ & $0.0083 \pm 0.0014$ & $0.0092 \pm 0.0016$ & $\mathbf{0.0081 \pm 0.0014}$ & \\
 & Spectrum & $0.0562 \pm 0.0370$ & -- & $0.0290 \pm 0.0100$ & $0.0227 \pm 0.0003$ & $0.0201 \pm 0.0106$ & $\mathbf{0.0180 \pm 0.0072}$ & \\
\midrule
\multirow{3}{*}{KF3D} & RMSE     & $0.6987 \pm 0.7455$ & -- & $0.1423 \pm 0.0143$ & $0.1366 \pm 0.0115$ & $0.1328 \pm 0.0136$ & $\mathbf{0.1292 \pm 0.0155}$ & \multirow{3}{*}{3} \\
 & $H^1$    & $0.7142 \pm 0.7662$ & -- & $0.1431 \pm 0.0143$ & $0.1374 \pm 0.0115$ & $0.1336 \pm 0.0137$ & $\mathbf{0.1300 \pm 0.0155}$ & \\
 & Spectrum & $0.1431 \pm 0.1788$ & -- & $0.0616 \pm 0.0693$ & $\mathbf{0.0250 \pm 0.0125}$ & $0.0283 \pm 0.0106$ & $0.0275 \pm 0.0161$ & \\
\midrule
\multirow{3}{*}{KS3D} & RMSE     & $2.2190 \pm 2.8475$ & -- & $0.3768 \pm 0.2084$ & $0.2893 \pm 0.1132$ & $\mathbf{0.2211 \pm 0.0310}$ & $0.2238 \pm 0.0440$ & \multirow{3}{*}{3} \\
 & $H^1$    & $2.2237 \pm 2.8529$ & -- & $0.3784 \pm 0.2092$ & $0.2910 \pm 0.1143$ & $\mathbf{0.2222 \pm 0.0313}$ & $0.2250 \pm 0.0446$ & \\
 & Spectrum & $0.1688 \pm 0.1718$ & -- & $0.0963 \pm 0.0129$ & $0.0931 \pm 0.0736$ & $0.0859 \pm 0.0373$ & $\mathbf{0.0837 \pm 0.0593}$ & \\
\midrule
\multirow{3}{*}{DT3D} & RMSE     & $0.2554 \pm 0.2923$ & -- & $0.0398 \pm 0.0171$ & $0.0358 \pm 0.0141$ & $0.0374 \pm 0.0147$ & $\mathbf{0.0304 \pm 0.0091}$ & \multirow{3}{*}{3} \\
 & $H^1$    & $0.2586 \pm 0.2964$ & -- & $0.0401 \pm 0.0172$ & $0.0361 \pm 0.0142$ & $0.0377 \pm 0.0148$ & $\mathbf{0.0306 \pm 0.0091}$ & \\
 & Spectrum & $0.1554 \pm 0.1019$ & -- & $0.0673 \pm 0.0481$ & $0.0739 \pm 0.0452$ & $0.0495 \pm 0.0297$ & $\mathbf{0.0267 \pm 0.0107}$ & \\
\bottomrule
\end{tabular}
\end{adjustbox}
\vspace{-5pt}
\end{table}

\subsection{In-Distribution Benchmark Results}
\label{sec:results:indist}

\begin{figure}[t]
\centering
\begin{subfigure}[t]{0.47\textwidth}
  \centering
  \includegraphics[width=1.05\textwidth]{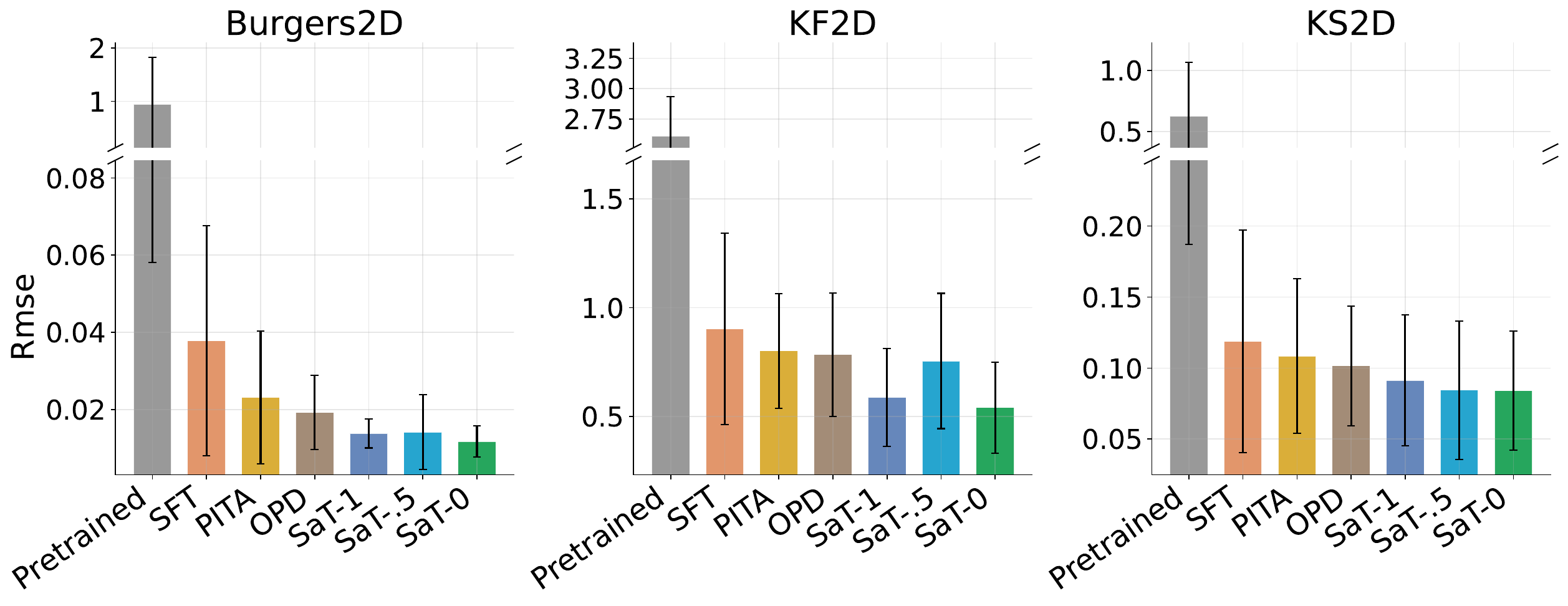}
  \caption{2D tasks.}
  \label{fig:id_2d}
\end{subfigure}
\hfill
\begin{subfigure}[t]{0.515\textwidth}
  \centering
  \includegraphics[width=\textwidth]{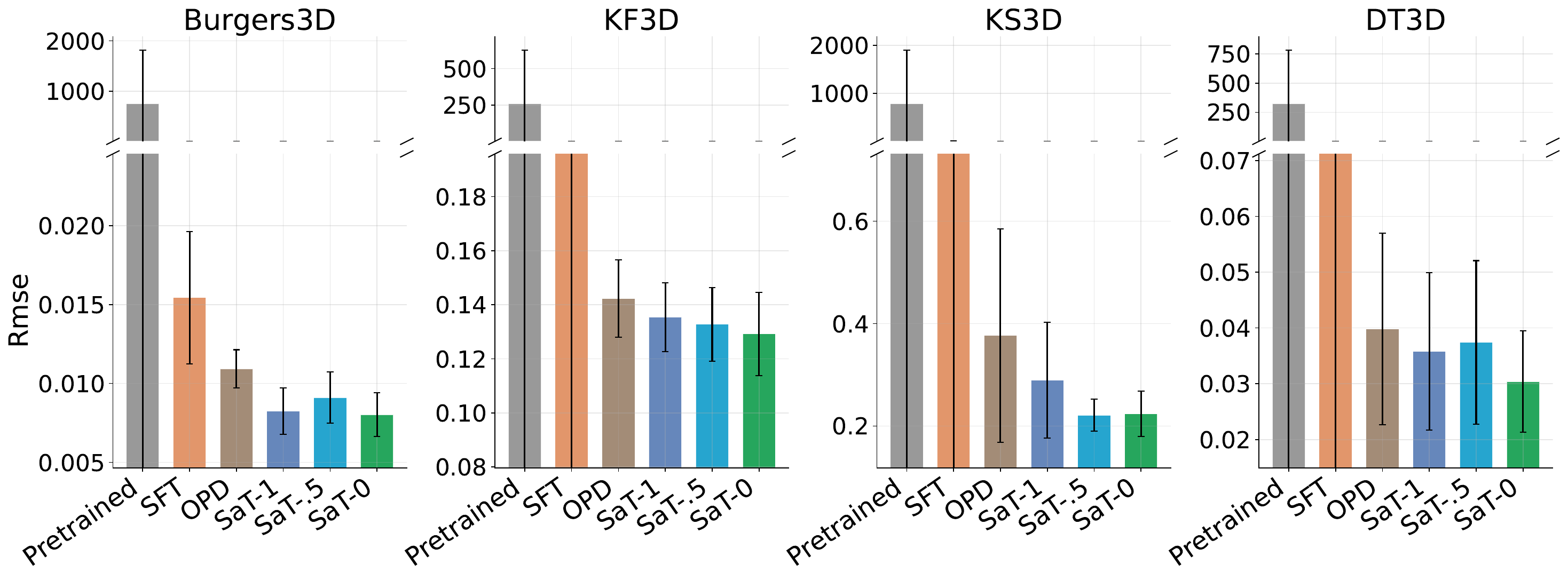}
  \caption{3D tasks.}
  \label{fig:idd_3d}
\end{subfigure}
\caption{In-distribution rollout RMSE averaged across evaluated architectures: four models for 2D tasks and three models for 3D tasks. SaT-0 achieves the lowest cross-model RMSE on six of seven tasks, while SaT-.5 is best on KS3D.}
\label{fig:id_bar}
\vspace{-5pt}
\end{figure}

\paragraph{SaT variants dominate long-rollout accuracy.}
Table~\ref{tab:cross_model_diagnostics_id} reports cross-model aggregate diagnostics on 50-step rollouts for the 6 foundation models across 7 PDE benchmarks; precise definitions of RMSE, the Sobolev $H^{1}$ error, and the normalized energy-spectrum error are given in \cref{sec:appendix:physical_diagnostics}, and the per-model RMSE breakdown is in \cref{sec:appendix:rmse_per_model,tab:rmse_comparison}. Across 24 model--task pairs (12 for 2D, 12 for 3D), a SaT variant gives the best per-model RMSE in 23 cases, with the only exception being MORPH-Ti on DT3D where OPD is slightly lower.
The aggregated trends per scenario are  even clearer: SaT-0 achieves the lowest cross-model RMSE on six of seven tasks, while SaT-.5 is best on KS3D and differs only marginally from SaT-0 ($0.2211$ vs.\ $0.2238$; \cref{tab:cross_model_diagnostics_id}). Relative to SFT, the best SaT variant reduces cross-model RMSE by up to $68.8\%$ on 2D cases (Burgers2D) and up to $90.0\%$ on 3D cases ($81.5\%$ on KF3D, $90.0\%$ on KS3D, and $88.1\%$ on DT3D).
These results are shown in~\cref{fig:id_bar}.
Thus, SaT improves both the per-architecture best case and the architecture-averaged behavior, which is the relevant criterion for post-training a family of heterogeneous PDE-FMs rather than tuning a single backbone. This ranking is not an artifact of pointwise RMSE alone: the trajectory-level Sobolev $H^1$ and normalized energy-spectrum of \cref{tab:cross_model_diagnostics_id} show a SaT variant best on all 7 in-distribution tasks for both diagnostics, with SaT-0 best on 6 of 7 tasks for each metric. These diagnostics strengthen the interpretation that solver correction improves physically meaningful dynamics, including gradient fidelity and scale-wise energy allocation, rather than merely reducing amplitude error.

\paragraph{Full dynamic loss stabilizes chaotic rollouts.}
Unrolled training (SaT-1) accounts for most of the improvement over SFT. Replacing its reference targets with solver targets lowers RMSE further in 22 of 24 model--task pairs and on every task's cross-model mean, by $2.4\%$ (Burgers3D) to $22.6\%$ (KS3D). This gap remains when SaT-1 receives as many targets per epoch from a larger static dataset (\cref{sec:appendix:static_data}) and when the methods are compared at equal wall-clock time (\cref{sec:appendix:budget}). The relative ranking of SaT-1, SaT-.5, and SaT-0 clarifies the role of the reference anchor in~\cref{eq:unified-loss}. SaT-0 uses the full dynamic loss ($\alpha=0$): every target is generated by the solver from a student-visited state, so the training signal remains on-policy and physically grounded even after the student has left the reference trajectory. This is exactly the regime where pointwise trajectory matching becomes ill-conditioned: small errors are amplified by chaotic dynamics, and an off-policy reference state can become a poor target for correcting the student's current state. Full dynamic loss therefore dominates the hardest settings, winning 9 of 12 3D model--task pairs. Conversely, the anchor-containing variants remain useful in simpler or more weakly chaotic 2D cases, where the reference trajectory is still informative over the training horizon: SaT-.5 wins 6 of 12 2D model--task pairs. This pattern supports the intended interpretation of $\alpha$: reference loss can accelerate alignment when the trajectory target remains meaningful, while full dynamic loss is more robust once autoregressive drift and chaotic decorrelation dominate.

\begin{figure}[t]
\centering
\includegraphics[width=\textwidth]{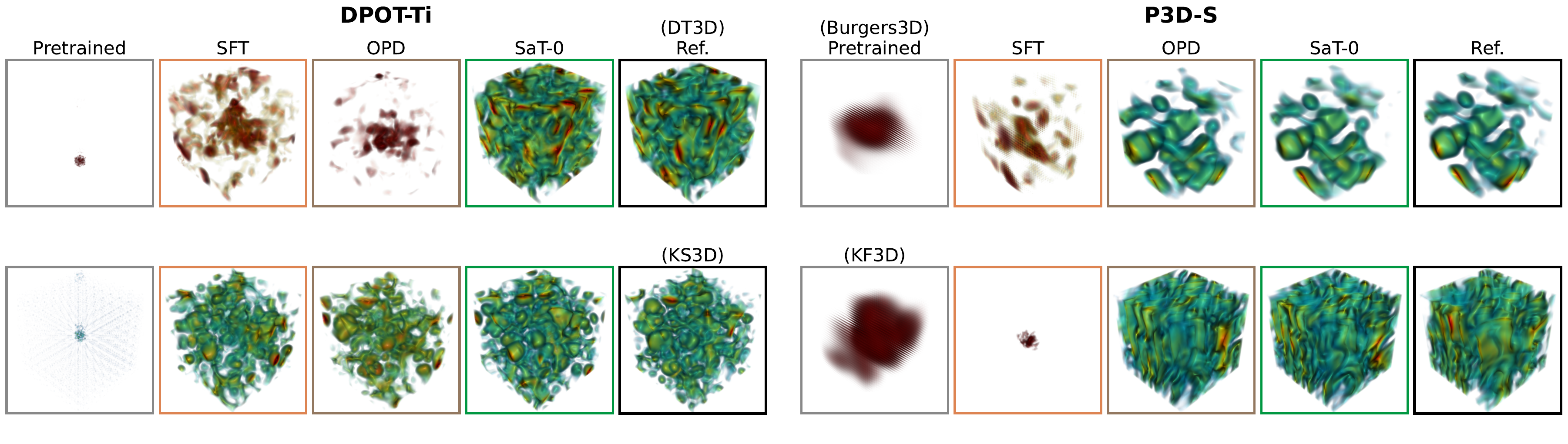}
\caption{Volumetric fields after 50 autoregressive rollout steps. \textbf{Left block}: DPOT-Ti on DT3D and KS3D. \textbf{Right block}: P3D-S on Burgers3D and KF3D. SFT collapses into diffuse, sparse, or underdeveloped structures, while OPD partially improves morphology but can still miss the solver reference. SaT-0 recovers coherent volumetric structure and closely matches the reference fields, especially for P3D-S, where SFT is weakest.}
\label{fig:fields_3d_grid}
\vspace{-5pt}
\end{figure}

\paragraph{SaT rescues temporally under-specified backbones.}
The strongest architecture-level result is not only that SaT improves average RMSE, but that its gains are largest for backbones whose temporal signal is weakest. DPOT-Ti is pretrained with autoregressive denoising and a multi-frame time aggregator, MPP-B and MORPH-Ti contain explicit temporal attention mechanisms, and Poseidon-T is pretrained with time-conditioned semigroup pairs. In contrast, PDE-S and its 3D extension P3D-S are single-frame transformer backbones, with neither temporal-history input nor temporal-history encoding into structures. This makes them especially vulnerable under SFT, which trains on reference states but never corrects the model's own rollout distribution. SaT directly supplies the missing temporal supervision through solver corrections on student-visited states. Under SFT, P3D-S is the weakest 3D model on all four tasks; with SaT-0, it becomes the best model on Burgers3D and DT3D, reducing RMSE up to $97.4\%$ on DT3D ($0.6687$ to $0.0175$). The same improvement shows in harder chaotic tasks: P3D-S improves from $6.2459$ to $0.2534$ on KS3D with SaT-.5 ($95.9\%$), and from $1.7530$ to $0.1478$ on KF3D with SaT-0 ($91.6\%$). The 2D sibling PDE-S shows the same pattern, with SaT-0 reducing RMSE up to $73.1\%$ on Burgers2D. The visual comparison in~\cref{fig:fields_3d_grid} shows the 3D transition: SFT collapses toward diffuse, sparse, or underdeveloped structures, whereas SaT-0 recovers volumetric fields with the correct large-scale morphology. Thus, SaT can supply the missing temporal training signal through the on-policy solver loop itself, allowing even backbones without intrinsic temporal-history encoding to produce stable long rollouts.

\begin{figure}[t]
\centering
\includegraphics[width=\textwidth]{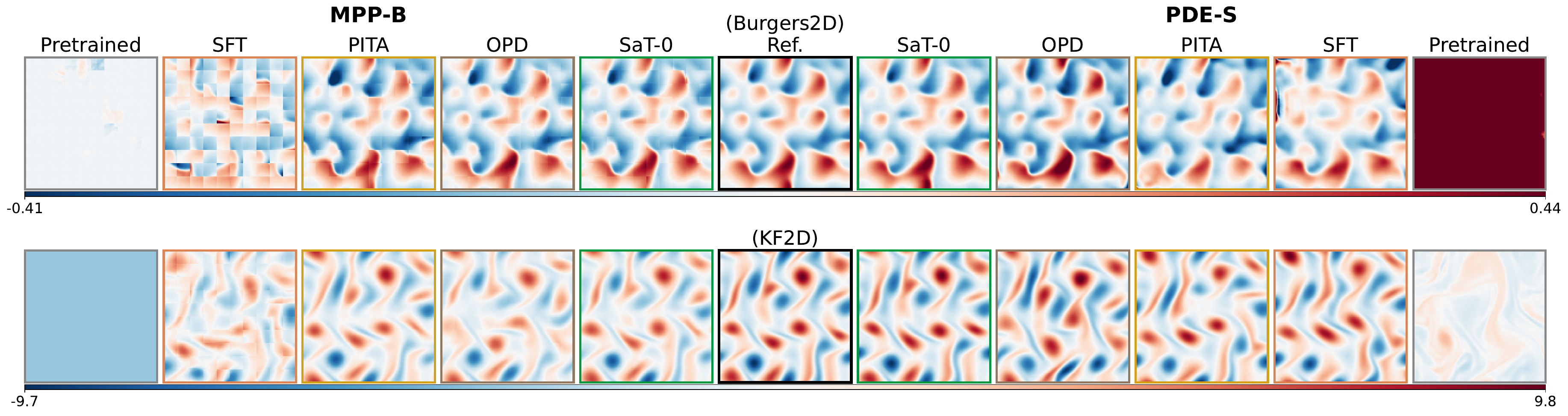}
\caption{Predicted 2D fields after 50 autoregressive rollout steps; MPP-B (left) and PDE-S (right) on Burgers2D (top) and KF2D (bottom). The solver reference is shown at the center. MPP-B's SFT prediction exhibits pronounced blocky artifacts, arising from the AViT's independent per-patch sparse output projection. SaT-0 removes these artifacts and recovers physically coherent fine-scale structure. PDE-S, whose Swin U-Net decoder fuses multi-scale skip connections, produces smoother SFT fields; SaT further sharpens the prediction toward the reference.}
\label{fig:fields_mpp_vs_pdes}
\end{figure}

\paragraph{On-policy correction removes architecture-specific artifacts.}
Beyond aggregate metrics, \cref{fig:fields_mpp_vs_pdes} exposes a 2D failure mode that SaT corrects. MPP-B's SFT predictions exhibit a pronounced $10 \times 10$ grid of blocky artifacts,
an architectural fingerprint of the independent per-patch sparse output projection, which reconstructs each patch without cross-patch convolution or overlap. SFT exacerbates this because the model never trains on its own patchwise artifacts; at inference, small inter-patch inconsistencies are repeatedly fed back and compound across autoregressive steps. SaT corrects the same failure at its source: the on-policy rollout exposes these boundary artifacts to the loss, and dense solver corrections force the next-step map to re-enter a spatially coherent physical field. As shown in~\cref{fig:fields_mpp_vs_pdes}, SaT-0 eliminates the blocky structure and recovers smooth fields close to the solver reference. In contrast, PDE-S's Swin U-Net decoder already provides overlapping receptive fields through hierarchical skip connections and shifted-window attention, so SFT is smoother; SaT further sharpens the solution toward the reference.

\vspace{-6pt}
\paragraph{Solver teachers are more reliable than learned teachers.}
OPD tests the central limitation discussed in~\cref{sec:introduction}: replacing the numerical solver with a larger PDE-FM also closes the distribution gap, but the teacher remains an approximate neural simulator. This distinction matters in a chaotic rollout. On KF2D, OPD degrades DPOT-Ti relative to SFT from $0.2316$ to $0.4018$ and PDE-S from $1.0938$ to $1.1566$. Even when OPD improves over SFT in some cases, the solver teacher remains stronger: on KS3D, the cross-model OPD RMSE is $0.3768$, while SaT-.5 reaches $0.2211$; on DT3D, OPD gives $0.0398$, while SaT-0 gives $0.0304$. The reason is structural rather than merely empirical. A learned teacher can provide a dense target, but its target is still produced by a model that may hallucinate missing scales, damp turbulent energy, or inherit pretraining mismatch. The numerical solver provides the PDE evolution at the student's actual state, so its correction remains tied to the governing dynamics even when the student has drifted away from the data trajectory. PITA shows the complementary limitation of residual-only alignment: it improves MPP-B on KF2D ($1.4343$ to $0.8410$) but degrades DPOT-Ti on the same task ($0.2316$ to $0.3674$), and on Burgers2D it improves DPOT-Ti ($0.0126$ to $0.0103$) but degrades PDE-S ($0.0446$ to $0.0515$). Soft physics constraints are therefore useful for some architecture--PDE pairs, but less consistent than explicit solver targets. Beyond accuracy, the solver teacher is also substantially cheaper: on KS3D, SaT-0 completes training in ${\sim}3.1$--$5.9$\,h depending on the backbone, while OPD requires ${\sim}18.7$--$22.0$\,h for the student training alone plus ${\sim}14.5$\,h of teacher preparation, for a total cost of ${\sim}33$--$37$\,h, or about $6.2$--$10.9{\times}$ the SaT-0 wall-clock time (\cref{sec:appendix:wallclock,tab:wallclock_ks3d}).

\subsection{Out-of-Distribution Zero-Shot Results}
\label{sec:results:ood}

\begin{table}[htbp]
\vspace{-5pt}
\centering
\caption{OOD zero-shot RMSE (mean $\pm$ std) for 2D and 3D tasks with perturbed physical parameters (see \cref{sec:appendix:zeroshot}). Best per task \textbf{bolded}. PITA does not support 3D tasks (--).}
\label{tab:ood_rmse_combined}
\small
\resizebox{\textwidth}{!}{
\begin{tabular}{@{}l cccccc c@{}}
\toprule
\textbf{Task} & \textbf{SFT} & \textbf{PITA} & \textbf{OPD} & \textbf{SaT-1} & \textbf{SaT-.5} & \textbf{SaT-0} & \textbf{N models} \\
\midrule
Burgers2D-OOD & $0.0856 \pm 0.0157$ & $0.0770 \pm 0.0079$ & $0.0738 \pm 0.0016$ & $0.0727 \pm 0.0014$ & $0.0728 \pm 0.0021$ & $\mathbf{0.0723 \pm 0.0010}$ & 4 \\
KS2D-OOD      & $0.1373 \pm 0.0310$ & $0.1200 \pm 0.0347$ & $0.1262 \pm 0.0282$ & $0.1251 \pm 0.0307$ & $0.1069 \pm 0.0261$ & $\mathbf{0.1060 \pm 0.0175}$ & 4 \\
KF2D-OOD      & $2.4524 \pm 0.2899$ & $2.4685 \pm 0.2256$ & $2.4226 \pm 0.2118$ & $2.3316 \pm 0.1738$ & $2.4989 \pm 0.2597$ & $\mathbf{2.2652 \pm 0.2140}$ & 4 \\
\midrule
Burgers3D-OOD & $0.0175 \pm 0.0024$ & -- & $0.0143 \pm 0.0020$ & $0.0131 \pm 0.0012$ & $0.0141 \pm 0.0020$ & $\mathbf{0.0129 \pm 0.0010}$ & 3 \\
KF3D-OOD      & $1.0658 \pm 1.1758$ & -- & $0.2021 \pm 0.0056$ & $0.1988 \pm 0.0042$ & $0.1963 \pm 0.0049$ & $\mathbf{0.1929 \pm 0.0045}$ & 3 \\
KS3D-OOD      & $2.6798 \pm 3.5134$ & -- & $0.2897 \pm 0.0961$ & $0.2004 \pm 0.0188$ & $0.1956 \pm 0.0168$ & $\mathbf{0.1920 \pm 0.0226}$ & 3 \\
DT3D-OOD      & $0.2487 \pm 0.2838$ & -- & $0.0435 \pm 0.0114$ & $0.0425 \pm 0.0063$ & $0.0432 \pm 0.0076$ & $\mathbf{0.0370 \pm 0.0028}$ & 3 \\
\bottomrule
\end{tabular}
}
\vspace{-5pt}
\end{table}

\begin{figure}[t]
\centering
\begin{subfigure}[t]{0.47\textwidth}
  \vspace{0pt}
  \centering
  \includegraphics[width=\textwidth]{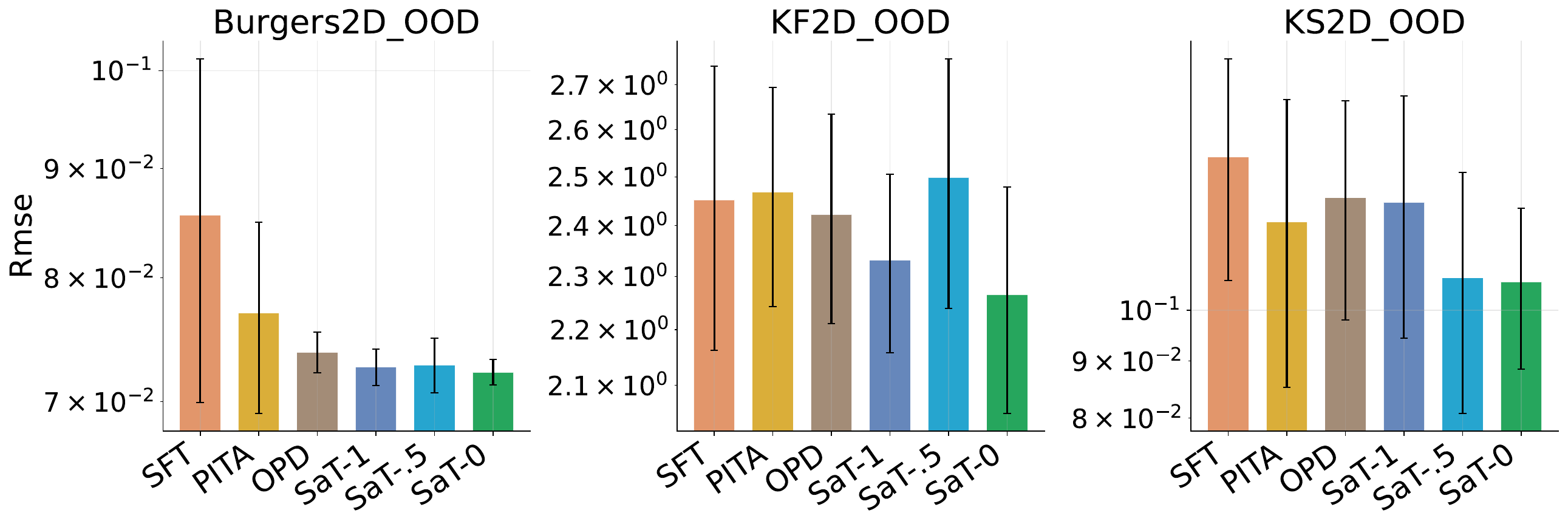}
  \caption{2D OOD tasks.}
  \label{fig:ood_2d}
\end{subfigure}
\hfill
\begin{subfigure}[t]{0.48\textwidth}
  \vspace{0pt}
  \centering
  \includegraphics[width=\textwidth]{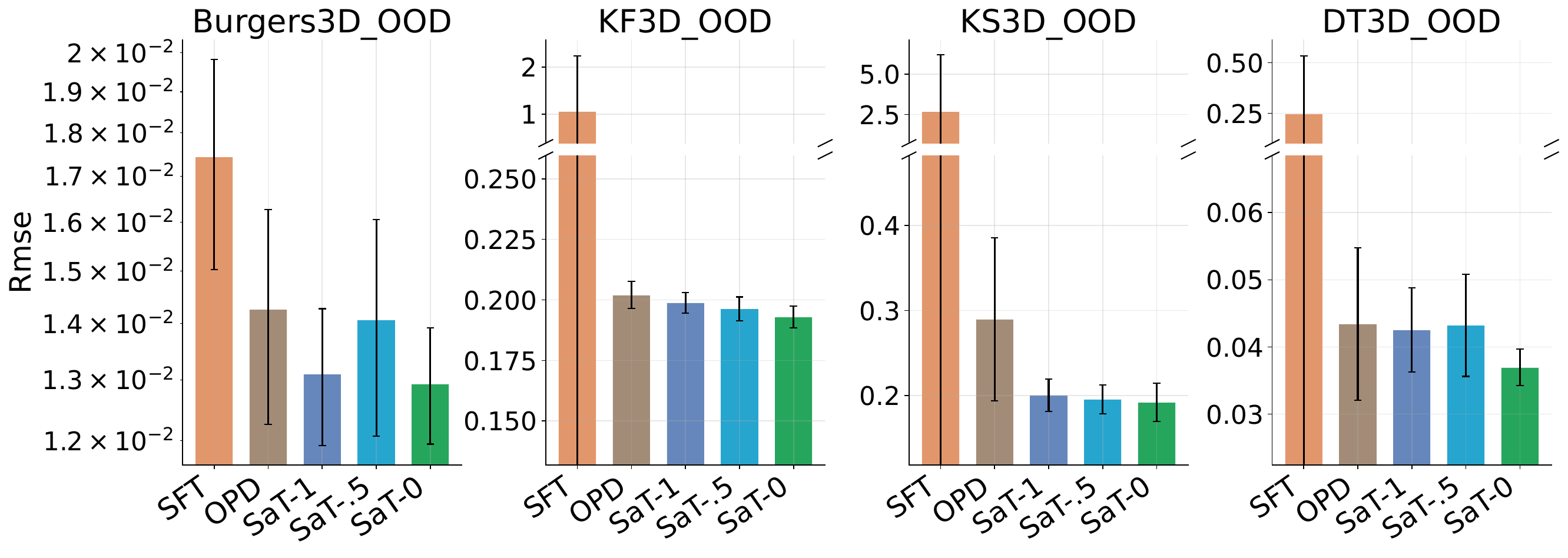}
  \caption{3D OOD tasks.}
  \label{fig:ood_3d}
\end{subfigure}
\caption{Zero-shot RMSE on OOD tasks with perturbed physical parameters, averaged across four 2D models or three 3D models. SaT-0 achieves the lowest cross-model RMSE on all 7 OOD tasks.}
\label{fig:ood_bar}
\vspace{-5pt}
\end{figure}

\paragraph{Full dynamic loss generalizes to shifted physics.}
Table~\ref{tab:ood_rmse_combined} and \cref{fig:ood_bar} report zero-shot RMSE under physical-parameter shifts: stronger nonlinear coupling for KS, stronger turbulence for KF, higher viscosity for DT3D, and weakened diffusion and convection for Burgers. SaT-0 gives the lowest cross-model RMSE on all seven OOD tasks, with the largest gains in 3D: $81.9\%$ on KF3D-OOD, $92.8\%$ on KS3D-OOD, and $85.1\%$ on DT3D-OOD relative to SFT. This supports the main mechanism rather than task-specific fitting: the student is trained on its own rollout states, including states that have drifted from the reference data, and the solver provides dynamically consistent supervision at each of them. The appendix diagnostics lead to the same conclusion beyond RMSE: \cref{tab:physical_diagnostics_ood} shows SaT-0 best on all seven OOD $H^1$ tasks, and a SaT variant best on six of seven normalized spectrum-error tasks. Per-model OOD results are reported in \cref{sec:appendix:ood_detailed,tab:ood_detailed_combined}.

\section{Concluding Remarks}
\label{sec:conclusion}

We introduced \textbf{Solver-as-Teacher (SaT)}, an on-policy post-training framework that uses a numerical solver as a dense teacher for the states visited by a PDE-FM during its own autoregressive rollout. This changes post-training from offline trajectory fitting into solver-guided correction of the model's induced dynamics. Across seven 2D and 3D PDE benchmarks and six pretrained backbones, the experiments show that SaT variants achieve the best 50-step rollout RMSE in 23 of 24 in-distribution model--task pairs, reduce cross-model RMSE over SFT by up to $90.0\%$, and give the lowest cross-model RMSE on all seven OOD parameter-shift benchmarks. The method comparisons further support the mechanism: full dynamic loss is strongest across most 3D cases and all OOD aggregate settings, solver teachers are more reliable than learned FM teachers in our comparisons, and the gains are largest for backbones whose temporal signal is weakest under standard SFT.
\paragraph{Limitations.}
One limiting area is that SaT by construction relies on access to reliable numerical solvers at training time. Its targets are solver-derived references. They are accurate only up to the governing equation and its discretization, and SaT requires a teacher whose dynamics remain valid at the states the student visits. A teacher coarsened until it diverges or changes the dynamics yields a correspondingly worse student: on KS2D, coarsening the teacher from $128^2$ to $80^2$, where it diverges after 36 steps, raises the student's cross-model RMSE from $0.090$ to $0.268$. We have not tested governing equations that are known to be incomplete or misspecified; pairing an inexpensive coarse solver with a learned closure is a natural direction. Also, our evaluation does not yet extend to complex geometries or unstructured meshes. Finally, SaT's efficiency is closely tied to successful solver parallelization. This poses an interesting direction for asynchronous training strategies.

Despite this, the presented results already demonstrate the promise of solver-guided, on-policy post-training for improving the robustness and long-horizon fidelity of PDE-FMs, and may serve as a key building block for future scientific foundation model pipelines.

\begin{ack}
The authors gratefully acknowledge the scientific support and HPC resources provided by the Erlangen National High Performance Computing Center (NHR@FAU) of the Friedrich-Alexander-Universität Erlangen-Nürnberg (FAU). We also acknowledge the EuroHPC Joint Undertaking for access to the supercomputer Leonardo, hosted by CINECA (Italy).
\end{ack}
\bibliography{reference}
\clearpage
\counterwithin{equation}{section}

\begin{center}
{\huge Appendix}
\end{center}

\appendix

\section{Training Method Computational Efficiency}
\label{sec:method_efficiency}

We profile the per-iteration computational cost of three training paradigms: Supervised Fine-Tuning (SFT), Solver-as-Teacher (SaT), and On-Policy Distillation with an FM teacher (OPD). All methods share the same student model (PDE-S, 33M parameters) and are evaluated on the KF2D task ($160^2$ resolution, batch size $B=4$, single NVIDIA L40S GPU). Each training step is decomposed into four timed components using asynchronous CUDA events: (1)~model forward (student rollout), (2)~solver/teacher (target generation), (3)~backward (gradient computation, including re-forward for DAgger), and (4)~optimizer (gradient clipping and AdamW update). We report the mean over 20 timed iterations after 5 warmup iterations.

\paragraph{SFT calibration.} For fair comparison, SFT unroll steps $S$ are set to match two reference points per SaT configuration: (i) the student rollout horizon $S=K$, and (ii) the total solver compositions $N_{\text{solver}}(K, m_{\max}) = \sum_{k=0}^{K-1}\min(m_{\max}, K-k)$. With $K \in \{1, 5\}$ and $m_{\max}=5$, this yields $S \in \{1, 5, 15\}$.

\paragraph{Gradient modes.} Both off-policy rollout training and on-policy training generate a $K$-step autoregressive student rollout $\hat{u}_0 \xrightarrow{\mathcal{G}_\theta} \hat{u}_1 \xrightarrow{\mathcal{G}_\theta} \cdots \xrightarrow{\mathcal{G}_\theta} \hat{u}_K$, and the gradient computation through this chain admits two modes. Because the present study uses rollout horizons up to $K=5$, BPTT remains feasible; however, DAgger/detach provides a more scalable option for larger $K$ or memory-constrained settings. We therefore benchmark both modes to quantify the time--memory trade-off, while leaving a broader comparison of long-horizon gradient estimators to prior work~\cite{list2025differentiability}.

\textbf{Backpropagation through time (BPTT).} The full rollout chain retains the computational graph, and a single backward call computes depth-$K$ Jacobian-vector products. This provides inter-step gradient signal---a parameter change at step~$0$ affects the loss at step~$K$---but activation memory scales as $O(K)$ because all intermediate states must be kept live.

\textbf{DAgger/detach.} Inspired by the DAgger framework for imitation learning~\cite{ross2011dagger}, this mode detaches the autoregressive chain: each rollout state is generated without gradient, then independently re-forwarded to compute depth-1 gradients via $K$ separate backward calls. The SFT analogue (``detach'' in Tables~\ref{tab:method_bptt}--\ref{tab:method_dagger}) applies the same per-step detach-and-reforward pattern to the supervised unroll.  Because no intermediate activations are retained across steps, activation memory is $O(1)$ regardless of the rollout horizon, at the cost of losing the inter-step gradient signal. This trade-off is primarily motivated by memory constraints when scaling to large foundation models or rolling out long horizons, as DAgger/detach keeps the memory footprint bounded and enables training configurations that would otherwise be infeasible.

\begin{table}[h]
\centering
\caption{Per-iteration component breakdown (ms) under BPTT gradient mode. SaT uses the Exponax pseudo-spectral solver; OPD uses PDE-L (518M params) as teacher. Measurements use $B=4$ at $160^2$ resolution on an NVIDIA L40S GPU.}
\label{tab:method_bptt}
\vspace{2pt}
\small
\begin{tabular}{@{}l rrrr r@{}}
\toprule
Method & Forward & Solver/Teacher & Backward & Optimizer & \textbf{Total} \\
\midrule
SFT ($S=1$)                  & 18.6 & ---   & 19.9  & 1.9 & \textbf{40.8}  \\
SFT ($S=5$)                  & 93.8 & ---   & 75.1  & 2.4 & \textbf{172.3} \\
SFT ($S=15$)                 & 270.2 & ---  & 228.5 & 2.8 & \textbf{503.9} \\
\midrule
SaT ($K=1$, $m=5$)        & 20.4 & 4.4   & 24.0  & 2.0 & \textbf{50.9}  \\
SaT ($K=5$, $m=5$)        & 90.9 & 41.5  & 75.3  & 2.3 & \textbf{210.1} \\
\midrule
OPD ($K=1$, $m=5$)        & 18.8 & 52.2  & 0.3   & 1.0 & \textbf{72.4}  \\
OPD ($K=5$, $m=5$)        & 92.0 & 708.5 & 2.3   & 1.5 & \textbf{804.5} \\
\bottomrule
\end{tabular}
\end{table}

\begin{table}[h]
\centering
\caption{Per-iteration component breakdown (ms) under DAgger/detach gradient mode. The backward column includes the re-forward passes (depth-1 each) that replace the BPTT chain. Same setup as~\cref{tab:method_bptt}.}
\label{tab:method_dagger}
\vspace{2pt}
\small
\begin{tabular}{@{}l rrrr r@{}}
\toprule
Method & Forward & Solver/Teacher & Backward & Optimizer & \textbf{Total} \\
\midrule
SFT ($S=1$, detach)          & 18.9  & ---   & 21.0  & 1.9 & \textbf{42.1}  \\
SFT ($S=5$, detach)          & 97.4  & ---   & 106.2 & 2.2 & \textbf{208.8} \\
SFT ($S=15$, detach)         & 296.8 & ---   & 321.2 & 2.2 & \textbf{630.3} \\
\midrule
SaT ($K=1$, DAgger)          & 17.8  & 4.5   & 42.9  & 1.9 & \textbf{67.1}  \\
SaT ($K=5$, DAgger)          & 77.8  & 40.3  & 202.0 & 2.0 & \textbf{322.1} \\
\midrule
OPD ($K=1$, DAgger)          & 17.0  & 51.6  & 17.0  & 0.9 & \textbf{86.6}  \\
OPD ($K=5$, DAgger)          & 74.1  & 712.8 & 91.7  & 1.3 & \textbf{879.9} \\
\bottomrule
\end{tabular}
\end{table}

\paragraph{SaT adds modest overhead; OPD is teacher-dominated.} At $K=5$, SaT (BPTT) requires 210\,ms per step --- the numerical solver accounts for only 20\% of the total cost (42\,ms). The remaining 80\% is the student's own forward and backward passes, which scale linearly with $K$. In contrast, OPD at $K=5$ requires 805\,ms, with the FM teacher consuming 88\% of the step cost (709\,ms). The teacher is approximately $17\times$ more expensive than the numerical solver at the same horizon, reflecting the 15.6$\times$ parameter ratio between PDE-L (518M) and the lightweight Exponax integrator.

\paragraph{Matched-horizon SFT comparison.} When $S$ is calibrated to the total solver compositions $N_{\text{solver}}$, SFT provides a compute-matched baseline. At $S=15$ (matching SaT $K=5$, $m_{\max}=5$), SFT (BPTT) takes 504\,ms --- $2.4\times$ more than SaT's 210\,ms for the equivalent number of model evaluations. This gap arises because SaT's solver branches (42\,ms) are substantially cheaper than the additional depth-15 backward pass through the full unroll chain that SFT requires.

\paragraph{BPTT vs.\ DAgger trade-off.} Table~\ref{tab:bptt_vs_dagger} compares the two gradient modes. DAgger is consistently slower due to (i)~$K$ separate backward calls vs.\ one, each incurring Python dispatch and gradient-accumulation overhead, and (ii)~additional depth-1 re-forward passes. The penalty is most pronounced for SaT ($+53\%$), where the re-forward cost dominates. For OPD the overhead is modest ($+9\%$) because the teacher's forward pass --- unchanged between modes --- accounts for the vast majority of step time. In practice, DAgger remains preferable for large $K$ due to its $O(1)$ activation memory, enabling training configurations that would otherwise exhaust GPU memory.

\begin{table}[h]
\centering
\caption{BPTT vs.\ DAgger per-step wall-clock time at $K=5$, $m_{\max}=5$ (or $S=15$ for SFT). DAgger trades wall-clock overhead for $O(1)$ activation memory. Same setup as Tables~\ref{tab:method_bptt}--\ref{tab:method_dagger}.}
\label{tab:bptt_vs_dagger}
\vspace{2pt}
\small
\begin{tabular}{@{}l rr r@{}}
\toprule
Config & BPTT (ms) & DAgger (ms) & Overhead \\
\midrule
SFT ($S=15$)          & 503.9 & 630.3 & $+25\%$ \\
SaT ($K=5$, $m=5$)  & 210.1 & 322.1 & $+53\%$ \\
OPD ($K=5$, $m=5$)  & 804.5 & 879.9 & $+9\%$  \\
\bottomrule
\end{tabular}
\end{table}

\subsection{End-to-End Training Wall-Clock Time}
\label{sec:appendix:wallclock}

The per-iteration breakdowns in Tables~\ref{tab:method_bptt}--\ref{tab:method_dagger} characterize \emph{compute cost per step}; here we report the aggregate wall-clock time observed for full training runs on the KS3D benchmark ($64^3$, single NVIDIA A100 GPU). Table~\ref{tab:wallclock_ks3d} lists the approximate total training time for each model--method pair, including both the SFT warm-up phase and the main training phase. SFT is the cheapest method for every backbone, finishing in ${\sim}1.4$--$2.6$\,h depending on the model. SaT variants add modest overhead: SaT-0, the best-performing variant, ranges from ${\sim}3.1$\,h (MORPH-Ti) to ${\sim}5.9$\,h (P3D-S), while SaT-1 and SaT-.5 fall in the same ballpark (${\sim}2.4$--$6.1$\,h). In contrast, OPD requires ${\sim}18.7$--$22.0$\,h per run for the student training alone, roughly $3.7$--$6.2{\times}$ the cost of SaT-0, because every on-policy step queries the frozen P3D-L teacher. Crucially, these OPD times exclude teacher preparation: the P3D-L teacher itself requires ${\sim}14.5$\,h of fine-tuning before it can serve as the OPD oracle, bringing the true end-to-end OPD cost to ${\sim}33$--$37$\,h, or about $6.2$--$10.9{\times}$ SaT-0.

\begin{table}[h]
\centering
\caption{End-to-end training wall-clock time for all model--method pairs on KS3D ($64^3$, single NVIDIA A100 GPU). All times are approximate.}
\label{tab:wallclock_ks3d}
\vspace{2pt}
\small
\begin{tabular}{@{}l l r@{}}
\toprule
Model & Method & Total Time \\
\midrule
DPOT-Ti  & SFT   & ${\sim}1.5$\,h \\
DPOT-Ti  & OPD   & ${\sim}18.7$\,h \\
DPOT-Ti  & SaT-1 & ${\sim}2.4$\,h \\
DPOT-Ti  & SaT-.5 & ${\sim}2.9$\,h \\
DPOT-Ti  & SaT-0 & ${\sim}3.6$\,h \\
\midrule
MORPH-Ti & SFT   & ${\sim}1.4$\,h \\
MORPH-Ti & OPD   & ${\sim}19.2$\,h \\
MORPH-Ti & SaT-1 & ${\sim}2.8$\,h \\
MORPH-Ti & SaT-.5 & ${\sim}3.0$\,h \\
MORPH-Ti & SaT-0 & ${\sim}3.1$\,h \\
\midrule
P3D-S    & SFT   & ${\sim}2.6$\,h \\
P3D-S    & OPD   & ${\sim}22.0$\,h \\
P3D-S    & SaT-1 & ${\sim}5.6$\,h \\
P3D-S    & SaT-.5 & ${\sim}6.1$\,h \\
P3D-S    & SaT-0 & ${\sim}5.9$\,h \\
\midrule
\multicolumn{3}{@{}l}{\textit{OPD teacher preparation}} \\
P3D-L    & -- & ${\sim}14.5$\,h \\
\bottomrule
\end{tabular}
\end{table}

\subsection{Accuracy at Equal Wall-Clock Time and Equal FLOPs}
\label{sec:appendix:budget}

To compare methods at matched training budgets, we retrained SFT, SaT-1, SaT-.5, SaT-0 and OPD from scratch on three tasks with two backbones each (DPOT-Ti and PDE-S on KS2D; DPOT-Ti and P3D-S on Burgers3D and KS3D). Each method uses its own recipe from \cref{tab:rmse_comparison}, with one seed, and all runs use NVIDIA A40 GPUs. Because these are separate runs, their absolute RMSE differs from \cref{tab:rmse_comparison}, so we report comparisons within each cell. Each entry gives the best method at that budget and its relative RMSE change: for SaT-0 and SaT-.5 relative to SaT-1, and for SaT-1 relative to SaT-0, so a positive value is an on-policy improvement. Each point uses the lowest test RMSE reached within the budget.

\begin{table}[h]\centering
\caption{Best method and relative RMSE change at equal wall-clock time. Positive values are improvements of the on-policy variant over SaT-1.}
\label{tab:wallclock_race}
\vspace{2pt}
\small
\begin{tabular}{@{}lcccc@{}}
\toprule
Cell & 1\,h & 3\,h & 6\,h & 9\,h \\
\midrule
Burgers3D / DPOT-Ti & SaT-0 $+3\%$ & SaT-0 $+27\%$ & SaT-0 $+28\%$ & SaT-0 $+28\%$ \\
Burgers3D / P3D-S & not converged & SaT-1 $-18\%$ & SaT-1 $-1\%$ & SaT-0 $+5\%$ \\
KS2D / DPOT-Ti & SaT-.5 $+39\%$ & SaT-.5 $+33\%$ & SaT-.5 $+33\%$ & SaT-.5 $+33\%$ \\
KS2D / PDE-S & SaT-0 $0\%$ & SaT-0 $+11\%$ & SaT-0 $+15\%$ & SaT-0 $+16\%$ \\
KS3D / DPOT-Ti & SaT-0 $+20\%$ & SaT-0 $+14\%$ & SaT-0 $+14\%$ & SaT-0 $+20\%$ \\
KS3D / P3D-S & not converged & SaT-.5 $+42\%$ & SaT-.5 $+26\%$ & SaT-.5 $+26\%$ \\
\bottomrule
\end{tabular}
\end{table}

An on-policy variant outperforms SaT-1 in five of six settings after three and six hours, and in all six after nine hours. When SaT-0 reaches the converged accuracy of SaT-1, it does so with a median $2.6\times$ less wall-clock time.

For FLOPs, we count the student's forward pass from the training graph and one solver macro-step from the compiled solver's XLA cost model. At $K=m=5$, a training step makes 15 teacher calls and costs 15 student forward-pass equivalents: five forward steps plus a backward pass of about ten. At $160^2$, a solver macro-step costs $0.11$\,GFLOP against $3.7$--$25$\,GFLOP for one student forward pass, an overhead below $3\%$. At $64^3$, the costs are $5.3$ and $13$\,GFLOP, and the overhead rises to $1.19\times$--$1.41\times$. \Cref{tab:flops_race} expresses budgets as fractions of the end-to-end cost of the SaT-1 recipe and uses the same convention as \cref{tab:wallclock_race}.

\begin{table}[h]\centering
\caption{Solver FLOPs overhead and best method at equal training FLOPs, as a fraction of the SaT-1 budget.}
\label{tab:flops_race}
\vspace{2pt}
\small
\begin{tabular}{@{}lccccc@{}}
\toprule
Cell & Overhead & $0.1\times$ & $0.25\times$ & $0.5\times$ & $1.0\times$ \\
\midrule
Burgers3D / DPOT-Ti & $1.41\times$ & SaT-1 $-7\%$ & SaT-0 $+19\%$ & SaT-0 $+28\%$ & SaT-0 $+28\%$ \\
Burgers3D / P3D-S & $1.19\times$ & not converged & SaT-0 $+64\%$ & SaT-0 $+4\%$ & SaT-0 $+0\%$ \\
KS3D / DPOT-Ti & $1.41\times$ & SaT-1 $-1\%$ & SaT-0 $+14\%$ & SaT-0 $+13\%$ & SaT-0 $+14\%$ \\
KS3D / P3D-S & $1.19\times$ & not converged & SaT-.5 $+54\%$ & SaT-.5 $+26\%$ & SaT-.5 $+26\%$ \\
\bottomrule
\end{tabular}
\end{table}

Charging the solver for every FLOP does not change the outcome: an on-policy variant leads in all four settings from a quarter of the SaT-1 budget onward, with a median improvement of $37\%$ at that budget and $20\%$ at the full budget.

\subsection{Inference Speed of the Surrogate and the Solver}
\label{sec:appendix:inference}

\Cref{tab:inference_speed} reports the time to advance one dataset frame at batch size 1 on one NVIDIA A40 in fp32 (median of 30 timed repetitions after 5 warm-up repetitions). The surrogate column is the fastest student for that task. Timing depends only on the architecture, so the students are timed with randomly initialized weights.

\begin{table}[h]\centering
\caption{Time per frame at batch size 1 (ms). The last column states which side is faster and by how much.}
\label{tab:inference_speed}
\vspace{2pt}
\small
\begin{tabular}{@{}lcrrl@{}}
\toprule
Task & Sub-steps & Solver & Surrogate & Faster \\
\midrule
KS2D ($160^2$) & 1 & 0.195 & 7.09 & solver, $36.4\times$ \\
Burgers2D ($160^2$) & 1 & 0.197 & 7.14 & solver, $36.3\times$ \\
KF2D ($160^2$) & 20 & 2.55 & 7.06 & solver, $2.8\times$ \\
KS3D ($64^3$) & 1 & 0.368 & 7.94 & solver, $21.5\times$ \\
Burgers3D ($64^3$) & 1 & 0.375 & 8.50 & solver, $22.7\times$ \\
KF3D ($64^3$) & 20 & 8.62 & 8.84 & tie, $1.0\times$ \\
DT3D ($64^3$) & 20 & 8.22 & 8.86 & tie, $1.1\times$ \\
KS3D ($96^3$) & 1 & 1.22 & 14.76 & solver, $12.1\times$ \\
KF3D ($96^3$) & 20 & 44.50 & 14.75 & surrogate, $3.0\times$ \\
TCF3D (PICT) & 1 & 23.34 & 8.94 & surrogate, $2.6\times$ \\
\bottomrule
\end{tabular}
\end{table}

For a lightweight pseudo-spectral solver with few sub-steps, the solver is faster. The surrogate becomes faster when a frame is expensive: many sub-steps on a fine grid (KF3D at $96^3$, $3.0\times$) or an implicit iterative solve (PICT, $2.6\times$).

\section{Solver--Model Bridge Efficiency}
\label{sec:bridge}

A key engineering challenge in the Solver-as-Teacher framework is the \emph{bridge} between the student FM (PyTorch) and the teacher solver (JAX/Exponax). During on-policy training, every rollout step requires transferring states from PyTorch to JAX for solver evaluation and transferring the resulting targets back. With multi-step lookahead depth $m_{\max}$, each training iteration involves $\mathcal{O}(K \cdot m_{\max})$ such round-trips, making bridge overhead a potential bottleneck. We implement and benchmark three bridge strategies:
\begin{enumerate}[leftmargin=*,itemsep=2pt]
  \item \textbf{DLPack zero-copy} (\texttt{gpu\_dlpack}): Both JAX and PyTorch allocate on the same GPU. Tensors are exchanged via the DLPack protocol (\texttt{jnp.from\_dlpack} / \texttt{torch.from\_dlpack}), which shares the underlying device pointer without any data movement.
  \item \textbf{CPU bridge} (\texttt{gpu\_cpu\_bridge}): The solver runs on GPU via JAX, but data is transferred through host memory: \texttt{Tensor.cpu().numpy()} $\to$ \texttt{jnp.array()} on the forward path and \texttt{np.array()} $\to$ \texttt{torch.from\_numpy().cuda()} on the return path. This incurs two GPU$\leftrightarrow$CPU copies per round-trip.
  \item \textbf{CPU solver} (\texttt{cpu\_solver}): JAX runs entirely on CPU (\texttt{JAX\_PLATFORMS=cpu}). This avoids GPU memory contention but is dramatically slower for spectral solvers, which are compute-heavy.
\end{enumerate}

\begin{figure}[h]
  \centering
  \includegraphics[width=\textwidth]{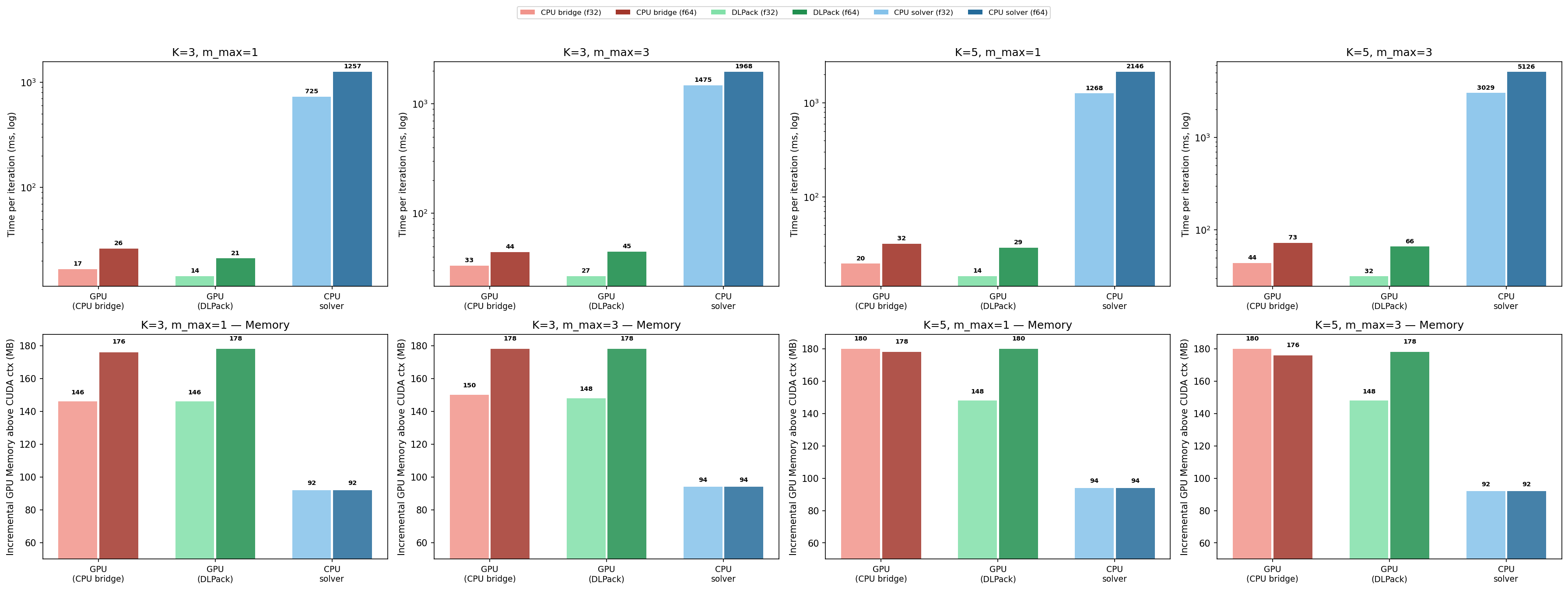}
  \caption{JAX$\leftrightarrow$PyTorch bridge benchmark. \textbf{Top row}: per-iteration wall-clock time (log scale) for three bridge strategies across four $(K, m_{\max})$ configurations, with \texttt{float32} (lighter) and \texttt{float64} (darker) bars. \textbf{Bottom row}: incremental GPU memory above the fixed CUDA context (427\,MB) for each mode. DLPack zero-copy consistently achieves the lowest latency while maintaining comparable memory footprint to the CPU bridge.}
  \label{fig:bridge}
\end{figure}

We benchmark the full depth-first multi-step branching loop, which mirrored the actual training pipeline, on Kolmogorov Flow ($160\!\times\!160$, $\mathrm{Re}\!\approx\!100$) with batch size $B\!=\!4$, rollout horizons $K\!\in\!\{3,5\}$, lookahead depths $m_{\max}\!\in\!\{1,3\}$, and solver precisions \texttt{float32} / \texttt{float64}. All timings are averaged over 50 iterations on an NVIDIA L40S GPU. The DLPack bridge reduces bridge overhead by $5\text{--}10\times$ compared to the CPU bridge, yielding $1.10$--$1.39\times$ end-to-end speedup across configurations. The CPU solver is $50$--$77\times$ slower than either GPU mode, confirming that GPU-resident spectral solvers are essential for practical training throughput. Incremental GPU memory above the fixed CUDA context (427\,MB) is $\sim\!148$\,MB (\texttt{float32}) / $\sim\!178$\,MB (\texttt{float64}) for DLPack, comparable to the CPU bridge, while the CPU solver uses only $\sim\!93$\,MB since JAX allocates on host. Figure~\ref{fig:bridge} visualizes these results.

\begin{table}[h]
\centering
\caption{Total iteration time (ms) and incremental GPU memory (above the fixed 427\,MB CUDA context) for each bridge mode and precision, reported as mean $\pm$ std across four $(K, m_{\max})$ configurations ($K\!\in\!\{3,5\}$, $m_{\max}\!\in\!\{1,3\}$). $B\!=\!4$, $160\!\times\!160$, NVIDIA L40S.}
\label{tab:bridge}
\vspace{2pt}
\small
\begin{tabular}{@{}l cc cc@{}}
\toprule
& \multicolumn{2}{c}{Iter Time (ms)} & \multicolumn{2}{c}{Incr.\ Memory (MB)} \\
\cmidrule(lr){2-3} \cmidrule(lr){4-5}
Mode & \texttt{float32} & \texttt{float64} & \texttt{float32} & \texttt{float64} \\
\midrule
DLPack zero-copy & $21.8 \pm 7.6$ & $40.2 \pm 17.3$ & $148 \pm 1$ & $178 \pm 1$ \\
CPU bridge       & $28.4 \pm 11.0$ & $43.8 \pm 17.9$ & $164 \pm 16$ & $177 \pm 1$ \\
CPU solver       & $1624.2 \pm 856.2$ & $2624.1 \pm 1482.0$ & $93 \pm 1$ & $93 \pm 1$ \\
\bottomrule
\end{tabular}
\end{table}

\section{Solver Backend Vectorization}
\label{sec:solver_backends}

The multi-step solver branching in SaT (Eq.~\ref{eq:branching}) requires $\sum_{k=0}^{K-1} m_k$ solver evaluations per training step, each applied to a mini-batch of $B$ samples. The total number of individual solver macro-steps is $B \times \sum m_k$, which grows as $O(B \cdot K \cdot m_{\max})$. For Kolmogorov Flow at $160\!\times\!160$ resolution with 20 ETDRK2 sub-steps per macro-step, the solver branching accounts for 80\%+ of per-iteration wall-clock time when executed naively. We implement and benchmark three solver backends that address two root bottlenecks: (1)~sequential per-sample solver calls (Python-level loop over the batch dimension) and (2)~suboptimal scheduling of the branch--depth computation graph.

The three backends represent a {\em no-vectorization baseline} and two vectorization strategies that differ in how the branch--depth computation graph is traversed:

\begin{enumerate}[leftmargin=*,itemsep=2pt]
  \item \textbf{No vectorization} --- iterates by depth level $j$; at each depth, all active branches are concatenated into a mega-batch, but each sample is solved individually via a sequential Python loop. This baseline isolates the raw solver cost without any hardware-level parallelism. Because both depth-first and branch-first traversal orders execute the same total number of single-sample solver calls ($B \times \sum m_k$), they yield identical wall-clock times; we therefore report a single baseline.
  \item \textbf{State vectorization} --- same depth-first scheduling, but the mega-batch at each depth level is processed in a single vectorized solver dispatch (via \texttt{jax.vmap}), eliminating the per-sample loop. At depth $j$, the batch size equals $B$ times the number of branches still active, maximizing GPU occupancy.
  \item \textbf{Solver vectorization} --- iterates by branch $k$; each branch's full chain $[\mathcal{S}^1(\hat{u}_k), \ldots, \mathcal{S}^{m_k}(\hat{u}_k)]$ is computed in a single fused dispatch (\texttt{jax.vmap} $+$ \texttt{jax.lax.scan}), vectorizing across all $B$ samples and chaining through $m_k$ steps without returning to the host framework. This issues $K$ sequential XLA dispatches (one per branch), each with batch size $B$.
\end{enumerate}

\paragraph{Setup.} We benchmark on 2D Kolmogorov Flow ($160\!\times\!160$, $\mathrm{Re}\!\approx\!100$) using the Exponax pseudo-spectral solver with 20 ETDRK2 sub-steps per macro-step. Batch size $B\!=\!4$, rollout horizon $K\!=\!10$, truncate strategy, NVIDIA L40S GPU. All timings are averaged over 10 iterations after 2 warmup iterations. Numerical equivalence (max absolute difference $= 0$) is verified across all backends.

\begin{table}[h]
\centering
\caption{Wall-clock time (s) and speedup relative to the no-vectorization baseline for three solver backends across two lookahead depths. $K\!=\!10$, $B\!=\!4$, $160\!\times\!160$, truncate strategy, NVIDIA L40S.}
\label{tab:solver_backends}
\vspace{2pt}
\small
\begin{tabular}{@{}l cc cc@{}}
\toprule
& \multicolumn{2}{c}{$m_{\max} = 5$ (40 steps)} & \multicolumn{2}{c}{$m_{\max} = 8$ (52 steps)} \\
\cmidrule(lr){2-3} \cmidrule(lr){4-5}
Backend & Time (s) & Speedup & Time (s) & Speedup \\
\midrule
No vectorization       & $0.464$ & $1.00\times$ & $0.596$ & $1.00\times$ \\
State vectorization   & $0.033$ & $14.20\times$ & $0.047$ & $12.74\times$ \\
Solver vectorization   & $0.102$ & $4.55\times$ & $0.126$ & $4.73\times$ \\
\bottomrule
\end{tabular}
\end{table}

\begin{figure}[h]
  \centering
  \includegraphics[width=\textwidth]{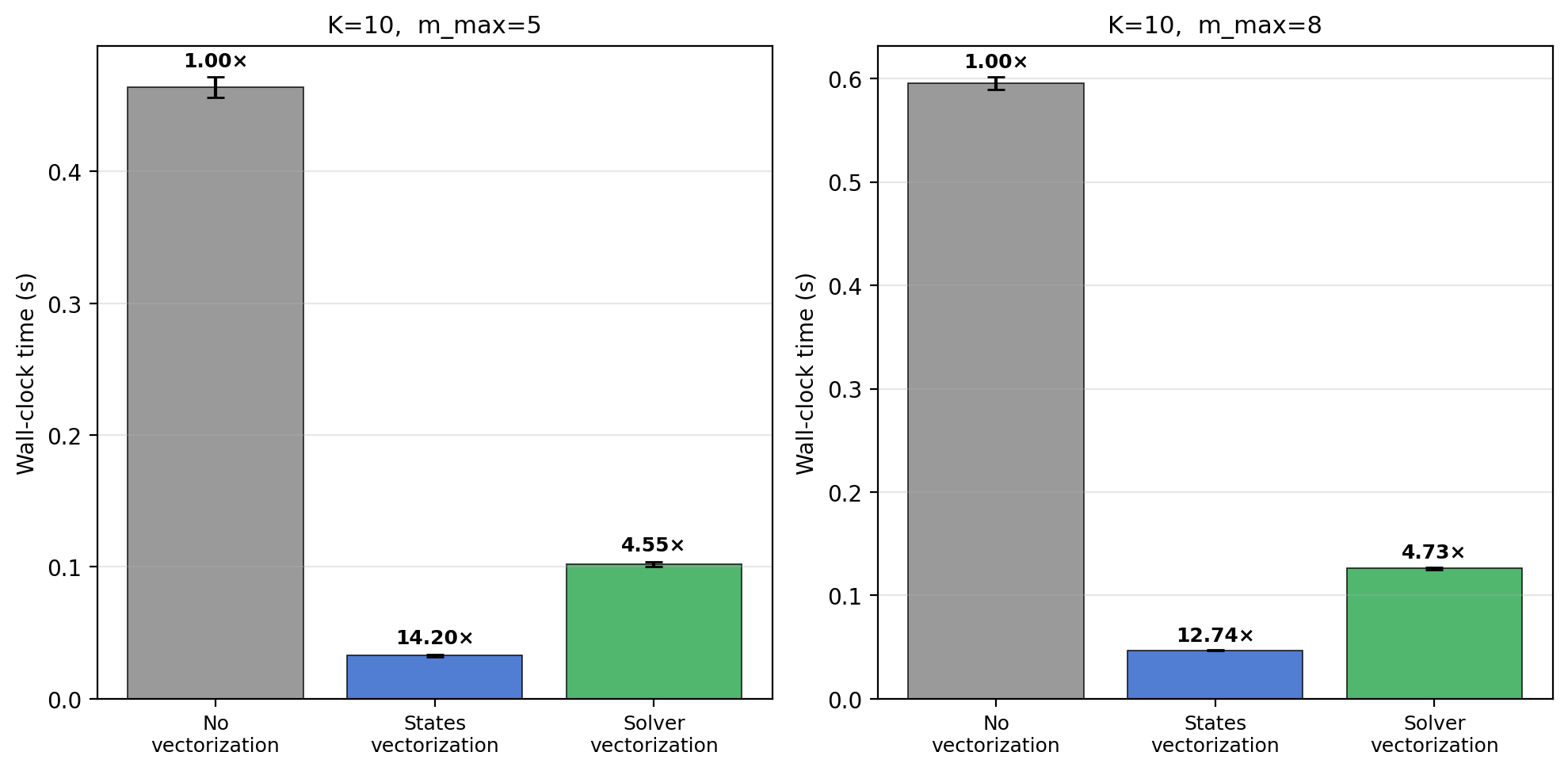}
  \caption{Wall-clock comparison of three solver backends on Kolmogorov Flow ($160\!\times\!160$, $B\!=\!4$, GPU). Speedup factors are annotated above each bar relative to the no-vectorization baseline. \emph{Left:} $m_{\max}\!=\!5$ (40 total solver macro-steps). \emph{Right:} $m_{\max}\!=\!8$ (52 total solver macro-steps). State vectorization consistently achieves the highest throughput by maximizing batch size per GPU kernel launch.}
  \label{fig:solver_backends}
\end{figure}

\paragraph{Analysis.} State vectorization provides the dominant speedup, achieving $14.2\times$ at $m_{\max}\!=\!5$ and $12.7\times$ at $m_{\max}\!=\!8$ over the sequential baseline. Its advantage stems from maximizing the batch size per GPU kernel: at depth $j\!=\!0$, all $K$ branches are active, yielding a mega-batch of $K \times B = 40$ samples in a single \texttt{vmap} dispatch. As branches terminate at deeper levels, the batch size decreases gracefully.

Solver vectorization achieves a more modest $4.6$--$4.7\times$ speedup. Although it eliminates inter-step bridge overhead by fusing each branch's chain into a single XLA program, each dispatch processes only $B\!=\!4$ samples---an order of magnitude fewer than the state-vectorization mega-batch. The $K\!=\!10$ sequential dispatches (one per branch) also incur more total launch overhead than the $m_{\max}$ dispatches of state vectorization.

In practice, we adopt state vectorization as the default backend due to its consistently superior throughput and minimal implementation complexity. Its only requirement is that the solver supports batched evaluation, which is satisfied by any \texttt{jax.vmap}-compatible stepper.

\section{Model Architecture Details}
\label{sec:appendix:model_details}

We provide detailed descriptions of the six PDE foundation models evaluated in our experiments. Table~\ref{tab:model_summary} summarizes the architecture family, spatial support, core mechanism, pretraining strategy, fine-tuning setting, and parameter count for each model. All models are pretrained on PDE corpora and post-trained within our unified pipeline.

\begin{table}[h]
\centering
\captionsetup{width=\textwidth}
\caption{Summary of the six PDE foundation models. \textbf{Params} reports the total parameter count of the instantiated wrapper used in our experiments. For DPOT-Ti, we list both the 2D and 3D parameter counts; the difference arises from the 3D patch embedding, positional embedding, and output reconstruction layers. All 2D models operate at $160^2$ resolution; all 3D models at $64^3$.}
\label{tab:model_summary}
\vspace{2pt}
\small
\resizebox{\textwidth}{!}{\begin{tabular}{@{}l l l l l r@{}}
\toprule
Model & Spatial & Architecture Core & Pretraining & Tuning & Params \\
\midrule
PDE-S~\cite{holzschuh2025pdetransformer} & 2D & Swin U-Net Transformer & 16 PDEs & Full & 33.2M \\
MPP-B~\cite{mccabe2024mpp} & 2D & Axial ViT (AViT) & Multi-physics & Full & 115.6M \\
DPOT-Ti~\cite{hao2024dpot} & 2D / 3D & AFNO spectral blocks & AR denoising & Full & 5.2M / 12.7M \\
Poseidon-T~\cite{herde2024poseidon} & 2D & SwinV2 U-Net (ScOT) & All-to-all semigroup & Full & 20.7M \\
P3D-S~\cite{holzschuh2025p3d} & 3D & Conv3D + windowed 3D attn. & 14 3D PDEs & Full & 11.2M \\
MORPH-Ti~\cite{rautela2025morph} & 3D & 4D axial ViT + cross-attn. & Heterogeneous 1D--3D & Full & 9.9M \\
\bottomrule
\end{tabular}}
\end{table}

\paragraph{PDE-S.}
PDE-Transformer~\cite{holzschuh2025pdetransformer} is a Swin-like U-Net Transformer. The architecture features patch-based input tokenization with patch size 4, a hierarchical encoder-decoder with depths $[2, 5] \to [8] \to [5, 2]$ and embed dimension 96, shifted-window attention with periodic padding support, and AdaLN-Zero conditioning on diffusion timestep and PDE class labels. The PDE-S variant uses 4 attention heads and operates at $160^2$ resolution. We use the deterministic prediction mode (timestep forced to zero), and the PDE class index is set according to the target task.

\paragraph{MPP-B (Multiple Physics Pretraining).}
MPP~\cite{mccabe2024mpp} employs an Axial Vision Transformer (AViT) with interleaved spatial axial attention (along H and W axes independently) and temporal attention blocks operating along the time axis, each with relative positional biases. The model uses sparse per-physics input/output projections conditioned on state labels, integer indices into a pretrained vocabulary of physical quantities (e.g., velocity Vx/Vy). MPP-B has embed dimension 768, 12 attention heads, 12 processor blocks, and patch size $[16, 16]$. It was pretrained on 12 physical systems spanning fluid dynamics, electromagnetics, and solid mechanics.

\paragraph{DPOT-Ti (Denoising Operator Transformer).}
DPOT~\cite{hao2024dpot} is built on the Adaptive Fourier Neural Operator (AFNO) architecture. Each of its $d=4$ spectral mixing blocks applies a learnable two-layer complex-valued MLP in the truncated Fourier domain, retaining only the lowest $k=32$ frequency modes in 2D and $k=8$ modes in our 3D configuration. This spectral filtering provides a global receptive field through FFT-based mixing. The architecture comprises: (1) a convolutional patch embed layer (patch size 8) that projects the input augmented with spatiotemporal coordinate channels into a latent space of dimension 512; (2) a time aggregator with learnable exponential-cosine temporal encoding; (3) a residual AFNO block stack with $1\!\times\!1$ convolutional feed-forward networks; and (4) a transposed-convolutional output layer that reconstructs the prediction at the original spatial resolution.

DPOT is the only model applied to both 2D and 3D tasks. The 2D variant (\texttt{DPOTNet}, $160^2$) uses Conv2d operations with patch kernel size $8^2 = 64$ and contains 5.2M parameters. The 3D variant (\texttt{DPOTNet3D}, $64^3$) keeps the same DPOT-Ti width, depth, and AFNO block count, but replaces the patch embedder, positional embedding, and output reconstruction with 3D counterparts. This increases the instantiated parameter count to 12.7M because the learned patch/output kernels scale from $8^2$ to $8^3$. Crucially, the AFNO spectral parameters ($w_1, w_2, b_1, b_2$) are shape-invariant between 2D and 3D and are directly transferred from the pretrained 2D checkpoint; the $1\!\times\!1$ block MLP weights are unsqueezed from Conv2d to Conv3d, while the 3D patch embedding, positional embedding, and output layer are randomly initialized and fine-tuned.

\paragraph{Poseidon-T.}
Poseidon~\cite{herde2024poseidon} is built on the ScOT (Scalable Operator Transformer) architecture, adapting SwinV2 for PDE operator learning. It features 4 encoder-decoder stages with depths $[4, 4, 4, 4]$, embed dimension 48, attention heads $[3, 6, 12, 24]$, window size 16, and patch size 4. Poseidon was pretrained with an all-to-all semigroup strategy that forms supervised pairs across different lead times within each trajectory, not with autoregressive pushforward unrolling. In our wrapper, time conditioning is disabled (\texttt{use\_conditioning=false}, \texttt{time=None}) to match the single-step pipeline. Poseidon includes built-in spectral (FFT-based) resolution adaptation: if the input resolution ($160^2$) differs from the pretrained image size ($128^2$), the model internally resamples via spectral truncation/padding. Since the released checkpoint uses 4 input/output channels whereas our 2D tasks use 2 effective channels, the embedding and recovery layers are reinitialized and trained together with the full model weights.

\paragraph{P3D-S (Physics-3D).}
P3D~\cite{holzschuh2025p3d} is a hybrid CNN-Transformer architecture for scalable 3D PDE surrogate modelling. The model combines: (1) a conditioned 3D encoder with stride-2 Conv3D downsampling, AdaLN conditioning, and skip connections; (2) a \texttt{P3DBackbone} with windowed 3D attention using SwinV2-style relative positional biases, window size 4, and encoder/decoder stages connected by PixelShuffle3D up/downsampling; and (3) a conditioned 3D decoder that upsamples back to original resolution. P3D-S has a hidden size of 64, 5 stages with depth $[2,2,2,2,2]$, 4 heads per stage, and 2 convolutional downsampling layers. The pretrained model operates deterministically (timestep and class conditioning zeroed). P3D requires a minimum input resolution of $64^3$ as its latent must be $\geq$ window size; for smaller inputs, the wrapper applies circular padding.

\paragraph{MORPH-Ti.}
MORPH~\cite{rautela2025morph} is a modality-agnostic PDE foundation model with a convolutional Vision Transformer architecture. It processes volumetric fields through: (1) a 3D hybrid patch embedding with Conv3D and inter-field cross-attention (32 heads); (2) spatio-temporal positional encoding; (3) $d=4$ encoder blocks with \emph{4D axial attention} decomposed along the temporal, depth, height, and width axes through four axial attention modules; and (4) a decoder that maps each patch token back to physical space via a linear projection. MORPH-Ti has latent dimension 256, 4 attention heads per axial module, MLP dimension 1024, and patch size 8. To mitigate blocky artefacts from the per-patch independent decoder, we added a lightweight residual Conv3D refinement head (zero-initialised) and enforced a minimum of 8 patches per spatial axis via circular padding.

\paragraph{Pipeline adaptations.}
Three models with native multi-timestep temporal input (DPOT, MPP, MORPH) are operated in single-frame mode ($T=1$) to match our pipeline's single-step prediction interface. For DPOT, the pretrained $T=10$ time aggregator is replaced by a $T=1$ aggregator and re-learned during post-training. For MPP and MORPH, the temporal attention modules remain loaded from the pretrained checkpoint but operate over a singleton temporal axis; in MORPH, inter-field cross-attention also becomes singleton because the wrapper uses $F=1$. The remaining three models (PDE-Transformer, Poseidon, P3D) natively process single spatial frames and require no temporal modification, although Poseidon's lead-time conditioning is explicitly disabled in our setup. Channel mismatches between pretrained weights and target tasks are handled differently per model: DPOT and Poseidon use random reinitialisation of embedding/recovery layers; P3D slices pretrained weights along the channel dimension to preserve learned features; MPP avoids channel mismatches through its sparse projections conditioned on state labels. Resolution mismatches are handled via bicubic interpolation of positional embeddings (DPOT-2D), built-in spectral resampling (Poseidon), or circular padding to meet minimum resolution requirements (P3D, MORPH).

\subsection{Configuration of Experiments}
\label{sec:appendix:finetuning}

All post-training runs are launched from the same composed configuration system. The training protocol, grid-search axes, and run-selection rule are summarized in~\cref{sec:appendix:train_strategy}; the present section focuses on the model-specific architecture and wrapper details needed to instantiate each pretrained backbone.

\section{Experimental Details}
\label{sec:appendix:experiment}

We describe the governing equations, solver methodology, and dataset generation parameters for all seven PDE benchmark tasks. All PDEs are solved on periodic domains using the Exponax library~\cite{koehler2024apebench}, which implements Exponential Time Differencing Runge--Kutta (ETDRK) pseudo-spectral methods in JAX. Initial conditions are random truncated Fourier series with wavenumber cutoff $k_{\max}=5$.

\subsection{Solver Methodology}
\label{sec:appendix:solver}

Exponax solves semi-linear PDEs of the form $\partial_t u = \hat{\mathcal{L}} u + \hat{\mathcal{N}}(u)$, where $\hat{\mathcal{L}}$ is a linear operator and handled exactly in Fourier space via the matrix exponential, while $\hat{\mathcal{N}}$ is a nonlinear term. Spatial derivatives are computed pseudo-spectrally using FFTs with $2/3$-rule dealiasing. For stiff PDEs requiring fine temporal resolution, each macro time step $\Delta t_{\mathrm{FM}}$ is subdivided into $N_{\mathrm{sub}}$ internal ETDRK2 sub-steps of size $\delta t = \Delta t_{\mathrm{FM}} / N_{\mathrm{sub}}$, with the macro step realized as: $\mathcal{S}_{\Delta t} = \underbrace{\mathcal{S}_{\delta t} \circ \cdots \circ \mathcal{S}_{\delta t}}_{N_{\mathrm{sub}}}$.

\subsection{Details of system equations}
\label{sec:appendix:difficulty}

The Kuramoto--Sivashinsky and Burgers equations are specified through APEBench's \emph{difficulty parameterization}~\cite{koehler2024apebench}. The underlying Exponax stepper solves
\begin{equation}
    \partial_t u \;=\; \sum_{j=0}^{J} a_j \,(\mathbf{1}\cdot\nabla^{j})\, u \;+\; b_0\, u^2 \;+\; b_1\,\tfrac{1}{2}(\mathbf{1}\cdot\nabla)(u^2) \;+\; b_2\,\tfrac{1}{2}\|\nabla u\|_2^2,
    \label{eq:difficulty}
\end{equation}
where $(\mathbf{1}\cdot\nabla^{j})\,u = \sum_{i=1}^{d} \partial_{x_i}^{j} u$, so the linear operators contain no cross-derivatives. On an $N^d$ grid of a periodic domain with side $L$ and time step $\Delta t$, the difficulties are dimensionless versions of these coefficients:
\begin{equation}
    \gamma_j = \frac{a_j\,\Delta t}{L^{j}}\,N^{j}\,2^{j-1}D \;\;(j \geq 1), \qquad
    \delta_1 = \frac{b_1\,\Delta t}{L}\,M N D, \qquad
    \delta_2 = \frac{b_2\,\Delta t}{L^{2}}\,M N^{2} D,
    \label{eq:difficulty_def}
\end{equation}
with $\gamma_0 = a_0\Delta t$, $\delta_0 = b_0\Delta t$, $D$ the number of spatial dimensions, and $M=1$ the expected maximum absolute value of the state. The stepper runs on $(0,1)^d$ with a unit time step, so $\boldsymbol{\gamma}$ and $\boldsymbol{\delta} = (\delta_0, \delta_1, \delta_2)$ fully determine the dynamics relative to the grid. Because the difficulties include $N$ and $D$, the same values describe different continuum problems at $160^2$ and at $64^3$.

The Navier--Stokes benchmarks use Exponax's physics-based stepper classes, which directly parameterize the governing equations through physical quantities (viscosity, Reynolds number, forcing parameters) rather than abstract difficulty coefficients.

\paragraph{Kuramoto--Sivashinsky (KS2D, KS3D).}
The KS equation models spatiotemporal chaos arising from the interplay of anti-diffusion (energy production at intermediate scales) and hyper-diffusion (energy dissipation at small scales):
\begin{equation}
    \partial_t u + \tfrac{b}{2}\|\nabla u\|_2^2 = -\nu\,\Delta u - \zeta \sum_{i=1}^{d} \partial_{x_i}^{4} u,
    \label{eq:ks}
\end{equation}
with $d \in \{2, 3\}$ and coefficients set by the difficulties through \cref{eq:difficulty_def}. We use the flame-front (gradient-norm) form of the nonlinearity rather than the convective form $u \cdot \nabla u$; the two coincide in 1D under $u = \partial_x \phi$ but are distinct operators in higher dimensions, and Exponax's difficulty interface parameterizes the gradient-norm term. In the difficulty parameterization, this corresponds to $\boldsymbol{\gamma} = (0, 0, -1.2, 0, -15)$ and $\boldsymbol{\delta} = (0, 0, -6)$, where $\gamma_2 = -1.2$ provides anti-diffusion, $\gamma_4 = -15$ provides hyper-diffusion, and $\delta_2 = -6$ scales the gradient-norm nonlinearity. Up to a rescaling of $u$, these values correspond to the classical KS equation ($b=\nu=\zeta=1$) on a periodic box of side $\ell \approx 0.566\,N$, i.e., $\ell \approx 90.5$ for KS2D ($N=160$) and $\ell \approx 36.2$ for KS3D ($N=64$); on $[0,2\pi]^d$ the classical equation has no linearly unstable mode.

\paragraph{Burgers (Burgers2D, Burgers3D).}
The single-channel Burgers equation combines viscous diffusion with a convective nonlinearity acting along the diagonal direction $\mathbf{1}$:
\begin{equation}
    \partial_t u + \tfrac{1}{2}(\mathbf{1}\cdot\nabla)(u^2) = \nu\,\Delta u,
    \label{eq:burgers}
\end{equation}
where $\tfrac{1}{2}(\mathbf{1}\cdot\nabla)(u^2) = u\sum_{i=1}^{d}\partial_{x_i} u$. The difficulty parameterization is $\boldsymbol{\gamma} = (0, 0, 1.5, 0, 0)$ and $\boldsymbol{\delta} = (0, -1.5, 0)$, where $\gamma_2 = 1.5$ provides forward diffusion and $\delta_1 = -1.5$ encodes the convective nonlinearity. By \cref{eq:difficulty_def}, the Reynolds number based on the domain size and the maximum initial amplitude is $2N|\delta_1|/\gamma_2 = 2N$, i.e., 320 for Burgers2D and 128 for Burgers3D.

\paragraph{Kolmogorov Flow --- Vorticity Formulation (KF2D).}
Kolmogorov flow solves the 2D incompressible Navier--Stokes equations in vorticity formulation with a monochromatic body force. The vorticity $\omega = \nabla \times \mathbf{v}$ evolves as:
\begin{equation}
    \partial_t \omega + (\mathbf{v} \cdot \nabla)\omega = \nu \nabla^2 \omega + \mu \omega + f, \qquad f = -\tfrac{2\pi n_f}{L}\,A \cos\!\left(\tfrac{2\pi n_f}{L} y\right),
    \label{eq:kolm2d}
\end{equation}
where $\nu$ is the kinematic viscosity, $\mu < 0$ is a linear drag coefficient that removes energy at the largest scales, $A$ is the forcing amplitude, and $n_f$ is the forcing wavenumber (injection mode). The forcing $f$ is the curl of the Kolmogorov body force $A \sin(2\pi n_f y / L)\,\mathbf{e}_x$ acting on the primary variables, which is why it enters the vorticity equation as a cosine carrying the extra factor $2\pi n_f / L$. The velocity field $\mathbf{v}$ is recovered from the vorticity via the Biot--Savart law in Fourier space. The KF2D task uses $\nu = 0.01$ ($\mathrm{Re} \approx 100$), $\mu = -0.2$, forcing mode $n_f = 4$, forcing amplitude $A = 1.0$, and $L = 2\pi$, so $f = -4\cos(4y)$. The APEBench data were generated with Exponax 0.1.0 and drag parameter $-0.1$, which that version applies with a factor of $D$, giving $\mu = -0.2$ in 2D.

\paragraph{Kolmogorov Flow --- Velocity Formulation (KF3D).}
The 3D Kolmogorov flow is solved directly in velocity formulation, evolving the incompressible velocity field $\mathbf{v} = (v_x, v_y, v_z)$:
\begin{equation}
    \partial_t \mathbf{v} + (\mathbf{v} \cdot \nabla)\mathbf{v} = -\nabla p + \nu \nabla^2 \mathbf{v} + \mu\,\mathbf{v} + \mathbf{f}, \qquad \nabla \cdot \mathbf{v} = 0,
    \label{eq:kolm3d}
\end{equation}
where $p$ is the pressure enforcing incompressibility (projected out in Fourier space via the Leray projector), and $\mathbf{f} = A \sin(2\pi n_f y / L)\,\mathbf{e}_x$ is a sinusoidal body force. The KF3D task uses $\nu = 0.01$, $\mu = -0.1$, $n_f = 4$, $A = 1$, and $L = 2\pi$. The velocity formulation produces 3 output channels ($v_x, v_y, v_z$), and initial conditions are projected to be divergence-free via \texttt{ex.spectral.make\_incompressible}. A 200-step warm-up period drives the flow into a statistically stationary turbulent regime before data recording begins.

\paragraph{Decaying Turbulence --- Velocity Formulation (DT3D).}
Decaying turbulence solves the unforced 3D incompressible Navier--Stokes equations:
\begin{equation}
    \partial_t \mathbf{v} + (\mathbf{v} \cdot \nabla)\mathbf{v} = -\nabla p + \nu \nabla^2 \mathbf{v}, \qquad \nabla \cdot \mathbf{v} = 0,
    \label{eq:decay3d}
\end{equation}
without body force or linear drag. Energy injected through the initial condition cascades from large to small scales and is dissipated by viscosity, resulting in a monotonically decaying energy spectrum. This task tests the model's ability to capture the temporal evolution of spectral energy transfer without the stationarity provided by external forcing.

\subsection{Dataset Generation and Solver Parameters}
\label{sec:appendix:dataset}

All 2D datasets are sourced from the original APEBench benchmark~\citep{koehler2024apebench}, while the 3D datasets are generated at higher resolution ($64^3$) by directly instantiating the corresponding Exponax stepper. Table~\ref{tab:pde_params} summarizes the complete solver parameters for all seven benchmark tasks.

\begin{table}[h]
\centering
\caption{Solver parameters for all seven PDE benchmark tasks. \textbf{Top}: 2D tasks at $160^2$ resolution. \textbf{Bottom}: 3D tasks at $64^3$ resolution. $\Delta t_{\mathrm{FM}}$ is the macro time step per FM prediction, $N_{\mathrm{sub}}$ is the number of ETDRK2 sub-steps per macro step, and $C$ is the number of data channels.}
\label{tab:pde_params}
\vspace{2pt}
\small
\begin{tabular}{@{}l l c c c c l@{}}
\toprule
Task & PDE & $C$ & Resolution & $\Delta t_{\mathrm{FM}}$ & $N_{\mathrm{sub}}$ & \begin{tabular}[t]{@{}l@{}}Key \\ Parameters\end{tabular} \\
\midrule
KS2D      & Kuramoto--Sivashinsky & 1 & $160^2$ & $1.0$ & 1 & \begin{tabular}[t]{@{}l@{}}$\boldsymbol{\gamma}\!=\!(0,0,-1.2,0,-15)$; \\ $\boldsymbol{\delta}\!=\!(0,0,-6)$\end{tabular} \\
Burgers2D & Burgers (single-ch.)  & 1 & $160^2$ & $1.0$ & 1 & \begin{tabular}[t]{@{}l@{}}$\boldsymbol{\gamma}\!=\!(0,0,1.5,0,0)$; \\ $\boldsymbol{\delta}\!=\!(0,-1.5,0)$\end{tabular} \\
KF2D      & Kolmogorov Flow (vort.) & 1 & $160^2$ & $0.1$ & 20 & \begin{tabular}[t]{@{}l@{}}$\nu\!=\!0.01$, drag $\!=\!-0.1$, \\ mode $\!=\!4$, $L\!=\!2\pi$ \\ \end{tabular} \\
\midrule
KS3D      & Kuramoto--Sivashinsky & 1 & $64^3$  & $1.0$ & 1 & \begin{tabular}[t]{@{}l@{}}$\boldsymbol{\gamma}\!=\!(0,0,-1.2,0,-15)$; \\ $\boldsymbol{\delta}\!=\!(0,0,-6)$\end{tabular} \\
Burgers3D & Burgers (single-ch.)  & 1 & $64^3$  & $1.0$ & 1 & \begin{tabular}[t]{@{}l@{}}$\boldsymbol{\gamma}\!=\!(0,0,1.5,0,0)$; \\ $\boldsymbol{\delta}\!=\!(0,-1.5,0)$\end{tabular} \\
KF3D      & Kolmogorov Flow (vel.) & 3 & $64^3$  & $0.1$ & 20 & \begin{tabular}[t]{@{}l@{}}$\nu\!=\!0.01$, drag $\!=\!-0.1$, \\ mode $\!=\!4$, $L\!=\!2\pi$\end{tabular} \\
DT3D      & Decaying Turb.\ (vel.)    & 3 & $64^3$  & $0.1$ & 20 & \begin{tabular}[t]{@{}l@{}}$\nu\!=\!0.01$, drag $\!=\!0$, \\ $L\!=\!2\pi$\end{tabular} \\
\bottomrule
\end{tabular}
\end{table}

\paragraph{Training data and splits.}
Each in-distribution task uses 50 training trajectories of 50 time steps and a held-out test set of 200-step trajectories, all generated from independent random initial conditions drawn from a truncated Fourier series with wavenumber cutoff $k_{\max}=5$. All 50 training trajectories are used for training. During post-training we monitor the rollout RMSE on the held-out trajectories at horizons 10 and 30: the plateau scheduler uses the 30-step value, and the checkpoint with the lowest 10-step value is retained. All methods use the same protocol. Reported metrics, in distribution and OOD, are 50-step autoregressive rollouts from the first five test trajectories.

\paragraph{Warm-up protocol.}
For the Kolmogorov flow tasks, the sinusoidal body forcing must first drive the system into a statistically stationary turbulent regime; we therefore discard a warm-up transient before recording data: 500 steps for KF2D and 200 steps for KF3D. Similarly, for Kuramoto--Sivashinsky (KS2D, KS3D), a 500-step warm-up develops the characteristic spatiotemporal chaos, as the energy-producing anti-diffusion requires time to populate the chaotic attractor from the smooth random Fourier initial conditions. Both Burgers tasks and decaying turbulence (DT3D) use no warm-up: Burgers' positive diffusion immediately steepens the initial condition into shock fronts, so the transient formation is itself the target dynamics; decaying turbulence captures the energy cascade from the initial vortex field, which is likewise the phenomenon of interest. These conventions follow the APEBench defaults; the KF3D warm-up is our own choice, since APEBench has no 3D Kolmogorov scenario.

\subsection{Out-of-Distribution Zero-Shot Datasets}
\label{sec:appendix:zeroshot}

To assess generalization beyond the training distribution, we evaluate post-trained models on out-of-distribution (OOD) test sets generated by perturbing the physical parameters of the in-distribution tasks. These datasets are test-only (10 trajectories, 200 steps each) and share the same spatial resolution and domain as their in-distribution counterparts. The OOD configurations are generated via the same Exponax solver infrastructure. The lower viscosity variant ($\nu = 0.005$) requires doubled sub-stepping ($N_{\mathrm{sub}} = 40$) to maintain solver stability. Table~\ref{tab:ood_params} summarizes the parameter perturbations for each OOD task.

\begin{table}[h]
\centering
\caption{Out-of-distribution (OOD) zero-shot evaluation datasets. Each row lists the parameter changes relative to the in-distribution training task.}
\label{tab:ood_params}
\vspace{2pt}
\small
\begin{tabular}{@{}l l l@{}}
\toprule
OOD Task & Base Task & Parameter Changes \\
\midrule
\multicolumn{3}{@{}l}{\textit{2D tasks ($160^2$)}} \\
Burgers2D-OOD & Burgers2D & $\gamma_2\!: 1.5 \to 0.5$; $\delta_1\!: -1.5 \to -0.5$ \\
KF2D-OOD      & KF2D      & $\nu\!: 0.01 \to 0.005$ ($N_{\mathrm{sub}}\!: 20 \to 40$); linear drag halved \\
KS2D-OOD      & KS2D      & $\delta_2\!: -6 \to -8$ \\
\midrule
\multicolumn{3}{@{}l}{\textit{3D tasks ($64^3$)}} \\
Burgers3D-OOD & Burgers3D & $\delta_1\!: -1.5 \to -0.5$ \\
KF3D-OOD      & KF3D      & $\nu\!: 0.01 \to 0.005$ ($N_{\mathrm{sub}}\!: 20 \to 40$)  \\
KS3D-OOD      & KS3D      & $\delta_2\!: -6 \to -8$  \\
DT3D-OOD      & DT3D      & $\nu\!: 0.01 \to 0.02$  \\
\bottomrule
\end{tabular}
\end{table}

\subsection{Training strategy}
\label{sec:appendix:train_strategy}

To ensure a rigorous empirical comparison, we employ a systematic hyperparameter sweep rather than relying on a single, hand-tuned configuration. For each available model--task--method pair, we perform a grid search over a defined hyperparameter space.
For each combination we then report the best-performing run of this search. This strict isolation guarantees that each method is evaluated at its peak capacity for a specific model and task, preventing the bias introduced by transferring suboptimal hyperparameters across disparate architectures.

Every run follows the same two-stage structure on an NVIDIA A100 GPU. First, an embedded SFT warm-up adapts the pretrained backbone to the downstream PDE distribution using single-step supervision ($K=1$), with 80 warm-up epochs for 2D runs and 40 warm-up epochs for 3D runs. Second, the main phase applies the method-specific objective: SFT trains against reference trajectories, PITA trains with its physics-informed residual objective, OPD uses the same on-policy rollout loop as SaT but replaces the numerical solver with a larger FM teacher, and SaT trains with the unified solver/anchor objective in~\cref{eq:unified-loss}. All methods share the same data split, trainer, checkpointing logic, rollout evaluation code, AdamW optimizer, gradient clipping at $\|\nabla\|_{\max}=1.0$, and composite per-step loss with MSE weight $1.0$ and Sobolev $H^1$ weight $0.5$ when such supervised targets are used.

\begin{table}[h]
\centering
\caption{Training grid used for model--task--method selection. The table lists the search axes used by the grid scripts and the realized ranges recorded in the complete run summary; individual retained configurations can be recovered from \texttt{experiment\_details\_2d.md}.}
\label{tab:training_strategy_grid}
\vspace{2pt}
\small
\begin{tabularx}{\textwidth}{@{}l X@{}}
\toprule
Axis & Values / protocol \\
\midrule
Methods & SFT, PITA (2D only), OPD, SaT-1 ($\alpha=1$), SaT-.5 ($\alpha=0.5$), SaT-0 ($\alpha=0$) \\
Optimization & AdamW; full fine-tuning in the standard grid; bf16 mixed precision \\
Learning rate & $10^{-3}$ \\
Epoch & 160 for 3D and 320 for 2D \\
Scheduler & Plateau or cosine scheduling, with 5\% warm-up and plateau factor $0.5$ when plateau is used \\
Batching & Base batch size is 6 for 3D and 12 for 2D; combined with gradient accumulation in $\{1,2,4,8\}$ \\
On-policy branching & SaT/OPD use $m_{\max}=5$ multi-step lookahead; $\lambda$ and $\gamma$ are set to 0.9 \\
\bottomrule
\end{tabularx}
\end{table}

OPD requires an additional teacher-preparation step. Since $K_{\max}=5$, we first adapt the larger full-size teacher backbone with the same two-stage strategy used for the students: a single-step SFT warm-up, followed by an unrolled SFT main phase whose rollout horizon increases linearly from $K=1$ to $K_{\max}=5$. This gives OPD a task-specialized teacher that has already been trained for the same autoregressive horizon used in the student grid, avoiding a comparison against an unadapted large pretrained model. During the subsequent OPD run, this teacher checkpoint is frozen and queried on the student-visited states; only the student parameters are updated. For SaT, the same rollout horizon and batching axes are used, but teacher targets are generated by Exponax from the student-visited states with the vectorized solver backend described in~\cref{sec:solver_backends}. After post-training, each retained checkpoint is evaluated with a 50-step autoregressive rollout on the first five held-out test trajectories using the common inference engine.

\section{Additional Physical Diagnostics}
\label{sec:appendix:physical_diagnostics}

The main evaluation uses rollout RMSE as the primary accuracy metric, since it directly measures pointwise deviation from the held-out solver trajectory. RMSE alone, however, does not distinguish a uniformly small amplitude error from a loss of spatial gradients or a distorted physical property like spectral energy distribution. We therefore report two additional trajectory-level diagnostics on the same autoregressive rollouts: a Sobolev $H^{1}$ error, which emphasizes gradient-scale mismatch, and a normalized energy-spectrum error.

All three metrics compare a predicted rollout $\hat{\tau}=\{\hat{u}_{1},\dots,\hat{u}_{K}\}$ against the corresponding reference $\tau^{\mathrm{ref}}=\{u^{\mathrm{ref}}_{1},\dots,u^{\mathrm{ref}}_{K}\}$. Metrics are averaged over all $K$ rollout steps and $B$ test samples; for notational brevity, we write $N=BK$ and let a single index $i\in\{1,\dots,N\}$ range over the flattened sample--time axis.

The trajectory-level RMSE is
\begin{equation}
\mathrm{RMSE} = \sqrt{\frac{1}{N}\sum_{i=1}^{N} \bigl\|\hat{u}_{i}-u^{\mathrm{ref}}_{i}\bigr\|^{2}},
\label{eq:appendix_rmse_metric}
\end{equation}
where $\|\cdot\|$ denotes the $\ell_{2}$ norm over all channels and grid points, normalised by the total number of elements (i.e.\ $\|v\|^{2}=\frac{1}{C|\Omega_h|}\sum_{c,\mathbf{x}}|v_{c}(\mathbf{x})|^{2}$).

For the Sobolev diagnostic, let $e_{i}=\hat{u}_{i}-u^{\mathrm{ref}}_{i}$ denote the error field and $\widehat{e}_{i}(\boldsymbol{\nu})$ its orthonormalized discrete Fourier transform at normalized frequency $\boldsymbol{\nu}$ ($\nu_{j}=q_{j}/N_{j}$, with integer wavenumber $q_j$ and grid size $N_j$ along axis $j$). The first-order Sobolev error is
\begin{equation}
\mathcal{E}_{H^{1}} = \sqrt{\frac{1}{N}\sum_{i=1}^{N} \frac{1}{C|\Omega_h|}\sum_{c}\sum_{\boldsymbol{\nu}} \bigl(1+|\boldsymbol{\nu}|^{2}\bigr) \bigl|\widehat{e}_{i,c}(\boldsymbol{\nu})\bigr|^{2}}.
\label{eq:appendix_h1_metric}
\end{equation}
By Parseval's theorem with orthonormalized Fourier coefficients, $\sum_{\boldsymbol{\nu}}|\widehat{e}_{i,c}|^{2}=\sum_{\mathbf{x}}|e_{i,c}|^{2}$, so setting $s=0$ recovers RMSE exactly. At $s=1$ the factor $(1+|\boldsymbol{\nu}|^{2})$ indicates unresolved fine-scale structure more strongly than RMSE.

The energy-spectrum diagnostic compares the radial distribution of Fourier energy. For a field $z_{i}$, let $\widehat{z}_{i,c}(\boldsymbol{\nu})$ denote the un-normalized DFT coefficient of channel~$c$. Grouping modes into radial bins $\mathcal{B}_{r}$ by their normalized-frequency magnitude, the per-sample spectrum and its normalization are
\begin{equation}
E_{z}^{i}(r) = \frac{1}{2}\sum_{c}\sum_{\boldsymbol{\nu}\in\mathcal{B}_{r}} \bigl|\widehat{z}_{i,c}(\boldsymbol{\nu})\bigr|^{2}, \qquad \widetilde{E}_{z}^{i}(r) = \frac{E_{z}^{i}(r)}{\sum_{r'}E_{z}^{i}(r')}.
\label{eq:appendix_energy_spectrum}
\end{equation}
Each of the $N$ spectra is normalized to sum to one before averaging: $\bar{E}_{z}(r)=N^{-1}\sum_{i}\widetilde{E}_{z}^{i}(r)$. The reported metric is the relative discrepancy
\begin{equation}
\mathcal{E}_{\mathrm{spec}} = \frac{\bigl\|\bar{E}_{\hat{u}}-\bar{E}_{u^{\mathrm{ref}}}\bigr\|_{2}}{\bigl\|\bar{E}_{u^{\mathrm{ref}}}\bigr\|_{2}}.
\label{eq:appendix_spectrum_error}
\end{equation}
Because the spectra are normalized per sample before comparison, $\mathcal{E}_{\mathrm{spec}}$ focuses on spectral \emph{shape} rather than absolute amplitude. It is therefore complementary to both RMSE and~$\mathcal{E}_{H^{1}}$: RMSE measures pointwise tracking, $H^{1}$ measures fine-scale structure error, and the spectrum error probes whether the rollout preserves the scale-wise energy distribution expected from the governing dynamics.

Table~\ref{tab:physical_diagnostics_ood} reports the same diagnostics under parameter shifts. SaT-0 is best on all seven Sobolev $H^1$ rows, which is consistent with the main text's conclusion that full dynamic loss is the most robust choice once the model is evaluated away from the training distribution. Spectrum errors are more sensitive to the specific PDE and parameter shift, but a SaT variant is still best on six of seven OOD tasks; the only exception is Burgers2D-OOD, where OPD has the lowest spectrum error. This indicates that solver-guided on-policy training improves not only pointwise tracking but also the scale-wise statistics needed for credible chaotic and turbulent rollouts.

\begin{table}[htbp]
\centering
\caption{Cross-model aggregate physical diagnostics on OOD zero-shot test sets. Metrics are mean $\pm$ std over evaluated architectures; lower is better. Best per row is \textbf{bolded}. PITA does not support 3D tasks (--).}
\label{tab:physical_diagnostics_ood}
\vspace{2pt}
\scriptsize
\setlength{\tabcolsep}{2.6pt}
\begin{adjustbox}{max width=\textwidth}
\begin{tabular}{@{}ll c c c c c c c@{}}
\toprule
Task & Metric & SFT & PITA & OPD & SaT-1 & SaT-.5 & SaT-0 & N models \\
\midrule
\multirow{2}{*}{Burgers2D-OOD} & $H^1$ & $0.0857 \pm 0.0157$ & $0.0771 \pm 0.0079$ & $0.0739 \pm 0.0016$ & $0.0727 \pm 0.0014$ & $0.0729 \pm 0.0021$ & $\mathbf{0.0723 \pm 0.0010}$ & \multirow{2}{*}{4} \\
 & Spectrum & $0.1939 \pm 0.0202$ & $0.1860 \pm 0.0214$ & $\mathbf{0.1852 \pm 0.0117}$ & $0.1918 \pm 0.0072$ & $0.1884 \pm 0.0117$ & $0.1893 \pm 0.0077$ & \\
\midrule
\multirow{2}{*}{KF2D-OOD} & $H^1$ & $2.4544 \pm 0.2901$ & $2.4703 \pm 0.2257$ & $2.4244 \pm 0.2118$ & $2.3333 \pm 0.1739$ & $2.5008 \pm 0.2599$ & $\mathbf{2.2669 \pm 0.2140}$ & \multirow{2}{*}{4} \\
 & Spectrum & $0.1959 \pm 0.0296$ & $0.2027 \pm 0.0272$ & $0.2226 \pm 0.0133$ & $0.2134 \pm 0.0216$ & $\mathbf{0.1911 \pm 0.0176}$ & $0.2147 \pm 0.0197$ & \\
\midrule
\multirow{2}{*}{KS2D-OOD} & $H^1$ & $0.1376 \pm 0.0311$ & $0.1203 \pm 0.0348$ & $0.1265 \pm 0.0282$ & $0.1253 \pm 0.0307$ & $0.1072 \pm 0.0262$ & $\mathbf{0.1063 \pm 0.0176}$ & \multirow{2}{*}{4} \\
 & Spectrum & $0.2416 \pm 0.1864$ & $0.1222 \pm 0.0465$ & $0.2181 \pm 0.2132$ & $0.2538 \pm 0.2230$ & $\mathbf{0.0967 \pm 0.0510}$ & $0.1365 \pm 0.0968$ & \\
\midrule
\multirow{2}{*}{Burgers3D-OOD} & $H^1$ & $0.0177 \pm 0.0026$ & -- & $0.0144 \pm 0.0020$ & $0.0132 \pm 0.0012$ & $0.0142 \pm 0.0020$ & $\mathbf{0.0130 \pm 0.0010}$ & \multirow{2}{*}{3} \\
 & Spectrum & $0.0803 \pm 0.0394$ & -- & $0.0380 \pm 0.0039$ & $0.0418 \pm 0.0234$ & $0.0298 \pm 0.0030$ & $\mathbf{0.0297 \pm 0.0118}$ & \\
\midrule
\multirow{2}{*}{KF3D-OOD} & $H^1$ & $1.0929 \pm 1.2117$ & -- & $0.2036 \pm 0.0055$ & $0.2004 \pm 0.0042$ & $0.1980 \pm 0.0049$ & $\mathbf{0.1945 \pm 0.0045}$ & \multirow{2}{*}{3} \\
 & Spectrum & $0.2196 \pm 0.1327$ & -- & $0.1562 \pm 0.0498$ & $\mathbf{0.1317 \pm 0.0258}$ & $0.1335 \pm 0.0161$ & $0.1319 \pm 0.0230$ & \\
\midrule
\multirow{2}{*}{KS3D-OOD} & $H^1$ & $2.6867 \pm 3.5221$ & -- & $0.2910 \pm 0.0965$ & $0.2016 \pm 0.0189$ & $0.1966 \pm 0.0169$ & $\mathbf{0.1929 \pm 0.0228}$ & \multirow{2}{*}{3} \\
 & Spectrum & $0.1683 \pm 0.1695$ & -- & $0.0944 \pm 0.0161$ & $0.0876 \pm 0.0569$ & $0.0723 \pm 0.0445$ & $\mathbf{0.0690 \pm 0.0507}$ & \\
\midrule
\multirow{2}{*}{DT3D-OOD} & $H^1$ & $0.2520 \pm 0.2880$ & -- & $0.0438 \pm 0.0115$ & $0.0428 \pm 0.0064$ & $0.0436 \pm 0.0077$ & $\mathbf{0.0372 \pm 0.0028}$ & \multirow{2}{*}{3} \\
 & Spectrum & $0.3981 \pm 0.0732$ & -- & $0.3301 \pm 0.0335$ & $0.3178 \pm 0.0207$ & $\mathbf{0.3024 \pm 0.0174}$ & $0.3060 \pm 0.0158$ & \\
\bottomrule
\end{tabular}
\end{adjustbox}
\end{table}

\section{Theoretical Analysis}
\label{app:theoretical_analysis}

We formalize the distribution-alignment advantage of Solver-as-Teacher
(SaT). In particular, we show that the dynamics component of the SaT
objective directly controls the one-step solver-correction error on
student-visited states. Combined with a standard no-regret
online-learning argument and a stability assumption on the numerical
solver, this yields a bound on the autoregressive rollout error.

The analysis is stated for the DAgger-style interpretation of SaT. At
iteration \(n\), the current model \(G_n\) first generates a rollout
from an initial condition \(u_0\sim\rho_0\):
\begin{align}
\hat u_{n,0}
&=
u_0,
&
\hat u_{n,k+1}
&=
G_n(\hat u_{n,k}),
\qquad
k=0,\ldots,K-1.
\label{eq:theory_student_rollout}
\end{align}
Let \(d_{G_n}^k\) denote the distribution of \(\hat u_{n,k}\). The
corresponding reference trajectory is generated by the solver:
\begin{align}
u_0^{\mathrm{ref}}
&=
u_0,
&
u_{k+1}^{\mathrm{ref}}
&=
S(u_k^{\mathrm{ref}}).
\label{eq:theory_reference_rollout}
\end{align}

For each student-visited state \(\hat u_{n,k}\), define
\begin{align}
m_k
:=
\min(m_{\max},K-k).
\end{align}
After collecting the rollout, the states are treated as fixed training
inputs. For any candidate model \(G\in\mathcal G\), let \(G^j\) denote
the \(j\)-fold composition of \(G\), and define the round-\(n\)
dynamics and anchor losses by
\begin{align}
\mathcal L_{\mathrm{dyn},n}(G)
&:=
\frac{1}{W_{\mathrm{dyn}}}
\sum_{k=0}^{K-1}
\sum_{j=1}^{m_k}
\gamma^k\lambda^{j-1}
\mathbb E_{u_0\sim\rho_0}
\left[
\ell\left(
G^j(\hat u_{n,k}),
S^j(\hat u_{n,k})
\right)
\right],
\label{eq:theory_dynamic_loss}
\\
\mathcal L_{\mathrm{anchor},n}(G)
&:=
\frac{1}{W_{\mathrm{anchor}}}
\sum_{k=0}^{K-1}
\gamma^k
\mathbb E_{u_0\sim\rho_0}
\left[
\ell\left(
G(\hat u_{n,k}),
u_{k+1}^{\mathrm{ref}}
\right)
\right].
\label{eq:theory_anchor_loss}
\end{align}
Here \(W_{\mathrm{dyn}}\) and \(W_{\mathrm{anchor}}\) are the same
normalization constants as in the main SaT objective. The complete
round-\(n\) loss is
\begin{align}
\mathcal L_n^\alpha(G)
:=
(1-\alpha)\mathcal L_{\mathrm{dyn},n}(G)
+
\alpha\mathcal L_{\mathrm{anchor},n}(G),
\qquad
\alpha\in[0,1].
\label{eq:theory_round_loss}
\end{align}
When evaluated at \(G=G_n\), this is precisely the expected version of
the training objective in the main text, since
\begin{align}
G_n^j(\hat u_{n,k})
=
\hat u_{n,k+j}.
\end{align}

We define the one-step on-policy solver-correction error of the
\(n\)-th iterate as
\begin{align}
\mathcal E_{\mathrm{on}}(G_n)
:=
\frac{1}{K}
\sum_{k=0}^{K-1}
\mathbb E_{\hat u_{n,k}\sim d_{G_n}^k}
\left[
\ell\left(
G_n(\hat u_{n,k}),
S(\hat u_{n,k})
\right)
\right].
\label{eq:theory_on_policy_error}
\end{align}

For the sequence of online losses
\(\{\mathcal L_n^\alpha\}_{n=1}^N\), define the best-in-hindsight
approximation error
\begin{align}
\varepsilon_N^\alpha
:=
\min_{G\in\mathcal G}
\frac{1}{N}
\sum_{n=1}^N
\mathcal L_n^\alpha(G),
\label{eq:theory_approximation_error}
\end{align}
and the online learning regret
\begin{align}
R_N
:=
\sum_{n=1}^N
\mathcal L_n^\alpha(G_n)
-
\min_{G\in\mathcal G}
\sum_{n=1}^N
\mathcal L_n^\alpha(G).
\label{eq:theory_regret}
\end{align}

Finally, for a fixed model \(G\), let
\(\hat u_{k+1}=G(\hat u_k)\) and
\(u_{k+1}^{\mathrm{ref}}=S(u_k^{\mathrm{ref}})\), with
\(\hat u_0=u_0^{\mathrm{ref}}\). We define the rollout error by
\begin{align}
\mathcal E_{\mathrm{SaT}}(G;S)
:=
\left(
\frac{1}{K}
\sum_{k=1}^K
\mathbb E
\left[
\left\|
\hat u_k-u_k^{\mathrm{ref}}
\right\|_{L^2(\Omega)}^2
\right]
\right)^{1/2}.
\label{eq:theory_rollout_error}
\end{align}
This is the population analogue of the trajectory-level rollout RMSE
used in the experiments.

\begin{theorem}[On-policy and rollout guarantees for SaT]
\label{thm:sat_theoretical_guarantee}
Suppose that \(m_{\max}\geq 1\),
\(\gamma,\lambda\in(0,1]\), and \(\alpha\in[0,1)\). Assume that the
solver transition is \(L_S\)-Lipschitz on the states visited by the
solver and student rollouts:
\begin{align}
\left\|
S(u)-S(v)
\right\|_{L^2(\Omega)}
\leq
L_S
\left\|
u-v
\right\|_{L^2(\Omega)}.
\label{eq:theory_solver_lipschitz}
\end{align}
The composite loss in the main text satisfies
\begin{align}
\ell(v,w)
\geq
\beta_1
\left\|
v-w
\right\|_{L^2(\Omega)}^2,
\label{eq:theory_loss_lower_bound}
\end{align}
because its Sobolev component is nonnegative.

Define
\begin{align}
\kappa_{K,m_{\max},\gamma,\lambda}
:=
\frac{
W_{\mathrm{dyn}}
}{
K\gamma^{K-1}
}.
\label{eq:theory_kappa_definition}
\end{align}
Then there exists an iterate
\(\hat n\in\{1,\ldots,N\}\) such that
\begin{align}
\mathcal E_{\mathrm{on}}(G_{\hat n})
\leq
\frac{
\kappa_{K,m_{\max},\gamma,\lambda}
}{
1-\alpha
}
\left(
\varepsilon_N^\alpha
+
\frac{R_N}{N}
\right).
\label{eq:theory_best_iterate_bound}
\end{align}
Moreover, the same iterate satisfies
\begin{align}
\mathcal E_{\mathrm{SaT}}(G_{\hat n};S)
\leq
\left(
\sum_{r=0}^{K-1}
L_S^r
\right)
\sqrt{
\frac{
\kappa_{K,m_{\max},\gamma,\lambda}
}{
\beta_1(1-\alpha)
}
\left(
\varepsilon_N^\alpha
+
\frac{R_N}{N}
\right)
}.
\label{eq:theory_rollout_bound}
\end{align}
\end{theorem}

\begin{proof}
We first show that the SaT training objective controls the one-step
on-policy correction error. Since all loss terms are nonnegative and
\(m_k\geq 1\), the \(j=1\) terms of the dynamics loss give
\begin{align}
\mathcal L_n^\alpha(G_n)
&=
(1-\alpha)
\mathcal L_{\mathrm{dyn},n}(G_n)
+
\alpha
\mathcal L_{\mathrm{anchor},n}(G_n)
\nonumber\\
&\geq
\frac{1-\alpha}{W_{\mathrm{dyn}}}
\sum_{k=0}^{K-1}
\gamma^k
\mathbb E
\left[
\ell\left(
G_n(\hat u_{n,k}),
S(\hat u_{n,k})
\right)
\right].
\label{eq:proof_keep_one_step}
\end{align}
Because \(\gamma\in(0,1]\), we have
\(\gamma^k\geq\gamma^{K-1}\) for every
\(k\in\{0,\ldots,K-1\}\). Hence
\begin{align}
\mathcal L_n^\alpha(G_n)
&\geq
\frac{
(1-\alpha)\gamma^{K-1}
}{
W_{\mathrm{dyn}}
}
\sum_{k=0}^{K-1}
\mathbb E
\left[
\ell\left(
G_n(\hat u_{n,k}),
S(\hat u_{n,k})
\right)
\right]
\nonumber\\
&=
\frac{
(1-\alpha)K\gamma^{K-1}
}{
W_{\mathrm{dyn}}
}
\mathcal E_{\mathrm{on}}(G_n).
\label{eq:proof_objective_controls_on_policy}
\end{align}
Rearranging and using
\eqref{eq:theory_kappa_definition}, we obtain
\begin{align}
\mathcal E_{\mathrm{on}}(G_n)
\leq
\frac{
\kappa_{K,m_{\max},\gamma,\lambda}
}{
1-\alpha
}
\mathcal L_n^\alpha(G_n).
\label{eq:proof_transfer_bound}
\end{align}

Averaging \eqref{eq:proof_transfer_bound} over
\(n=1,\ldots,N\) gives
\begin{align}
\frac{1}{N}
\sum_{n=1}^N
\mathcal E_{\mathrm{on}}(G_n)
&\leq
\frac{
\kappa_{K,m_{\max},\gamma,\lambda}
}{
1-\alpha
}
\frac{1}{N}
\sum_{n=1}^N
\mathcal L_n^\alpha(G_n)
\nonumber\\
&=
\frac{
\kappa_{K,m_{\max},\gamma,\lambda}
}{
1-\alpha
}
\left[
\min_{G\in\mathcal G}
\frac{1}{N}
\sum_{n=1}^N
\mathcal L_n^\alpha(G)
+
\frac{R_N}{N}
\right]
\nonumber\\
&=
\frac{
\kappa_{K,m_{\max},\gamma,\lambda}
}{
1-\alpha
}
\left(
\varepsilon_N^\alpha
+
\frac{R_N}{N}
\right).
\label{eq:proof_average_on_policy}
\end{align}
Since the minimum of \(N\) nonnegative numbers is no larger than their
average, there exists
\(\hat n\in\{1,\ldots,N\}\) satisfying
\eqref{eq:theory_best_iterate_bound}.

It remains to relate the one-step correction error to rollout RMSE.
Fix \(G=G_{\hat n}\), and define
\begin{align}
e_k
&:=
\left\|
\hat u_k-u_k^{\mathrm{ref}}
\right\|_{L^2(\Omega)},
&
\delta_k
&:=
\left\|
G(\hat u_k)-S(\hat u_k)
\right\|_{L^2(\Omega)}.
\label{eq:proof_error_definitions}
\end{align}
Using the student and solver recursions together with
\eqref{eq:theory_solver_lipschitz}, we have
\begin{align}
e_{k+1}
&=
\left\|
G(\hat u_k)-S(u_k^{\mathrm{ref}})
\right\|_{L^2(\Omega)}
\nonumber\\
&\leq
\left\|
G(\hat u_k)-S(\hat u_k)
\right\|_{L^2(\Omega)}
+
\left\|
S(\hat u_k)-S(u_k^{\mathrm{ref}})
\right\|_{L^2(\Omega)}
\nonumber\\
&\leq
\delta_k
+
L_S e_k.
\label{eq:proof_error_recursion}
\end{align}
Since \(e_0=0\), recursively applying
\eqref{eq:proof_error_recursion} yields
\begin{align}
e_k
\leq
\sum_{j=0}^{k-1}
L_S^{k-1-j}
\delta_j.
\label{eq:proof_unrolled_error}
\end{align}
The right-hand side is a causal discrete convolution between
\(\{\delta_j\}\) and the kernel
\(\{1,L_S,\ldots,L_S^{K-1}\}\). By the discrete Young inequality,
\begin{align}
\left(
\sum_{k=1}^K
e_k^2
\right)^{1/2}
\leq
\left(
\sum_{r=0}^{K-1}
L_S^r
\right)
\left(
\sum_{j=0}^{K-1}
\delta_j^2
\right)^{1/2}.
\label{eq:proof_young}
\end{align}
Squaring, taking expectations, and dividing by \(K\) gives
\begin{align}
\mathcal E_{\mathrm{SaT}}^2(G;S)
&\leq
\left(
\sum_{r=0}^{K-1}
L_S^r
\right)^2
\frac{1}{K}
\sum_{j=0}^{K-1}
\mathbb E[\delta_j^2].
\label{eq:proof_rollout_from_local_error}
\end{align}
By \eqref{eq:theory_loss_lower_bound},
\begin{align}
\frac{1}{K}
\sum_{j=0}^{K-1}
\mathbb E[\delta_j^2]
&\leq
\frac{1}{\beta_1 K}
\sum_{j=0}^{K-1}
\mathbb E
\left[
\ell\left(
G(\hat u_j),
S(\hat u_j)
\right)
\right]
\nonumber\\
&=
\frac{1}{\beta_1}
\mathcal E_{\mathrm{on}}(G).
\label{eq:proof_local_error_from_loss}
\end{align}
Combining
\eqref{eq:proof_rollout_from_local_error} and
\eqref{eq:proof_local_error_from_loss}, and then taking square roots,
gives
\begin{align}
\mathcal E_{\mathrm{SaT}}(G;S)
\leq
\left(
\sum_{r=0}^{K-1}
L_S^r
\right)
\sqrt{
\frac{
\mathcal E_{\mathrm{on}}(G)
}{
\beta_1
}
}.
\label{eq:proof_rollout_from_on_policy}
\end{align}
Substituting
\eqref{eq:theory_best_iterate_bound} into
\eqref{eq:proof_rollout_from_on_policy} proves
\eqref{eq:theory_rollout_bound}.
\end{proof}

\begin{remark}[Dependence on the anchor coefficient]
\label{rem:theory_alpha}
The restriction \(\alpha<1\) is essential. The guarantee is obtained
from the dynamics component of the objective, whose coefficient is
\(1-\alpha\). When \(\alpha=1\), this component vanishes, and the
objective reduces to SaT-1, namely uniformly weighted unrolled SFT with
no on-policy solver-correction signal. Accordingly, the bound becomes
vacuous at \(\alpha=1\). In its explicit dependence on \(\alpha\), the
bound is strongest for SaT-0 and becomes weaker as more weight is
assigned to the reference anchor.
\end{remark}

\begin{remark}[Explicit form of the transfer constant]
\label{rem:theory_kappa}
If the dynamics normalizer is chosen as the sum of the dynamic weights,
\begin{align}
W_{\mathrm{dyn}}
=
\sum_{k=0}^{K-1}
\sum_{j=1}^{m_k}
\gamma^k\lambda^{j-1},
\label{eq:theory_weight_sum_normalizer}
\end{align}
then the transfer constant becomes
\begin{align}
\kappa_{K,m_{\max},\gamma,\lambda}
=
\frac{
\displaystyle
\sum_{k=0}^{K-1}
\sum_{j=1}^{m_k}
\gamma^k\lambda^{j-1}
}{
K\gamma^{K-1}
}.
\label{eq:theory_explicit_kappa}
\end{align}
For \(m_{\max}=1\) and \(\gamma=1\), this gives
\(\kappa_{K,1,1,\lambda}=1\). With multi-step branching, the theorem
uses only the fraction of the normalized objective assigned to
one-step targets, which explains the dependence on
\(m_{\max}\) and \(\lambda\).
\end{remark}

\begin{remark}[Why reference-state SFT requires coverage]
\label{rem:theory_sft_coverage}
Standard SFT controls the one-step error under solver-generated
reference states:
\begin{align}
\mathcal E_{\mathrm{SFT}}(G)
:=
\frac{1}{K}
\sum_{k=0}^{K-1}
\mathbb E_{u\sim d_S^k}
\left[
\ell\left(
G(u),
S(u)
\right)
\right],
\label{eq:theory_sft_error}
\end{align}
where \(d_S^k\) is the distribution of \(u_k^{\mathrm{ref}}\). To
convert this quantity into an error under the deployment distribution
\(d_G^k\), one needs an absolute-continuity condition. Define
\begin{align}
C(G)
:=
\max_{0\leq k<K}
\left\|
\frac{\mathrm d d_G^k}
{\mathrm d d_S^k}
\right\|_{L^\infty(d_S^k)},
\label{eq:theory_coverage_coefficient}
\end{align}
with \(C(G)=+\infty\) if \(d_G^k\) is not absolutely continuous with
respect to \(d_S^k\) for some \(k\). A change of measure gives
\begin{align}
\mathcal E_{\mathrm{on}}(G)
\leq
C(G)
\mathcal E_{\mathrm{SFT}}(G).
\label{eq:theory_sft_change_measure}
\end{align}
Consequently, an SFT-based rollout guarantee necessarily contains the
additional factor \(C(G)\):
\begin{align}
\mathcal E_{\mathrm{SaT}}(G;S)
\leq
\left(
\sum_{r=0}^{K-1}
L_S^r
\right)
\sqrt{
\frac{
C(G)\mathcal E_{\mathrm{SFT}}(G)
}{
\beta_1
}
}.
\label{eq:theory_sft_rollout_bound}
\end{align}
In deterministic systems with a fixed initial condition, once the
student and reference trajectories reach different states,
\(d_G^k\) and \(d_S^k\) may be distinct Dirac measures, in which case
\(C(G)=+\infty\). Thus, a small SFT loss on reference states alone need
not imply a small deployment error.
\end{remark}

\begin{remark}[Why SaT-1 does not provide solver correction]
\label{rem:theory_unrolled_sft}
SaT-1 rolls out the student but compares its prediction against the
time-aligned reference target \(u_{k+1}^{\mathrm{ref}}\). The locally
correct solver target at the student-visited state is instead
\(S(\hat u_k)\). These two targets differ whenever the student has
departed from the reference trajectory. Indeed,
\begin{align}
\left\|
G(\hat u_k)-S(\hat u_k)
\right\|_{L^2(\Omega)}
&\leq
\left\|
G(\hat u_k)-u_{k+1}^{\mathrm{ref}}
\right\|_{L^2(\Omega)}
+
\left\|
u_{k+1}^{\mathrm{ref}}-S(\hat u_k)
\right\|_{L^2(\Omega)}
\nonumber\\
&\leq
\left\|
G(\hat u_k)-u_{k+1}^{\mathrm{ref}}
\right\|_{L^2(\Omega)}
+
L_S
\left\|
\hat u_k-u_k^{\mathrm{ref}}
\right\|_{L^2(\Omega)}.
\label{eq:theory_unrolled_sft_mismatch}
\end{align}
Therefore, even if the time-aligned anchor loss is small, an additional
state-distribution mismatch term remains. SaT-0 and SaT-.5 remove this
ambiguity through the explicit target \(S(\hat u_k)\).
\end{remark}

\begin{remark}[Scope of the no-regret argument]
\label{rem:theory_dagger_scope}
The online-to-batch part of
Theorem~\ref{thm:sat_theoretical_guarantee} treats the rollout states
\(\hat u_{n,k}\) as fixed after they are collected. It therefore
corresponds most directly to the DAgger/detach implementation, in which
the states are generated without gradient and then independently
re-forwarded for optimization. The BPTT implementation uses the same
solver targets and the same objective values, but additionally
differentiates through the state-generation process. Establishing a
no-regret result for that parameter-dependent state distribution would
require a separate nonconvex optimization analysis.

The theorem is consequently conditional on the realized regret \(R_N\).
When \(R_N=o(N)\), the online optimization contribution
\(R_N/N\) vanishes, leaving only the best-in-class approximation error
\(\varepsilon_N^\alpha\).
\end{remark}

\begin{remark}[Horizon dependence and the DAgger interpretation]
\label{rem:theory_horizon}
The rollout amplification factor is
\begin{align}
A_K(L_S)
:=
\sum_{r=0}^{K-1}
L_S^r
=
\begin{cases}
\dfrac{1-L_S^K}{1-L_S},
&
L_S\neq 1,
\\[6pt]
K,
&
L_S=1.
\end{cases}
\label{eq:theory_horizon_factor}
\end{align}
When \(L_S<1\), local errors contract and \(A_K(L_S)\) is uniformly
bounded in \(K\). When \(L_S=1\), the worst-case dependence is linear,
whereas \(L_S>1\) permits geometric error amplification, as expected
in unstable or chaotic regimes.

In the classical bounded-error DAgger interpretation \citep{ross2011dagger}, SFT controls a
per-step error \(\epsilon\) only under the expert-state distribution.
The probability of having departed from that distribution by step
\(k\) can then scale as \(O(k\epsilon)\), producing
\(O(K^2\epsilon)\) cumulative error. On-policy data aggregation instead
controls the error on learner-visited states and gives
\(O(K\epsilon)\) cumulative error when the expert can recover from
local deviations. The continuous-state result above is the analogous
statement for the SaT loss: distribution alignment is captured by
\(\mathcal E_{\mathrm{on}}\), while error propagation is captured by
\(A_K(L_S)\).
\end{remark}

\section{\texorpdfstring{Higher Resolution: $96^3$}{Higher Resolution: 96\textasciicircum 3}}
\label{sec:appendix:highres}

We retrained DPOT-Ti and P3D-S with SaT-1 and SaT-0 on all four 3D tasks at $96^3$, one run per cell, with full fine-tuning, learning rate $10^{-3}$, a plateau scheduler, $K=5$, $\gamma=\lambda=0.9$, effective batch size 48, and 160 epochs including a 40-epoch single-step warm-up, on NVIDIA H100 GPUs. Data follow the $64^3$ setup except for the grid size. \Cref{tab:highres} reports cross-model 50-step RMSE (mean $\pm$ std over the two backbones) in units of $10^{-2}$.

\begin{table}[h]\centering
\caption{Cross-model 50-step RMSE at $96^3$ ($\times10^{-2}$; DPOT-Ti and P3D-S).}
\label{tab:highres}
\vspace{2pt}
\small
\begin{tabular}{@{}lcc@{}}
\toprule
Task & SaT-1 & SaT-0 \\
\midrule
Burgers3D & $1.27 \pm 0.49$ & $\mathbf{1.11 \pm 0.40}$ \\
KF3D & $12.97 \pm 3.25$ & $\mathbf{12.46 \pm 2.90}$ \\
KS3D & $143.67 \pm 113.91$ & $\mathbf{21.65 \pm 7.46}$ \\
DT3D & $4.06 \pm 1.06$ & $\mathbf{2.95 \pm 0.01}$ \\
\bottomrule
\end{tabular}
\end{table}

Clearly, SaT-0 improves on SaT-1 on all four tasks at $96^3$, as at $64^3$.

\section{Static Solver Data versus On-Policy Targets}
\label{sec:appendix:static_data}

To separate the effect of more solver data from that of querying the solver at student-visited states, we retrained SaT-1 on an extended static dataset generated by the same solver from fresh initial conditions, with the same number of targets per epoch as SaT-0 (SaT-1-Ext). All other columns are the values from \cref{tab:rmse_comparison}. \Cref{tab:static_data} reports 50-step RMSE in units of $10^{-2}$.

\begin{table}[h]\centering
\caption{Extended static data (SaT-1-Ext) against on-policy targets, 50-step RMSE ($\times10^{-2}$). Best per row \textbf{bolded}.}
\label{tab:static_data}
\vspace{2pt}
\small
\resizebox{\textwidth}{!}{
\begin{tabular}{@{}llrrrrrr@{}}
\toprule
Task & Model & Pretrained & SFT & SaT-1 & SaT-1-Ext & SaT-.5 & SaT-0 \\
\midrule
Burgers2D & DPOT-Ti & 245.02 & 1.26 & 1.45 & 1.20 & \textbf{0.66} & 1.10 \\
KS2D & DPOT-Ti & 138.53 & 6.07 & 6.00 & 5.54 & \textbf{4.39} & 5.49 \\
KS3D & DPOT-Ti & 47.35 & 17.59 & 18.44 & 18.19 & 17.92 & \textbf{16.77} \\
KF3D & DPOT-Ti & 44.58 & 16.50 & 12.41 & 11.49 & 11.62 & \textbf{10.99} \\
KS3D & P3D-S & 235865.41 & 624.59 & 44.65 & 29.70 & \textbf{25.34} & 27.52 \\
KF3D & P3D-S & 77721.85 & 175.30 & 15.18 & 15.00 & 14.95 & \textbf{14.78} \\
\bottomrule
\end{tabular}
}

\end{table}

More static data improves SaT-1 in all six cells, by $1.2\%$ to $33.5\%$, but an on-policy variant remains best in every row, by $1.5\%$ to $45.0\%$ over SaT-1-Ext. The best on-policy variant remains lower than both SaT-1 and SaT-1-Ext ($25.34$ vs.\ $29.70$ on KS3D; $14.78$ vs.\ $15.00$ on KF3D).

\section{On-Policy Distillation with Task-Specific Learned Teachers}
\label{sec:appendix:learned_teacher}

OPD in the main text uses a large fine-tuned foundation model as teacher. To test whether a teacher that matches the task's inductive bias closes the gap, we replaced it with a task-specific operator trained from scratch, an FNO in 2D and an AFNO in 3D (OPD-FNO/AFNO), and kept the student protocol unchanged. OPD and SaT-0 are the values from \cref{tab:cross_model_diagnostics_id}. \Cref{tab:learned_teacher} reports cross-model 50-step RMSE in units of $10^{-2}$.

\begin{table}[h]\centering
\caption{OPD with a large FM teacher, OPD with a task-specific FNO/AFNO teacher, and SaT-0; cross-model 50-step RMSE ($\times10^{-2}$).}
\label{tab:learned_teacher}
\vspace{2pt}
\small
\begin{tabular}{@{}lccc@{}}
\toprule
Task & OPD & OPD-FNO/AFNO & SaT-0 \\
\midrule
Burgers2D & $1.93 \pm 0.96$ & $2.27 \pm 1.51$ & $\mathbf{1.18 \pm 0.40}$ \\
KS2D & $10.2 \pm 4.22$ & $15.2 \pm 2.34$ & $\mathbf{8.41 \pm 4.19}$ \\
KF2D & $78.5 \pm 28.3$ & $89.3 \pm 37.7$ & $\mathbf{54.1 \pm 20.9}$ \\
Burgers3D & $1.09 \pm 0.12$ & $2.21 \pm 0.07$ & $\mathbf{0.80 \pm 0.14}$ \\
KS3D & $37.7 \pm 20.8$ & $93.1 \pm 93.5$ & $\mathbf{22.4 \pm 4.40}$ \\
KF3D & $14.2 \pm 1.43$ & $21.5 \pm 2.50$ & $\mathbf{12.9 \pm 1.55}$ \\
DT3D & $3.98 \pm 1.71$ & $5.68 \pm 0.72$ & $\mathbf{3.04 \pm 0.91}$ \\
\bottomrule
\end{tabular}
\end{table}

Although these operators match the periodic, pseudo-spectral setting, students distilled from them are $1.14\times$ to $2.47\times$ worse than students distilled from the large teacher, and SaT-0 is lower than both learned teachers on all seven tasks. A learned operator is fitted on the data manifold and need not remain accurate after the student drifts, whereas a numerical solver stays consistent with its governing equation and discretization.

\section{On-Policy Physics-Informed Baseline}
\label{sec:appendix:pino}

To test whether on-policy sampling with PDE residuals suffices, we post-trained the 2D students with a PINO-style residual loss~\cite{li2024physics} evaluated at student-visited states, without solver targets (PINO-OP-0), and with an additional reference-trajectory anchor of weight $0.5$ (PINO-OP-.5). SaT-0 is the value from \cref{tab:cross_model_diagnostics_id}. \Cref{tab:pino_op} reports cross-model 50-step RMSE in units of $10^{-2}$.

\begin{table}[h]\centering
\caption{On-policy residual baselines and SaT-0; cross-model 50-step RMSE ($\times10^{-2}$).}
\label{tab:pino_op}
\vspace{2pt}
\small
\begin{tabular}{@{}lccc@{}}
\toprule
Task & PINO-OP-0 & PINO-OP-.5 & SaT-0 \\
\midrule
Burgers2D & $11.46 \pm 7.27$ & $3.01 \pm 2.47$ & $\mathbf{1.18 \pm 0.40}$ \\
KS2D & $33.55 \pm 7.73$ & $16.14 \pm 5.78$ & $\mathbf{8.41 \pm 4.19}$ \\
KF2D & $181.31 \pm 31.06$ & $90.91 \pm 29.48$ & $\mathbf{54.09 \pm 20.89}$ \\
\bottomrule
\end{tabular}
\end{table}

Clearly, SaT-0 substantially outperforms both PINO-OP-0 and PINO-OP-.5 across all three reported tasks. We attribute the gap between SaT and PINO-OP to two complementary properties of SaT's multi-step targets. First, SaT supervises the state directly, whereas PINO-OP optimizes a residual objective, which is harder to optimize: adding a reference-trajectory anchor reduces the error of PINO-OP by $49.9\%$ to $73.7\%$, and SaT-0 reduces it by a further $40.5\%$ to $60.8\%$ by using solver-derived targets at student-visited states. Second, and most importantly, the targets are on-policy and finite-horizon. SaT matches the student's $j$-step prediction from $\hat{u}_k$ to the solver output $\mathcal{S}^{j}(\hat{u}_k)$ for several horizons $j$. Both sides of each loss term $\ell(\hat{u}_{k+j}, \mathcal{S}^{j}(\hat{u}_k))$ start from the same student-visited state, so each term isolates the error introduced within its window and penalizes it at the magnitude it reaches after propagating through the intervening dynamics. PINO-OP instead constrains each visited state through a local residual, which measures an error only at its instantaneous size. The difference matters most in chaotic regimes, where errors along unstable directions grow rapidly. This comparison covers the three 2D tasks; without an anchor, PINO-OP-0 is poorly conditioned on these chaotic systems, which makes the comparison favorable to SaT.

\section{Wall-Bounded Turbulent Channel Flow with a Finite-Volume Solver}
\label{sec:appendix:tcf}

TCF3D is turbulent channel flow at $\mathrm{Re}_\tau=180$ from FluidGym~\cite{becktepe2026fluidgym}, on a $64\times65\times64$ grid refined toward the walls, with no-slip walls in the wall-normal direction and periodic boundaries in the streamwise and spanwise directions. The state has four channels $(u_x,u_y,u_z,p)$, and one step advances $\Delta t=0.005$. The teacher is PICT~\cite{franz2025pict}, a PyTorch finite-volume solver with an implicit PISO pressure projection and an iterative linear solve at every step. It shares no component with Exponax, and SaT uses it through the same solver interface without changes. Pressure enforces incompressibility rather than evolving as a state, so the post-training losses and all reported metrics use the three velocity channels.

Each run uses a 40-epoch single-step warm-up and 160 main epochs, full fine-tuning, learning rate $10^{-3}$ with a cosine schedule, and effective batch size 32. The checkpoint is selected by 50-step RMSE on a held-out validation set. OPD uses P3D-L, fine-tuned on TCF3D, as its teacher, with a batch size of 6. We report the best run per model and method in \cref{tab:tcf}.

\begin{table}[h]\centering
\caption{TCF3D velocity RMSE after 50 steps. Best per row \textbf{bolded}.}
\label{tab:tcf}
\vspace{2pt}
\small
\setlength{\tabcolsep}{3pt}
\begin{tabular}{@{}lccccc@{}}
\toprule
Model & SFT & OPD & SaT-1 & SaT-.5 & SaT-0 \\
\midrule
DPOT-Ti & 0.4431 & 0.5252 & 0.4162 & 0.4138 & \textbf{0.4092} \\
MORPH-Ti & 0.6402 & 0.5776 & 0.5253 & 0.5192 & \textbf{0.5060} \\
P3D-S & 21.4909 & 0.8161 & 0.1388 & \textbf{0.0863} & 0.1022 \\
\midrule
Mean $\pm$ std & $7.5247 \pm 9.8759$ & $0.6396 \pm 0.1266$ & $0.3601 \pm 0.1627$ & $0.3398 \pm 0.1843$ & $\mathbf{0.3391 \pm 0.1721}$ \\
\bottomrule
\end{tabular}
\end{table}

\begin{figure}[htbp]\centering
  \includegraphics[width=\textwidth]{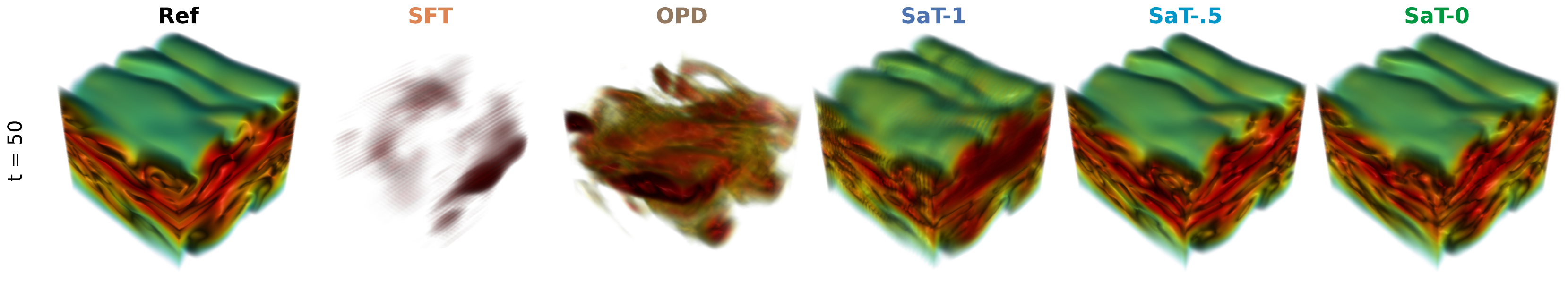}
  \caption{TCF3D, velocity magnitude at $t=50$ (P3D-S). SFT has diverged and retains no flow structure. OPD loses the layered channel-flow profile. All three SaT variants keep the layered profile and the near-wall streaks: SaT-1 blurs the streaks and smears the small-scale eddies, while SaT-.5 and SaT-0 remain closest to the reference, consistent with \cref{tab:tcf}.}
  \label{fig:tcf_traj}
\end{figure}

A SaT variant gives the lowest velocity RMSE on all three backbones (SaT-0 on DPOT-Ti and MORPH-Ti, SaT-.5 on P3D-S), and SaT-0 the lowest cross-model RMSE; SFT diverges on P3D-S. Figure~\ref{fig:tcf_traj} shows the P3D-S velocity field of every method after 50 steps.

\section{Detailed In-Distribution Per-Model RMSE}
\label{sec:appendix:rmse_per_model}

Table~\ref{tab:rmse_comparison} reports the per-model 50-step rollout RMSE that is aggregated across architectures in the main text (\cref{tab:cross_model_diagnostics_id}). Bold values indicate the best method per model within each task; bold underlined values indicate the best across all models for that task.

\begin{table}[htbp]
\centering
\caption{Per-model rollout RMSE on 2D and 3D in-distribution tasks across 7 PDE benchmarks and 6 foundation models. Best per model \textbf{bolded}; best per task \underline{\textbf{underlined}}. PITA does not support 3D tasks (--).}
\label{tab:rmse_comparison}
\vspace{2pt}
\scriptsize
\resizebox{\textwidth}{!}{
\begin{tabular}{@{}ll|rrrr|rrr@{}}
\toprule
\textbf{Task} & \textbf{Model} & \textbf{Pretrained} & \textbf{SFT} & \textbf{PITA} & \textbf{OPD} & \textbf{SaT-1} & \textbf{SaT-.5} & \textbf{SaT-0} \\
\midrule
\multirow{4}{*}{\textbf{Burgers2D}}
 & DPOT-Ti    & 2.4502 & 0.0126 & 0.0103 & 0.0160 & 0.0145 & \textbf{0.0066} & 0.0110 \\
 & PDE-S      & 0.6937 & 0.0446 & 0.0515 & 0.0347 & 0.0150 & 0.0287 & \textbf{0.0120} \\
 & Poseidon-T & 0.3508 & 0.0103 & 0.0086 & 0.0084 & 0.0078 & \underline{\textbf{0.0042}} & 0.0063 \\
 & MPP-B      & 0.2760 & 0.0838 & 0.0222 & 0.0180 & 0.0181 & \textbf{0.0170} & 0.0176 \\
\midrule
\multirow{4}{*}{\textbf{KF2D}}
 & DPOT-Ti    & 2.7181 & 0.2316 & 0.3674 & 0.4018 & 0.2340 & 0.2966 & \underline{\textbf{0.1988}} \\
 & PDE-S      & 2.3416 & 1.0938 & 1.0617 & 1.1566 & 0.7190 & 1.1282 & \textbf{0.6562} \\
 & Poseidon-T & 2.2831 & 0.8494 & 0.9367 & 0.6581 & 0.5699 & 0.6701 & \textbf{0.5583} \\
 & MPP-B      & 3.0967 & 1.4343 & 0.8410 & 0.9215 & 0.8307 & 0.9265 & \textbf{0.7504} \\
\midrule
\multirow{4}{*}{\textbf{KS2D}}
 & DPOT-Ti    & 1.3853 & 0.0607 & 0.0566 & 0.0716 & 0.0600 & \textbf{0.0439} & 0.0549 \\
 & PDE-S      & 0.3646 & 0.1341 & 0.1757 & 0.1415 & 0.1219 & 0.1307 & \textbf{0.1089} \\
 & Poseidon-T & 0.4196 & 0.0397 & 0.0531 & 0.0489 & 0.0344 & \underline{\textbf{0.0278}} & 0.0335 \\
 & MPP-B      & 0.3358 & 0.2401 & 0.1484 & 0.1445 & 0.1489 & \textbf{0.1351} & 0.1389 \\
\midrule
\multirow{3}{*}{\textbf{Burgers3D}}
 & DPOT-Ti  & 0.1435 & 0.0103 & -- & 0.0125 & 0.0099 & 0.0109 & \textbf{0.0094} \\
 & MORPH-Ti & 1.2382 & 0.0154 & -- & 0.0108 & \textbf{0.0084} & 0.0095 & 0.0085 \\
 & P3D-S    & 2251.4734 & 0.0206 & -- & 0.0095 & 0.0064 & 0.0070 & \underline{\textbf{0.0061}} \\
\midrule
\multirow{3}{*}{\textbf{KS3D}}
 & DPOT-Ti  & 0.4735 & 0.1759 & -- & 0.2012 & 0.1844 & 0.1792 & \underline{\textbf{0.1677}} \\
 & MORPH-Ti & 1.3442 & 0.2352 & -- & 0.2595 & 0.2369 & 0.2306 & \textbf{0.2285} \\
 & P3D-S    & 2358.6541 & 6.2459 & -- & 0.6696 & 0.4465 & \textbf{0.2534} & 0.2752 \\
\midrule
\multirow{3}{*}{\textbf{DT3D}}
 & DPOT-Ti  & 0.1975 & 0.0557 & -- & 0.0638 & 0.0529 & 0.0570 & \textbf{0.0371} \\
 & MORPH-Ti & 0.8630 & 0.0418 & -- & \textbf{0.0308} & 0.0362 & 0.0335 & 0.0366 \\
 & P3D-S    & 972.1893 & 0.6687 & -- & 0.0248 & 0.0184 & 0.0217 & \underline{\textbf{0.0175}} \\
\midrule
\multirow{3}{*}{\textbf{KF3D}}
 & DPOT-Ti  & 0.4458 & 0.1650 & -- & 0.1222 & 0.1241 & 0.1162 & \underline{\textbf{0.1099}} \\
 & MORPH-Ti & 0.9374 & 0.1780 & -- & 0.1550 & 0.1338 & 0.1326 & \textbf{0.1299} \\
 & P3D-S    & 777.2185 & 1.7530 & -- & 0.1497 & 0.1518 & 0.1495 & \textbf{0.1478} \\
\bottomrule
\end{tabular}
}
\end{table}

\section{Detailed OOD Zero-Shot Results}
\label{sec:appendix:ood_detailed}

Table~\ref{tab:ood_detailed_combined} reports per-model RMSE for all OOD tasks. The main text (Table~\ref{tab:ood_rmse_combined}) aggregates these results as mean $\pm$ std across four 2D architectures or three 3D architectures; here we provide the full per-model breakdown. Bold values indicate the best method per model within each task; bold underlined values indicate the best across all models for that task.
\begin{table}[htbp]
\centering
\caption{Per-model RMSE on 2D and 3D OOD zero-shot tasks. Best per model \textbf{bolded}; best per task \underline{\textbf{underlined}}. PITA does not support 3D tasks (--).}
\label{tab:ood_detailed_combined}
\vspace{2pt}
\scriptsize
\resizebox{\textwidth}{!}{
\begin{tabular}{@{}ll cccccc@{}}
\toprule
\textbf{Task} & \textbf{Model} & \textbf{SFT} & \textbf{PITA} & \textbf{OPD} & \textbf{SaT-1} & \textbf{SaT-.5} & \textbf{SaT-0} \\
\midrule
\multirow{4}{*}{\textbf{Burgers2D-OOD}}
 & DPOT-Ti    & 0.0746 & 0.0713 & 0.0733 & 0.0728 & \textbf{0.0708} & 0.0723 \\
 & PDE-S & 0.0843 & 0.0905 & 0.0748 & \textbf{0.0725} & \textbf{0.0725} & \textbf{0.0725}\\
 & Poseidon-T & 0.0719 & 0.0717 & 0.0715 & 0.0707 & \underline{\textbf{0.0706}} & 0.0707 \\
 & MPP-B      & 0.1115 & 0.0746 & 0.0758 & 0.0746 & 0.0747 & \textbf{0.0735} \\
\midrule
\multirow{4}{*}{\textbf{KS2D-OOD}}
 & DPOT-Ti    & 0.0888 & 0.0898 & 0.0889 & 0.0937 & \underline{\textbf{0.0851}} & 0.0853 \\
 & PDE-S & 0.1462 & 0.1790 & 0.1517 & 0.1748 & 0.1498 & \textbf{0.1335} \\
 & Poseidon-T & 0.1395 & 0.1023 & 0.1552 & 0.1246 & \textbf{0.0865} & 0.0995 \\
 & MPP-B      & 0.1746 & 0.1088 & 0.1090 & 0.1071 & 0.1061 & \textbf{0.1059} \\
\midrule
\multirow{4}{*}{\textbf{KF2D-OOD}}
 & DPOT-Ti    & 2.0029 & 2.1528 & 2.1561 & 2.0670 & 2.0947 & \underline{\textbf{1.9956}} \\
 & PDE-S & 2.6950 & 2.7148 & 2.6940 & \textbf{2.3248} & 2.7282 & 2.3858 \\
 & Poseidon-T & 2.3918 & 2.3602 & 2.2884 & 2.3838 & 2.4473 & \textbf{2.1338} \\
 & MPP-B      & 2.7200 & 2.6462 & 2.5519 & 2.5508 & 2.7253 & \textbf{2.5457} \\
\midrule
\multirow{3}{*}{\textbf{Burgers3D-OOD}}
 & DPOT-Ti  & 0.0154 & -- & 0.0171 & 0.0147 & 0.0168 & \textbf{0.0143} \\
 & MORPH-Ti & 0.0161 & -- & 0.0130 & 0.0127 & 0.0133 & \textbf{0.0125} \\
 & P3D-S    & 0.0208 & -- & 0.0127 & \underline{\textbf{0.0119}} & 0.0121 & 0.0120 \\
\midrule
\multirow{3}{*}{\textbf{KF3D-OOD}}
 & DPOT-Ti  & 0.2234 & -- & 0.1964 & 0.1933 & 0.1895 & \underline{\textbf{0.1866}} \\
 & MORPH-Ti & 0.2455 & -- & 0.2097 & 0.1994 & 0.1989 & \textbf{0.1955} \\
 & P3D-S    & 2.7287 & -- & 0.2001 & 0.2037 & 0.2006 & \textbf{0.1967} \\
\midrule
\multirow{3}{*}{\textbf{KS3D-OOD}}
 & DPOT-Ti  & 0.1748 & -- & 0.2031 & 0.1763 & 0.1724 & \underline{\textbf{0.1614}} \\
 & MORPH-Ti & 0.2161 & -- & 0.2424 & 0.2221 & \textbf{0.2117} & 0.2155 \\
 & P3D-S    & 7.6484 & -- & 0.4237 & 0.2029 & 0.2027 & \textbf{0.1990} \\
\midrule
\multirow{3}{*}{\textbf{DT3D-OOD}}
 & DPOT-Ti  & 0.0533 & -- & 0.0594 & 0.0514 & 0.0540 & \textbf{0.0404} \\
 & MORPH-Ti & 0.0428 & -- & 0.0338 & 0.0379 & 0.0381 & \underline{\textbf{0.0337}} \\
 & P3D-S    & 0.6499 & -- & 0.0371 & 0.0383 & 0.0376 & \textbf{0.0368} \\
\bottomrule
\end{tabular}
}
\end{table}

\clearpage
\section{Full 3D Trajectory Visualizations}
\label{sec:appendix:3d_trajectories}

We visualize full 50-step autoregressive rollouts on the four 3D PDE tasks at $64^3$ resolution.  Each figure shows volumetric iso-surface snapshots at steps $t \in \{1, 10, 20, 30, 40, 50\}$ (rows), comparing the solver reference, the pretrained backbone, SFT, OPD, and SaT-0 (columns).  These complement the quantitative RMSE results in Tables~\ref{tab:rmse_comparison} and~\ref{tab:ood_detailed_combined} by exposing how field morphology degrades over time for each training method.

\begin{figure}[htbp]
  \centering
  \includegraphics[width=\textwidth]{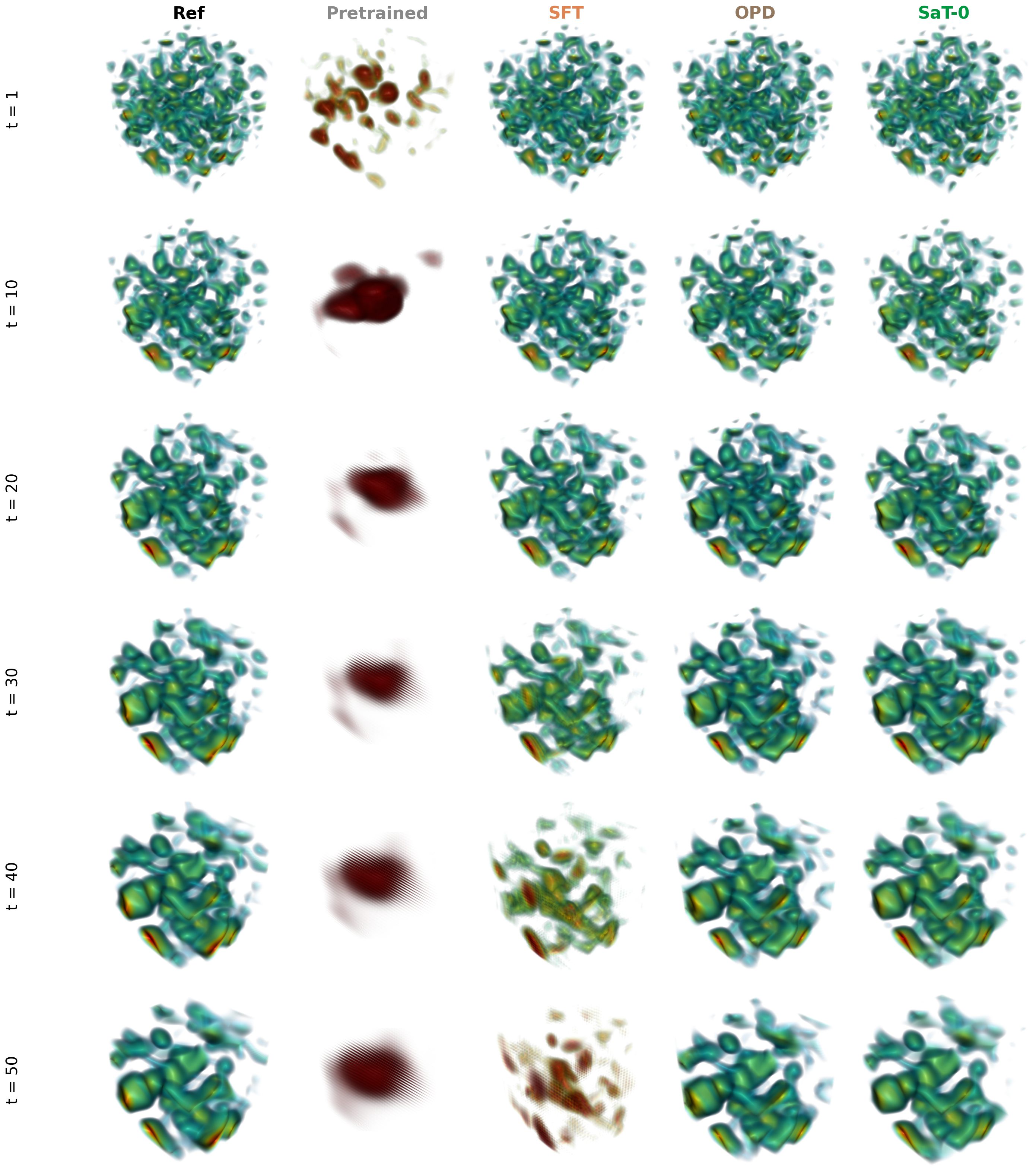}
  \caption{Burgers3D (P3D-S, trajectory 0). The pretrained backbone collapses into a single large-scale blob by $t=10$; SFT tracks the reference well until $t\!\approx\!30$ before developing grid-level artifacts at $t=40$--$50$. OPD and SaT-0 both preserve the fine-scale shock structure throughout the full 50-step rollout, with SaT-0 maintaining the closest visual match to the reference at late times.}
  \label{fig:burgers3d_traj}
\end{figure}

\begin{figure}[htbp]
  \centering
  \includegraphics[width=\textwidth]{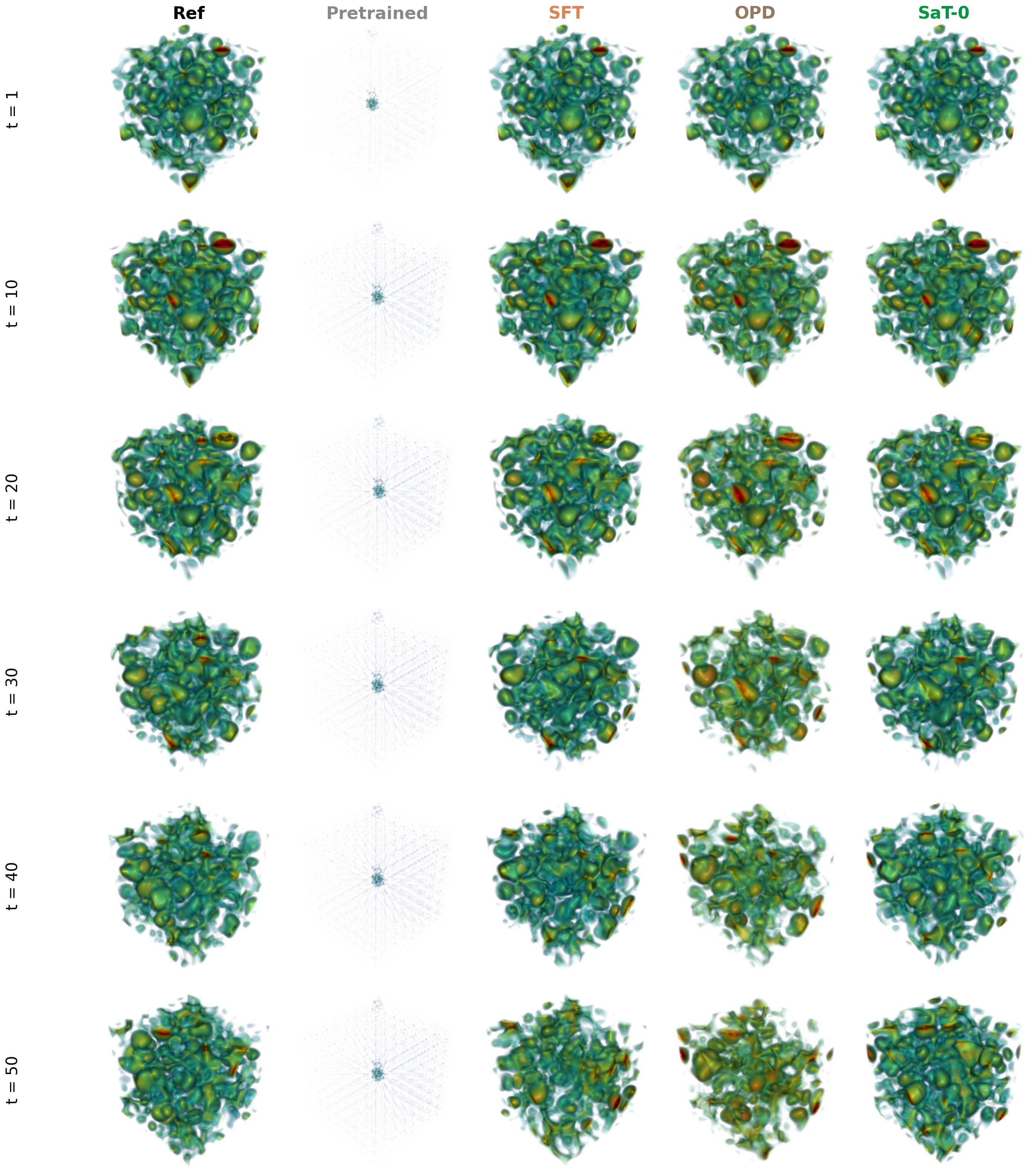}
  \caption{KS3D --- Kuramoto--Sivashinsky (DPOT-Ti, trajectory 3). The chaotic dynamics of the KS equation make long-horizon tracking particularly challenging. The pretrained backbone diverges immediately. SFT maintains qualitative agreement until $t\!\approx\!20$ but loses fine-scale cellular patterns at later steps. Both OPD and SaT-0 preserve the characteristic cellular structure throughout, with SaT-0 exhibiting the sharpest structures and most accurate magnitude at $t=50$.}
  \label{fig:ks3d_traj}
\end{figure}

\begin{figure}[htbp]
  \centering
  \includegraphics[width=\textwidth]{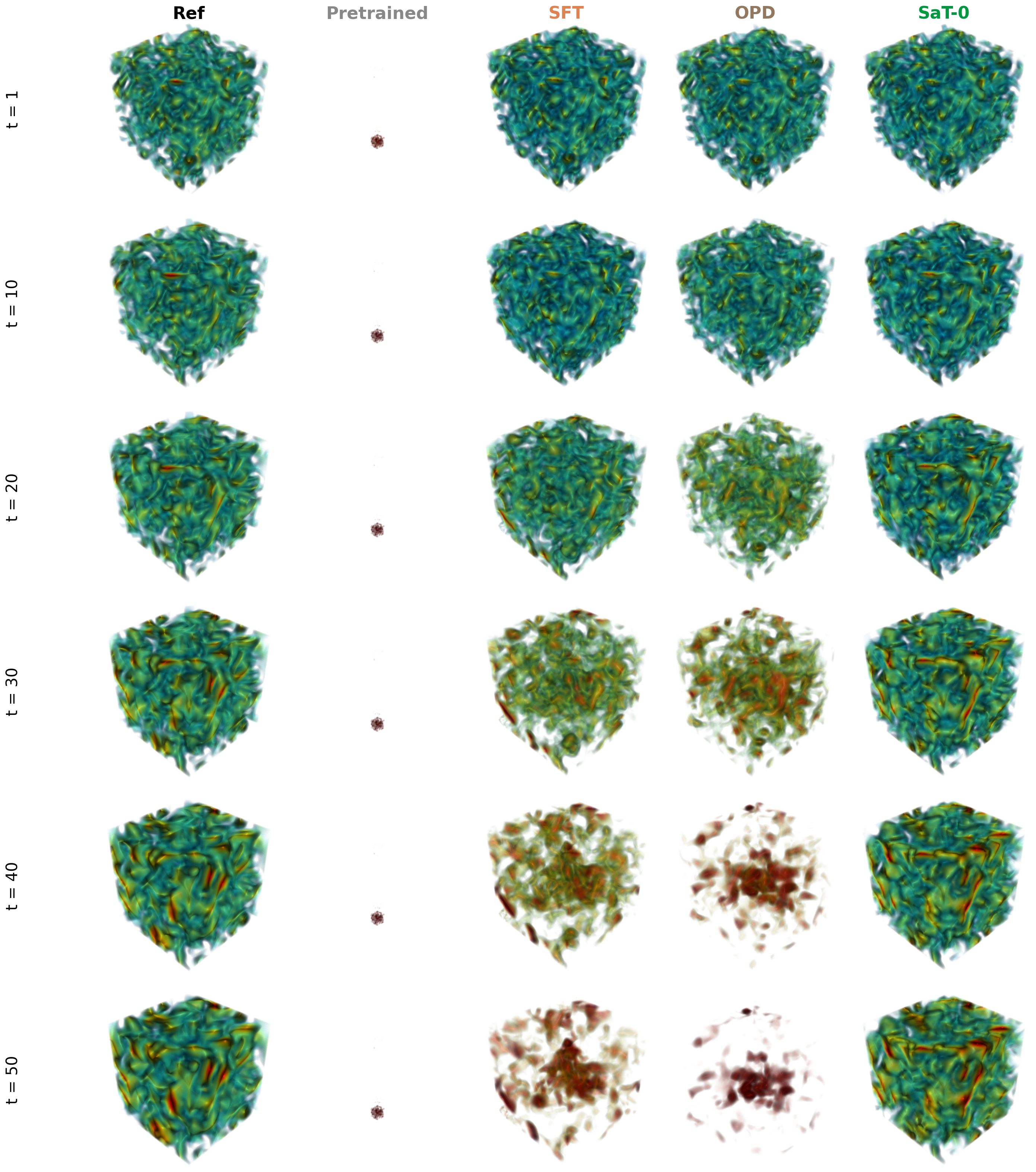}
  \caption{DT3D --- Decaying Turbulence velocity (DPOT-Ti, trajectory 2). As kinetic energy decays, fine-scale vortices merge and dissipate. The pretrained model fails to reproduce this cascade. SFT captures the large-scale decay but over-smooths the remaining eddies. OPD and SaT-0 reproduce both the energy decay rate and the vortex morphology more faithfully; SaT-0 retains the sharpest eddy boundaries at late times.}
  \label{fig:dt3d_traj}
\end{figure}

\begin{figure}[htbp]
  \centering
  \includegraphics[width=\textwidth]{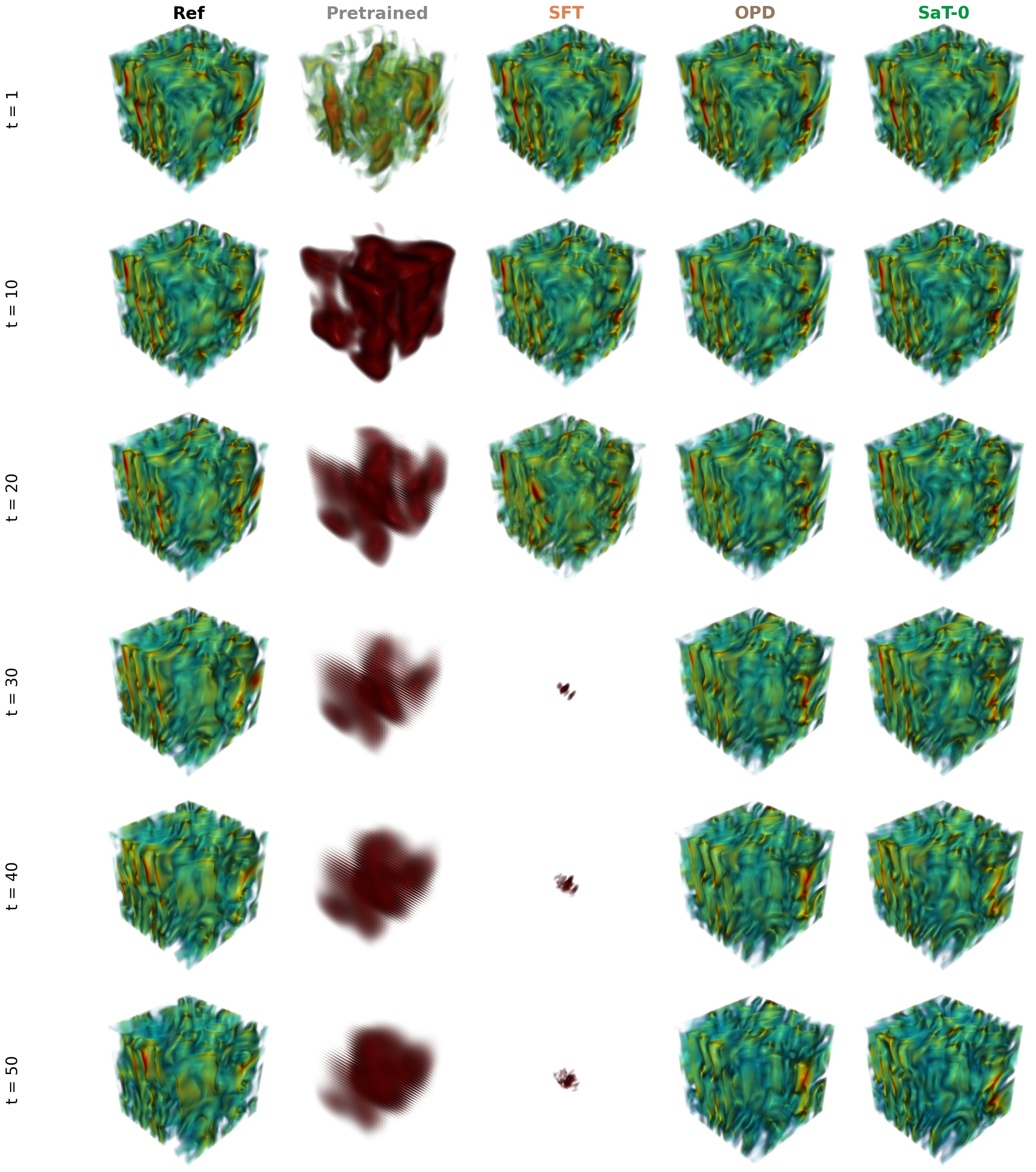}
  \caption{KF3D --- Kolmogorov Flow velocity (P3D-S, trajectory 0). The pretrained model fails to produce physically meaningful velocity iso-surfaces. SFT captures the large-scale flow topology but progressively over-smooths fine vortical structures after $t=20$. OPD and SaT-0 retain multi-scale vortex features up to $t=50$; SaT-0 shows slightly sharper structures at late times.}
  \label{fig:kf3d_traj}
\end{figure}

\begin{figure}[htbp]
  \centering
  \includegraphics[width=\textwidth]{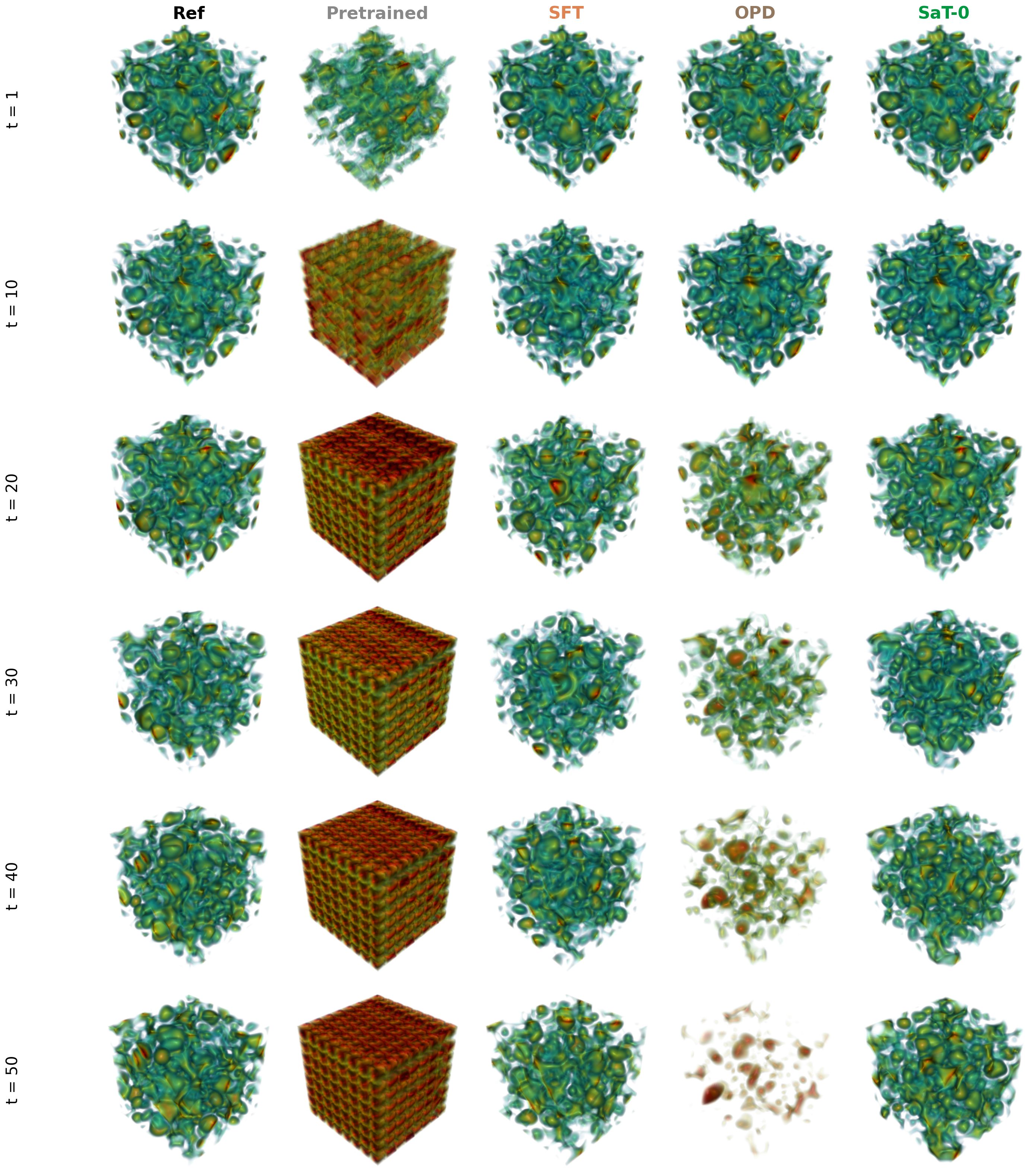}
  \caption{KS3D --- Kuramoto--Sivashinsky (MORPH-Ti, trajectory 4). The chaotic dynamics make long-horizon tracking particularly challenging. The pretrained backbone diverges immediately. OPD maintains qualitative agreement at early steps but loses fine-scale cellular patterns at later steps. Both SFT and SaT-0 preserve the characteristic cellular structure throughout the full 50-step rollout, with SaT-0 showing sharper structures at late times.}
  \label{fig:ks3d_morph_traj}
\end{figure}

\begin{figure}[htbp]
  \centering
  \includegraphics[width=\textwidth]{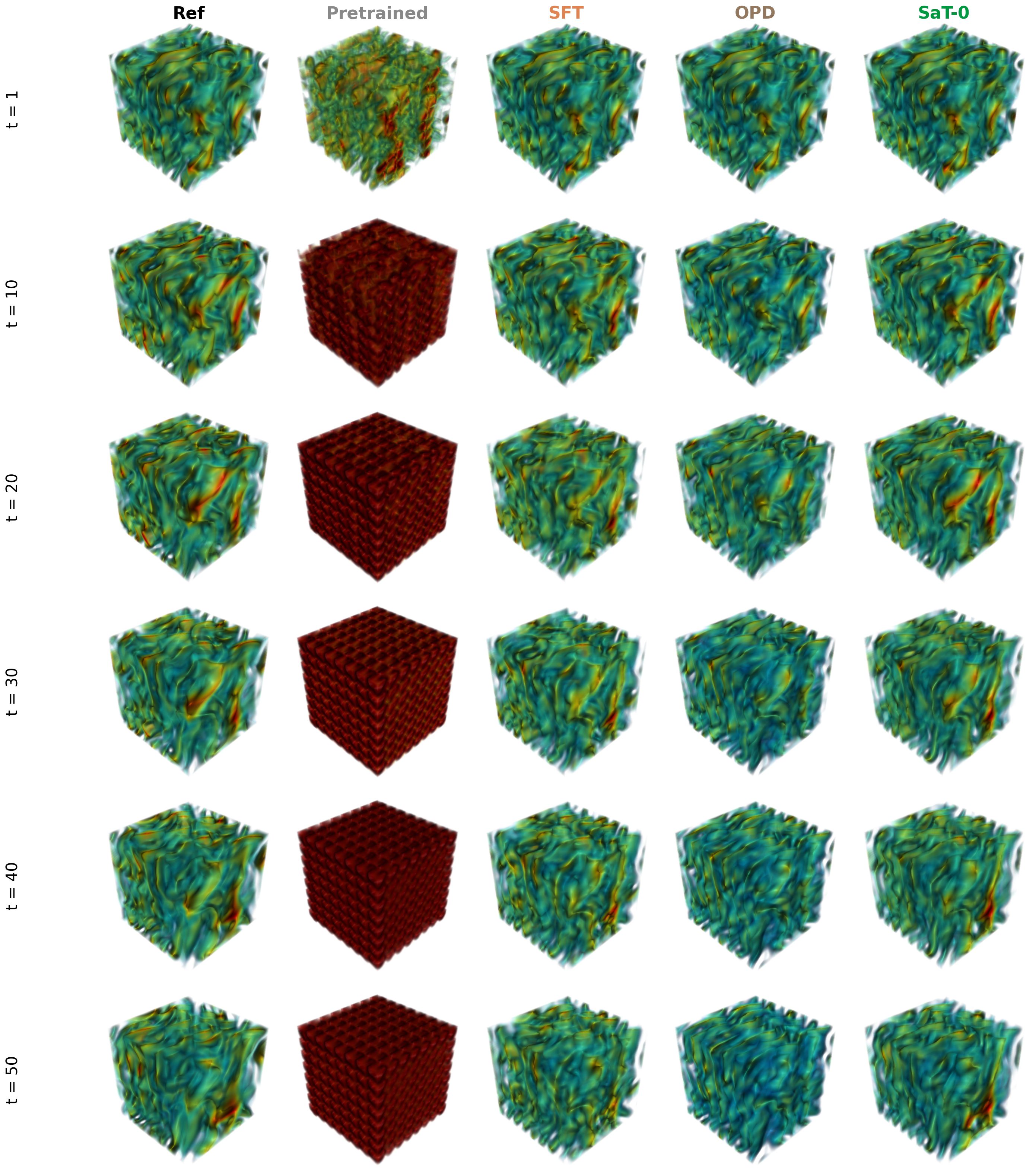}
  \caption{KF3D --- Kolmogorov Flow velocity (MORPH-Ti, trajectory 4). The pretrained model fails to produce physically meaningful velocity iso-surfaces. OPD and SFT capture the large-scale flow topology but progressively over-smooths fine vortical structures after $t=20$. SaT-0 retains multi-scale vortex features up to $t=50$ and shows sharper structures at late times.}
  \label{fig:kf3d_morph_traj}
\end{figure}

\end{document}